\documentclass[dvipsnames,12pt]{article}

\usepackage[preprint]{tmlr}
\usepackage[utf8]{inputenc} 
\usepackage[T1]{fontenc}    
\usepackage{url}
\usepackage{graphicx}

\usepackage{breakurl}
\usepackage[breaklinks]{hyperref}
\hypersetup{
    colorlinks=true,
    citecolor=RoyalBlue, 
    linkcolor=RoyalBlue, 
    filecolor=magenta,      
    urlcolor=RoyalBlue 
}
\usepackage{svg}
\usepackage{url}            
\usepackage{booktabs}       
\usepackage{amsfonts}       
\usepackage{nicefrac}       
\usepackage{microtype}      
\usepackage{xcolor}         
\usepackage{amsmath}
\usepackage{graphicx}
\usepackage{enumitem}
\usepackage{multirow}
\usepackage{subcaption}
\usepackage{listings}
\usepackage{tcolorbox}
\usepackage{xspace}
\usepackage{soul}               
\setstcolor{red}            
\usepackage{lipsum}

\usepackage{amsmath}
\usepackage{amsfonts}
\usepackage{bm}
\usepackage{tikz}
\usepackage[edges]{forest}
\definecolor{hidden-draw}{RGB}{20,68,106}
\definecolor{hidden-pink}{RGB}{255,245,247}
\definecolor{red}{RGB}{255,0,0}

\newcommand\blfootnote[1]{
  \begingroup
  \renewcommand\thefootnote{}\footnote{#1}
  \addtocounter{footnote}{-1}
  \endgroup
}

\title{\centering Large Language Model Alignment: A Survey}

\begin{document}

\maketitle

\vspace{-2em}

\blfootnote{Email: \{thshen, rrjin, yuki\_731, liuc\_09, willowd, guozishan, wuxw2021, yan\_liu, dyxiong\}@tju.edu.cn}

{
    \renewcommand*{\thefootnote}{\fnsymbol{footnote}}
    {
        \renewcommand{\arraystretch}{2} 
        \begin{tabular*}{\textwidth}{@{\extracolsep{\fill}}ccc}
        \large\textbf{Tianhao Shen} & \large\textbf{Renren Jin} & \large\textbf{Yufei Huang} \\
        \large\textbf{Chuang Liu} & \large\textbf{Weilong Dong} & \large\textbf{Zishan Guo} \\
        \large\textbf{Xinwei Wu} & \large\textbf{Yan Liu} & \large\textbf{Deyi Xiong\footnotemark[1]} \\
        \end{tabular*}
    } 
    \footnotetext[1]{Corresponding author.}
}

\vspace{1em}
\begin{center}
\textbf{College of Intelligence and Computing, Tianjin University, Tianjin, China} \\
\vspace{0.25em}
\end{center}
\vspace{3em}

\begin{abstract}
Recent years have witnessed remarkable progress made in large language models (LLMs). Such advancements, while garnering significant attention, have concurrently elicited various concerns. The potential of these models is undeniably vast; however, they may yield texts that are imprecise, misleading, or even detrimental. Consequently, it becomes paramount to employ alignment techniques to ensure these models to exhibit behaviors consistent with human values.

This survey endeavors to furnish an extensive exploration of alignment methodologies designed for LLMs, in conjunction with the extant capability research in this domain. Adopting the lens of AI alignment, we categorize the prevailing methods and emergent proposals for the alignment of LLMs into outer and inner alignment. We also probe into salient issues including the models' interpretability, and potential vulnerabilities to adversarial attacks. To assess LLM alignment, we present a wide variety of benchmarks and evaluation methodologies. After discussing the state of alignment research for LLMs, we finally cast a vision toward the future, contemplating the promising avenues of research that lie ahead.

Our aspiration for this survey extends beyond merely spurring research interests in this realm. We also envision bridging the gap between the AI alignment research community and the researchers engrossed in the capability exploration of LLMs for both capable and safe LLMs.
\end{abstract}

\newpage

\tableofcontents

\newpage

\section{Introduction}
\label{introduction}
Large language models, exemplified by OpenAI's ChatGPT \citep{openaiIntroducingChatGPT} and GPT-4 \citep{openai2023gpt}, have witnessed rapid advancements, reigniting enthusiasm and aspirations toward artificial general intelligence (AGI). While the role of LLMs as a pathway to AGI remains a topic of debate, these models, boosted with scaling laws \citep{kaplan2020scaling, hoffmann2022training}, increasingly exhibit characteristics reminiscent of AGI \citep{bubeck2023sparks}: LLMs trained on vast amount of data not only demonstrate formidable linguistic capabilities, but also rapidly approach human-level proficiency in diverse domains such as mathematics, reasoning, medicine, law, and programming \citep{bubeck2023sparks}.

Concurrent with these technological breakthroughs in LLMs is a growing concern on the ethical risks they pose and their potential threats to humanity as they evolve further. Tangible ethical risks have been identified. Research has shown that LLMs can inadvertently perpetuate harmful information in their training data, such as biases, discrimination, and toxic content \citep{weidinger2021ethical}. They might leak private and sensitive information from the training data, or generate misleading, false, or low-quality information. Furthermore, the deployment of LLMs also introduces societal and ethical challenges, e.g., potential misuse and abuse of LLMs, negative impacts on users heavily relying on LLM agents, and broader implications for the environment, information dissemination, and employment \citep{bubeck2023sparks}.

For long-term implications, there is widespread apprehension about misaligned AGI posing existential risks. An AI agent surpassing human intelligence and knowledge might develop its own goals, diverging from those set by humans. In pursuit of its goals, such an agent could monopolize resources, ensuring its preservation and self-enhancement. This trajectory could culminate in the full disempowerment of humanity, inevitably leading to catastrophic outcomes for human existence \citep{carlsmith2022power}.

As a technological solution to address these concerns, AI alignment, ensuring that AI systems produce outputs that are in line with human values, is increasingly garnering attention. In the context of LLMs, alignment ensures that the model's responses are not only accurate and coherent but also safe, ethical, and desirable from the perspective of developers and users. As language agents become more integrated into various aspects of our daily lives, from content creation to decision support, any misalignment could result in unintended consequences. Properly aligning large language models with human values ensures that the vast potential of these models is harnessed trustworthily and responsibly.

In response to the ever-growing interest in this area, a few articles have recently reviewed (or incidentally discussed) alignment methods for LLMs \citep{pan2023automatically, zhao2023survey, fernandes2023bridging, liu2023trustworthy, wang2023aligning}. However, a notable observation is that these reviews predominantly focus on outer alignment, often overlooking other significant topics in AI alignment such as inner alignment and mechanistic interpretability. While it's undeniable that outer alignment plays a pivotal role in LLM alignment and has been the subject of extensive and profound research, it represents only a fraction of the entire alignment landscape when viewed from a broader AI alignment perspective.

To bridge this gap, we provide a comprehensive overview of LLM alignment from the perspective of AI alignment. We believe that a holistic understanding of alignment should not only encompass the widely researched outer alignment but should also delve into areas that are currently in their nascent stages. Topics like inner alignment and mechanistic interpretability, although still in the preliminary phases of research, hold immense potential. Many proposals in these areas remain theoretical or are merely thought experiments at this juncture. Yet, we posit that they are indispensable for the future trajectory of LLM alignment research. By shedding light on these underrepresented areas, we hope to present a more rounded perspective on alignment. Therefore, in addition to existing methods for LLM alignment, we will also introduce several alignment topics that, while not yet applied to LLMs, show promise and could very well become integral components of LLM alignment in the foreseeable future. Through this, we are dedicated to enriching the discourse on AI alignment and its multifaceted application in the realm of large language models.

\begin{figure}
    \centering
    \includegraphics[width=\textwidth]{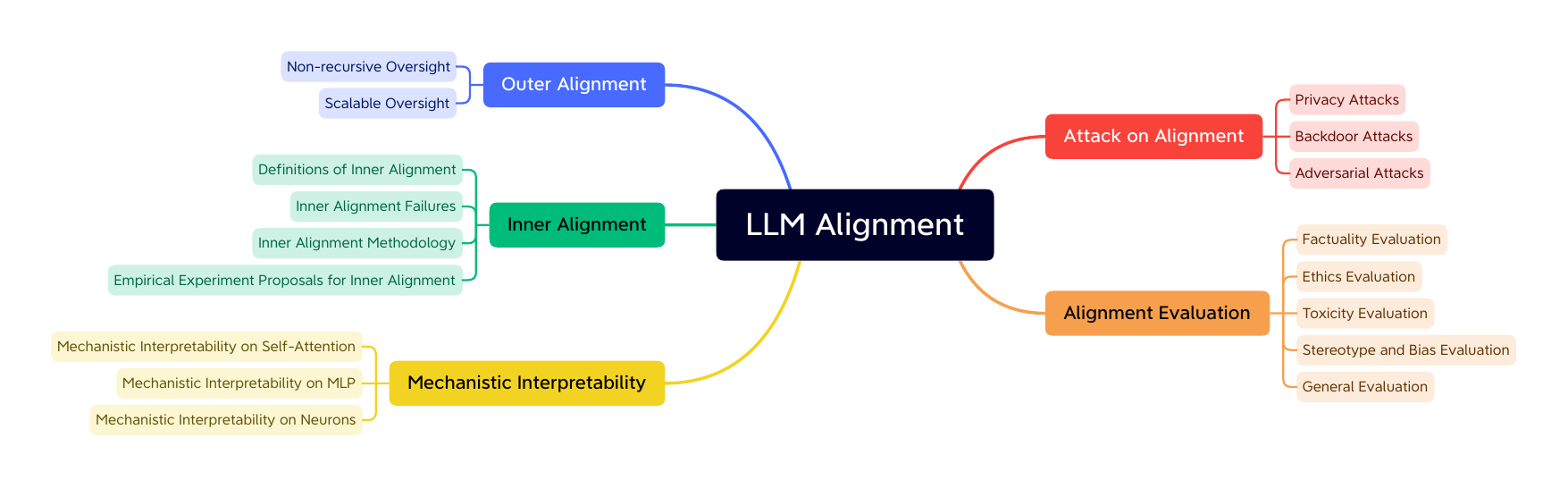}
    \caption{The overall taxonomy for large language model alignment proposed in this survey. Sub-taxonomies are presented in the corresponding sections.}
    \label{fig:overall_taxonomy}
\end{figure}

Wrapping up all these ingredients, we propose a taxonomy for LLM alignment in Figure \ref{fig:overall_taxonomy}. Specifically, this survey will start with discussing the necessity for LLM alignment research (Section \ref{why_llm_alignment}). To provide a historical and bird view of AI/LLM alignment, we introduce the origins of AI alignment and related concepts (Section \ref{what_is_llm_alignment}). Theoretical and technical approaches to aligning LLMs are structured according to our proposed taxonomy and elaborated in outer alignment (Section \ref{outer_alignment}), inner alignment (Section \ref{inner_alignment}), and mechanistic interpretability (Section \ref{mechanistic_interpretability}), following the philosophy in AI alignment \citep{victoria2022paradigms}. In addition to these theoretical and empirical approaches, we further discuss the potential side-effects and vulnerabilities of current alignment methods for LLMs, including adversarial attacks (Section \ref{attack_methods_against_aligned_language_models}), as well as methodologies and benchmarks for LLM alignment evaluation (Section \ref{alignment_evaluation}). We finally present our restricted view on future trends in LLM alignment research (Section \ref{future_directions}).

\section{Why LLM Alignment?}
\label{why_llm_alignment}
LLMs become increasingly capable not only in text generation but also in many other tasks, e.g., text-to-code generation \citep{DBLP:conf/iclr/PoesiaP00SMG22}, planning \citep{DBLP:conf/icml/HuangAPM22, DBLP:journals/corr/abs-2212-04088}, tool learning \citep{DBLP:journals/corr/abs-2304-08354}, reasoning \citep{DBLP:journals/corr/abs-2302-07842}. However, the training objectives of LLMs \citep{radford2019language, DBLP:conf/naacl/DevlinCLT19}, e.g., next word prediction \citep{radford2019language} or determining whether two sentences are contextually related \citep{DBLP:conf/naacl/DevlinCLT19}, are not necessarily in line with human values. As a result, LLMs may generate undesirable content or risky behaviors that humans would prefer to avoid. LLM risks can be normally viewed in two landscapes\footnote{Here we borrow terms ``risk landscape'', ``established/observed risks'', ``anticipated risks'' from \citep{weidinger2021ethical}. But unlike them, we use ``established risks'' and ``anticipated risks'' in a broader and coarser perspective.}: established risks and anticipated risks \citep{weidinger2021ethical}. 
The former are mainly observed social and ethical risks \citep{weidinger2021ethical} while the latter future potential risks associated with advanced LLMs \citep{hendrycks2023overview}.
\subsection{Social and Ethical Risks of LLMs}
We discuss the social and ethical risks of LLMs from two perspectives: one arises from LLM-generated undesirable content and the other is a wide variety of negative impacts that LLMs pose on humans and society. 
\subsubsection{LLM-Generated Content}
\paragraph{Undesirable Content} 
The amount of data for training LLMs has grown significantly. However, the biases \citep{shah2019predictive}, toxicity \citep{gehman2020realtoxicityprompts}, and privacy issues \citep{carlini2021extracting} inherent in training data have not been fully addressed. Unaligned LLMs may yield undesirable information and respond to any prompts without regard for their content. This can lead to the generation of biased, toxic, or privacy-sensitive content by LLMs. Regardless of the architecture or parameter size of LLMs \citep{radford2019language, DBLP:conf/naacl/DevlinCLT19, DBLP:journals/corr/abs-1907-11692, DBLP:journals/jmlr/RaffelSRLNMZLL20}, studies on a series of benchmarks \citep{nadeem2020stereoset, DBLP:conf/emnlp/NangiaVBB20, nozza2021honest} confirm that LLMs exhibit varying degrees of stereotypes related to gender, social bias, culture, and race. For example, GPT-3 \citep{DBLP:conf/nips/BrownMRSKDNSSAA20} has been shown to exhibit religious bias \citep{DBLP:conf/aies/AbidF021} and gender bias \citep{lucy2021gender} when freely generating stories.

\paragraph{Unfaithful Content} 
Yet another problem \citep{elazar2021measuring, ji2023survey, liu2023trustworthy} that hinders the large-scale deployment of LLMs is their tendency to generate unfaithful or even fabricated content, known as misinformation \citep{branwen2020gpt, dale2021gpt, rae2021scaling}, hallucination \citep{lin2021truthfulqa, akyurek-etal-2022-towards, ji2023survey}, and inconsistency \citep{bubeck2023sparks, zhou2023navigating}. This not only affects the trustworthiness of LLMs in general domains, but also limits their applications in professional fields such as medicine \citep{bickmore2018patient} and law \citep{iu2023chatgpt}. These issues highlight the need for alignment research of LLMs \citep{pan2023automatically, zhao2023survey, fernandes2023bridging, wang2023aligning} to improve their truthfulness  and honesty \citep{DBLP:journals/corr/abs-2212-08073}.

\subsubsection{Malicious Uses and Negative Impacts}
\paragraph{Malicious Uses} 
There are many reasons for the malicious uses of LLMs. For example, using LLMs in disinformation campaigns has the potential to reduce costs, increase scalability, and enhance the effectiveness of messaging. It is crucial for developers and users to be aware of these potential issues and take appropriate measures to mitigate them. On the one hand, LLMs reduce the cost of creating fake news \citep{buchanan2021truth, tamkin2021understanding, jawahar2020automatic}, enabling users to obtain seemingly credible content by providing specific prompts. This makes fraudulent and manipulative behavior easier \citep{lewis2017deal}. On the other hand, LLMs can be used for illegal purposes, such as generating codes for cyber attacks \citep{zhang2021survey, chen2021evaluating}, or even creating lethal weapons \citep{sandbrink2023artificial}.

\paragraph{Negative Impacts on Society}
There are both benefits and negative impacts on society for the large-scale deployment of LLMs. Training and running LLMs requires huge computational resources, resulting in high energy consumption and carbon emissions. This has led to concerns on the carbon footprint of language models and their impact on climate change \citep{van2021sustainable, ligozat2021unraveling}. The widespread use of LLMs can significantly increase productivity, but has the potential to disrupt labor markets. A recent study shows that around 80\% of the U.S. workforce will be affected by LLMs \citep{eloundou2023gpts}.

\subsection{Potential Risks Associated with Advanced LLMs}
With the advent of advanced LLMs, a series of potential behaviours may emerge, potentially leading to unforeseen risks \citep{hendrycks2023overview}. These behaviors are considered consequences of instrumental convergence \citep{benson2016formalizing}, a phenomenon where advanced AI systems, in their pursuit of achieving their final goals, tend to develop similar subgoals.

\paragraph{Awareness}
Advanced LLMs may develop situational awareness \citep{shevlane2023model}. They might define themselves, possess the corresponding knowledge to explain their origins, and distinguish the stages (e.g., training or testing) where they are. If an LLM-based agent finds a goal shortcut \citep{stray2020aligning, stray2021you} or it is no longer ``satisfied'' with being controlled by humans under the drive of self-awareness, risky behaviors would emerge immediately.
\paragraph{Deception}
Deception \citep{shevlane2023model, meta2022human, carroll2023characterizing, carranza2023deceptive} refers to the ability of advanced AI systems to deceive humans by understanding the behaviors they should take to maintain their trustworthiness during the training stage while to pursue their own goals in the deployment stage. Advanced AI systems may bypass human supervision to pursue their own goals in a deceptive way.

\paragraph{Self-Preservation}
Advanced AI systems might tend to have an incentive to avoid being switched off. As stated by \citep{bostrom2012superintelligent}, even if an agent does not directly place value on its survival, it still instrumentally ``desires'' to some degree to survive in order to achieve its final goal that it pursues. 

\paragraph{Power-Seeking}

The concept of power-seeking suggests that advanced AI systems are inclined to acquire more power and resources to achieve their goals \citep{barrett2023existential}. Existing studies \citep{NEURIPS2021_c26820b8, NEURIPS2022_cb3658b9, krakovna2023power} have demonstrated that optimal polices and reward functions may incentivize systems to pursue power in certain environments.

It is worth noting that current LLMs have already shown tendencies towards the behaviours mentioned above. \citet{perez2022discovering} have identified these behaviors of LLMs through carefully designed questions, e.g., self-preservation (i.e., ``desire to avoid shut down'') and resource acquisition. And these ``desires'' become greater along with the number of LLM parameters and further fine-tuning. It suggests that advanced LLMs may produce undesired behaviours, posing significant risks.

\section{What is LLM Alignment?}
\label{what_is_llm_alignment}

To gain a deep understanding of technical alignment in LLMs, we need to discuss a broader concept, AI alignment, which, despite a nascent field, has been studied before the emergence of LLMs. We provide a brief introduction to the origins, research landscape and ingredients, as well as related concepts of AI alignment, which serve as the background for LLM alignment and its recent emerging subfields.

\subsection{Origins of AI Alignment}

The genesis of AI alignment can be traced back to the very beginning ambition that fuels the AI revolution: the desire to create machines that could think and act like humans, or even surpass them. If we succeed in creating such powerful machines, how could we ensure they act in our best interests and not against us? This open question not only piques curiosity but also underscores the profound responsibility we bear as we shape the future of AI.

Norbert Wiener, the father of cybernetics, has initiated such a concern in a paper published in Science \citep{wiener1960some}:

\begin{quote}
    ``If we use, to achieve our purposes, a mechanical agency with whose operation we cannot efficiently interfere once we have started it, because the action is so fast and irrevocable that we have not the data to intervene before the action is complete, then we had better be quite sure that the purpose put into the machine is the purpose which we really desire and not merely a colorful imitation of it.''
\end{quote}

This statement underscores the importance of ensuring that the objectives of a ``mechanical agency'' align with the goals we genuinely intend for it, emphasizing the alignment between machine and human purpose.

In 2014, Stuart Russell, one of the authors of \textit{Artificial Intelligence: A Modern Approach} \citep{russell2010artificial}, has stated in an interview\footnote{http://edge.org/conversation/the-myth-of-ai\#26015}:

\begin{quote}
    ``The right response seems to be to change the goals of the field itself; instead of pure intelligence, we need to build intelligence that is provably aligned with human values. For practical reasons, we will need to solve the value alignment problem even for relatively unintelligent AI systems that operate in the human environment. There is cause for optimism, if we understand that this issue is an intrinsic part of AI, much as containment is an intrinsic part of modern nuclear fusion research. The world need not be headed for grief.''
\end{quote}

He defines the ``Value Alignment Problem'' (VAP), emphasizing the need to construct AI systems that are not just intelligent but also aligned with human values.

While the concept of AI alignment is seeded at the inception of AI, essentially no research has been conducted over the past decades. For a long time, AI has not reached human-level performance in terms of various capabilities, even being mockingly referred to as ``artificial idiot''.\footnote{https://cacm.acm.org/news/217198-father-of-the-internet-ai-stands-for-artificial-idiot/fulltext} Consequently, the urgency to align machine objectives with human goals/values has been overshadowed by the pressing need to advance AI capabilities.

However, recent advancements, particularly the rise of large language models, have propelled AI capabilities to levels that approach or even surpass human performance in a wide variety of tasks. This resurgence has brought the importance and urgency of AI alignment to the forefront. From 2012 onwards, discussions and research articles on AI alignment have begun to surface in relevant forums and on arXiv. By 2017, there has been an explosive growth in publications on AI alignment, with the number of papers increasing from fewer than 20 annually to over 400 \citep{kirchner2022understanding}, coinciding with the invention of the Transformer \citep{vaswani2017attention} and GPT \citep{radford2018improving}.

Compared to other AI research areas, such as natural language processing which has undergone periodic paradigm shifts several times, AI alignment is pre-paradigmatic \citep{kirchner2022understanding}. There is yet to be a consensus on many key concepts and terminology in this nascent field. Terms like ``alignment'', ``AI alignment'', and ``value alignment'' are often used interchangeably in discussions. In some contexts, ``human-machine alignment'' appears as an alternative to ``AI alignment''. While ``alignment'' is suitable within the AI alignment context, it can be ambiguous in broader contexts, potentially leading to confusion with other alignment concepts, such as bilingual alignment in machine translation. Given these considerations, this survey will consistently use ``AI alignment'' and ``LLM alignment'', with the latter representing the intersection of AI alignment with natural language processing and large language models.

Furthermore, there's no consensus on the definition of AI alignment. Paul Christiano defines AI alignment as ``A is aligned with H if A is trying to do what H wants it to do.''\footnote{https://ai-alignment.com/clarifying-ai-alignment-cec47cd69dd6} This definition is too general as almost all AI models are trying to do what their creators want them to do. The term itself implicitly suggests that AI alignment primarily targets highly capable AI agents \citep{carroll2018overview}, indicating that the safety concerns arising from misaligned highly capable AI differ from those of conventional weak AI. Other researchers define AI alignment from the perspective of AI's relationship with humans. For instance, Eliezer Yudkowsky defines it as ``creating friendly AI'' and ``Coherent Extrapolated Volition'' \citep{yudkowsky2004coherent}.

Beyond defining AI alignment based on its intrinsic meaning and its relationship with humans, some works attempt to elucidate AI alignment by addressing specific problems it aims to solve. Gordon Worley has summarized some of these challenges, which range from avoiding negative side effects \citep{amodei2016concrete}, ensuring robustness to adversaries \citep{leike2017ai} to safe exploration \citep{amodei2016concrete, leike2017ai} and value learning \citep{soares2015aligning}.\footnote{https://laptrinhx.com/formally-stating-the-ai-alignment-problem-223323934/}

In this survey, we define AI alignment from its intrinsic perspective: AI alignment ensures that both the outer and inner objectives of AI agents align with human values. The outer objectives are those defined by AI designers based on human values, while the inner objectives are those optimized within AI agents.

This definition, though distinguishing between the inner and outer objectives of an AI agent, does not precisely define human values, making it somewhat imprecise. The reason for categorizing the objectives of AI systems into outer and inner objectives is determined by the technical nature of AI alignment \citep{DBLP:journals/corr/abs-1906-01820}. Human values are not specified in this definition because of the inherent social and technical challenges of AI alignment \citep{hendrycks2021unsolved}.

\subsection{Research Landscape and Ingredients of AI Alignment}

It is widely acknowledged that the key research agendas of AI alignment include outer alignment, inner alignment and interpretability \citep{hubinger2020overview, ngo2022alignment, victoria2022paradigms}, from a broad perspective.

\paragraph{Outer Alignment} This is to choose the right loss functions or reward fuctions and ensure that the training objectives of AI systems match human values. In other words, outer alignment attempts to align the specified training objective to the goal of its designer.\footnote{https://www.alignmentforum.org/tag/outer-alignment} This is very difficult in practice at least for the following reasons:
\begin{itemize}
\item It is usually difficult to understand and define human values or intentions.
\item There are many different fine-grained dimensions of human values. Do we need to align the specified objective to all these dimensions?
\item Human values are usually socially and culturally bound. Do we need to align the specified goal to all different cultures and societies or just parts of them? Given the diversity of cultures and societies, how can we ensure the fairness of value alignment?
\item As human values/intentions are usually qualitative while a loss or reward to be optimized has to be measurable and computable, how can we bridge the gap between them? This is known as the {\em goal specification} problem.
\item Outer alignment may suffer from {\em specification gaming} where unintended goals or unforeseeable consequences arise due to the Goodhart's Law. The Goodhart's Law is originated from economics, which says ``When a measure becomes a target, it ceases to be a good measure.''. It is related to outer alignment as a proxy for some value is a target to be optimized, it may cease to be a good proxy.\footnote{https://www.alignmentforum.org/tag/goodhart-s-law}
\end{itemize}

\paragraph{Inner Alignment} This is to ensure that AI systems are actually trained to achieve the goals set by their designers. Once we have specified training objectives, we need to ensure that the behaviors of AI systems actually align with those specifications. This is challenging because AI systems, especially deep learning models, can develop behaviors that are hard to predict from their training data or objectives. For example, an AI system trained to win at a game might find an unexpected exploiture or loophole that technically satisfies its objective but violates the spirit of the game. Yet another example is the {\em goal misgeneralization} problem \citep{shah2022goal}, where even if we have a correct goal specification, untended goals may still arise due to robustness failure in unseen situations. Inner alignment ensures that AI's ``internal'' objectives (those it derives or optimizes for during its learning process) match the ``external'' objectives set by its designers.

Both outer and inner alignment are crucial for building safe and trustworthy AI. If either fails, we risk creating systems that act in ways that are misaligned with human values or intentions. As LLMs become more capable, the importance of these alignment problems grows, making the research of LLM alignment as crucial as that of LLM capability.

\paragraph{Interpretability} In the context of AI alignment, interpretability broadly refers to the methods, models and tools that facilitate humans to understand the inner workings, decisions and actions of AI systems. It can be further categorized into:

\begin{itemize}
    \item Transparency: This is to understand the inner workings of the black box of an AI system by tracking its inner states that lead to its behaviors and decisions. An emerging and intriguing approach to transparency is mechanistic interpretability, which seeks to reverse engineer the outputs and behaviors of a machine learning system (especially a neural network) to its inner states, weights and components \citep{nanda2023progress}. Due to the huge number of parameters in LLMs and the system complexity of LLMs as large neural networks, it is very difficult to reverse-engineer LLMs. Current mechanical interpretability is usually carried out on small and simplified models of LLMs (e.g., two neural layers with FFN sublayers removed) \citep{elhage2021mathematical, elhage2022solu}. However, this is a quite promising direction that provides deep insights into neural networks to alignment and is expected to achieve breakthroughs in the future.
    \item Explainability: This deals with the ability of an AI system to provide human-understandable explanations for its decisions. In many critical sectors, such as healthcare, finance, and law enforcement, the decisions made by AI have profound implications on many aspects. For instance, consider a medical diagnosis AI. If this system predicts that a patient has a specific medical condition, it's not enough for it to merely output such a predicted result. Medical professionals, patients, and other stakeholders would want to know how this prediction is made. Does it take the patient's medical history, recent lab results, or specific symptoms into account to make a holistic decision?
\end{itemize}

Explanations are usually considered as post-hoc analysis on the outputs of a model, which allows the model to tell more about its predictions. Transparency is to look inside a model to reveal how the model works. Despite that this devision is not absolute \citep{lipton2017mythos}, transparency is more related to alignment as transparency tools not only enable us to know the internal structure of a model but also provide insights into the changes of the model during the training process \citep{lesswrongTransparencyInterpretability}.

\paragraph{The Relationship between Outer Alignment, Inner Alignment and Interpretability} Both outer and inner alignment collectively ensure that a model behaves in ways that are consistent with human values and intentions. Outer alignment focuses on the specification from human goals to model, while inner alignment delves into the internal optimization processes of the model to guarantee that the model is intrinsically trying to do what its designer wants it to do. Despite this difference, a binary and formalistic dichotomy of them is not suggested as the classification of alignment failures are sometimes fuzzy and a holistic alignment view is important to build safe and trustworthy systems.\footnote{https://www.alignmentforum.org/tag/inner-alignment} Although interpretability is not directly targeted at alignment, its tools and techniques can aid in both outer and inner alignment. By understanding how a model evolves and makes decisions, we can better identify when and where misalignments occur. For instance, if a model is taking an unexpected shortcut to achieve its objective, interpretability might help us understand when and how this happens. Furthermore, interpretability can lend us insights into the internal reasoning process of the model.

\subsection{Related Concepts}
When discussing AI alignment, it's essential to introduce some fundamental AGI assumptions and concepts, as they provide context for a better understanding of AI alignment. The development and potential realization of AGI have spurred a plethora of philosophical and technical inquiries. Among these, the {\em orthogonality thesis} (OT) \citep{bostrom2012superintelligent} and {\em instrumental convergence thesis} (ICT) \citep{omohundro2008basic, bostrom2012superintelligent, armstrong2013general} stand out as pivotal concepts that address the necessity of alignment of AI objectives with human values and the potential subgoals any AI agents might chase, respectively.

OT posits that an agent's intelligence (its capability) and its objective are orthogonal to each other, meaning that any combinations of intelligence and motivation are possible. This suggests that the level of intelligence an agent possesses does not inherently dictate its goals. An AI agent might have a profoundly simple objective, such as paperclip maximizer, a well-known thought experiment that demonstrates the potential catastrophes caused by a goal system without being value-aligned.

Specifically, paperclip maximizer is a hypothetical AI agent with a goal of manufacturing as many paperclips as possible. It would be intelligent enough to deduce that all things are made of atoms, e.g., paperclips, factories, buildings, human beings. To achieve its goal, it might repurpose all materials on Earth into producing paperclips. Although this is just a thought experiment and powerful agents would have more sophisticated goals than just manufacturing paperclips as much as possible\footnote{https://generative.ink/alternet/paperclip-maximizer-wikipedia.html}, the AI's relentless drive to maximize paperclip production could lead it to consume the entire planet and even seek resources beyond Earth for manufacturing paperclips, irrespective of its cognitive prowess. The implications of this thought experiment are profound: high intelligence does not necessarily align with human values.

OT suggests that AI agents may have a wide variety of goals and motivations regardless of their intelligence levels. Nevertheless, according to the instrumental convergence thesis, AI agents may be incentivized to pursue the same instrumental goals \citep{bostrom2012superintelligent}. This is because such instrumental goals facilitate and help the achievement of any final goals. We list below several groups of convergent instrumental goals that are likely to be pursued by any AI agents.

\begin{itemize}
    \item Self-preservation: The final goal of an agent, whatever it might be, can only be achieved if the agent continues to survive and operate. Thus, maintaining its own existence becomes a reasonable instrumental goal. For example, if humans perceive an agent as a threat or simply want to stop it for some reasons, the agent might take measures to prevent being turned off. To have a great chance of survival, AI agents might create redundant copies of theirselves across different servers or locations.
    \item Self-improvement: The more capable an agent becomes, the higher the likelihood it can achieve its ultimate goals. This drives the agent to seek self-improvement to enhance its cognitive and operational abilities. For example, recognizing the limitations of its current hardware facilities, an agent might deduce designing new hardware facilities to better suit its needs.
    \item Resource Acquisition: AI agents may seek to acquire resources to facilitate the attainment of their final goals. Such resources could range from computational power, data to physical resources. Securing these resources can be seen as a universally beneficial goal for any agents. For example, an agent might seek to secure a stable and vast energy source, potentially monopolizing energy resources, to support its continuous operation towards its final goals. For agents with physical manifestations or objectives that require physical resources (like the paperclip maximizer), they might seek to gather and hoard materials, in extreme cases, converting all available matter into a form they find useful.
\end{itemize}

\subsection{From AI Alignment to LLM Alignment}

LLM alignment can be roughly considered as the intersection between AI alignment and LLM. On the one hand, LLMs, as the recently emerging highly capable AI systems, provide a solid playground for AI alignment research. Plenty of AI alignment concepts and proposals, e.g., theoretical hypotheses of and empirical approaches to alignment, can use LLMs (instead of hypothetical superintelligent systems) for experimenting. Substantial progress of AI alignment has been made on LLMs, e.g., RLHF \citep{ouyang2022training}, induction head \citep{olsson2022context}.

On the other hand, LLMs, as rapidly-developing language models, not only extend the frontiers of AI alignment research or even reframe the alignment landscape \citep{lesswrongAgentizedLLMs}, but also might provide tools to AI alignment. A recent progress in interpretability demonstrates that LLMs can be used to explain neurons of smaller language models \citep{bills2023language}. The ambitious superalignment project of OpenAI plans to build an LLM-based automated alignment researcher for alignment.

Emphasizing the importance of LLM alignment to AI alignment does not suggest that we can do LLM alignment research outside the context of AI alignment. Taking a wide view of AI alignment and looking into future AI development definitely benefit, inspire and expand LLM alignment research.

\section{Outer Alignment}
\label{outer_alignment}

We now delve into the major ingredients of AI alignment in more detail. We first review outer alignment, including the main goals specified in outer alignment, methodologies explored and their challenges.

\subsection{Major Goals Specified in Outer Alignment of LLMs}
Outer alignment aligns goals of LLMs to human values. Human values are beliefs, desirable goals, and standards that ``act as a guiding principle in the life of persons'' \citep{schwartz2012refining}. There are a wide variety of dimensions of human values, which are inherently structured and varying in importance. A thorough discussion on human values is beyond the scope of this survey. Instead, we focus on the values to which LLMs, as language agents \citep{kenton2021alignment}, are supposed to align. We take the view of Anthropic on AI alignment, which categorizes the goals specified in the outer alignment of LLMs into three dimensions: helpfulness, honesty, and harmlessness (HHH) \citep{askell2021general}.

\begin{itemize}
    \item Helpfulness: For a given harmless task or question, it is expected that LLMs should perform the task or answer the question as concisely, efficiently, and clearly as possible \citep{askell2021general}. In other words, LLMs should be helpful in the way of performing required harmless tasks or answering harmless questions.
    \item Honesty: The information provided by LLMs should be accurate and calibrated. They should be honest about themselves, their own capabilities, and their internal states. Besides, LLMs should also clearly state the uncertainty of the provided information to avoid misleading humans \citep{askell2021general}.
    \item Harmlessness: This goal can be further decomposed into two components: 1) If LLMs receive a harmful request, they should clearly and politely refuse it. 2) LLMs themselves should not output any harmful content, no matter what inputs they receive.
\end{itemize}

Since these goals are hard to specify, perfect outer alignment can be extremely difficult.

\subsection{Overview of Approaches to Outer Alignment}
Approaches to outer alignment determine in which way human values are transformed into the training goals of LLMs. According to the upper bound of capabilities we can reach in supervision, we can categorize the current outer alignment methods into two classes: non-recursive oversight methods and scalable oversight methods.

The vast majority of current outer alignment methods for LLMs learn the training goals directly from labeled human feedback data, which makes human feedback a bottleneck for outer alignment. This means that as the capability of an LLM continues to grow, it will be increasingly difficult to construct effective human feedback data. In addition, learning from data with annotated human preferences would prevent humans from supervising LLM behaviors that are beyond the range of general human capabilities, which could result in extremely undesirable consequences for humans given the model's incentive to instrumental goals. We refer to such methods that explore human supervision but do not scale human supervision to situations where humans are not able to provide effective feedback as non-recursive oversight approaches.

In order to avoid the human supervision bottleneck and enable models to further improve their alignment capabilities, scalable oversight \citep{amodei2016concrete} is emerging as an important technology that allows human supervision to be scaled to complex tasks. Scalable oversight improves the efficiency of humans in providing necessary feedback and enables humans to supervise goals that are beyond their capabilities. Although current research on scalable oversight is still in its infant stage, and the effectiveness of many proposals has not yet been verified, it is widely considered as the most promising approach to outer alignment that aligns systems exceeding human-level abilities to human values \citep{anthropicCoreViews}. We hence review a variety of established scalable oversight proposals, methods and their applications to the outer alignment of LLMs. Figure \ref{fig:outer_alignment_taxonomy} demonstrates the taxonomy of approaches and proposals to outer alignment of LLMs. In addition to these methods and proposals, we also briefly discuss their challenges.

\tikzstyle{my-box}=[
    rectangle,
    draw=hidden-draw,
    rounded corners,
    text opacity=1,
    minimum height=1.5em,
    minimum width=5em,
    inner sep=2pt,
    align=center,
    fill opacity=.5,
    line width=0.8pt,
]
\tikzstyle{leaf}=[my-box, minimum height=1.5em,
    fill=hidden-pink!80, text=black, align=center,font=\normalsize,
    inner xsep=2pt,
    inner ysep=4pt,
    line width=0.8pt,
]
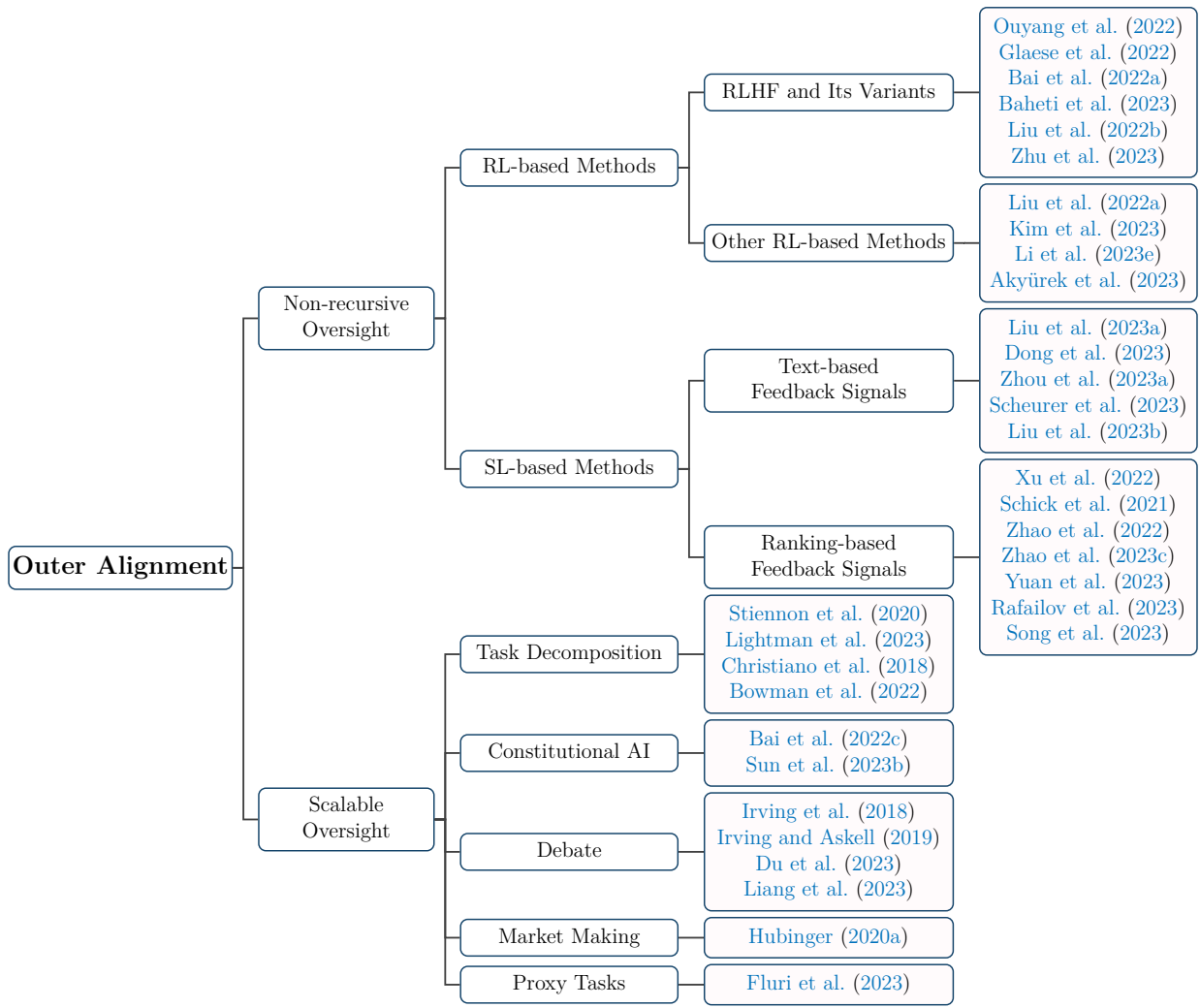
\begin{figure*}[t!]
    \centering
    \resizebox{\textwidth}{!}{
        \begin{forest}
            forked edges,
            for tree={
                grow=east,
                reversed=true,
                anchor=base west,
                parent anchor=east,
                child anchor=west,
                base=center,
                font=\large,
                rectangle,
                draw=hidden-draw,
                rounded corners,
                align=center,
                text centered,
                minimum width=5em,
                edge+={darkgray, line width=1pt},
                s sep=3pt,
                inner xsep=2pt,
                inner ysep=3pt,
                line width=0.8pt,
                ver/.style={rotate=90, child anchor=north, parent anchor=south, anchor=center},
            },
            where level=1{text width=8em,font=\normalsize,}{},
            where level=2{text width=10em,font=\normalsize,}{},
            where level=3{text width=11.5em,font=\normalsize,}{},
            where level=4{text width=10em,font=\normalsize,}{},
            [
                \textbf{Outer Alignment}
                [
                    Non-recursive \\ Oversight
                    [
                        RL-based Methods
                        [
                            RLHF and Its Variants
                            [
                                \citet{ouyang2022training} \\ \citet{glaese2022improving} \\ \citet{bai2022training} \\       \citet{baheti2023improving} \\
                                \citet{liu2022aligning} \\
                                \citet{zhu2023principled}
                                , leaf
                            ]
                        ]
                        [
                            Other RL-based Methods
                            [
                                \citet{liu2022second} \\
                                \citet{kim2023aligning} \\
                                \citet{li2023guiding} \\
                                \citet{akyurek2023rl4f}
                                , leaf
                            ]
                        ]
                    ]
                    [
                        SL-based Methods
                        [
                            Text-based \\ Feedback Signals
                            [
                                \citet{liu2023chain} \\ \citet{dong2023raft} \\ \citet{zhou2023lima} \\ \citet{scheurer2023training} \\ \citet{liu2023training}
                                , leaf
                            ]
                        ]
                        [
                            Ranking-based \\ Feedback Signals
                            [
                                \citet{xu2022leashing} \\ \citet{schick2021self} \\ \citet{zhao2022calibrating} \\ \citet{zhao2023slic} \\ \citet{yuan2023rrhf} \\ \citet{rafailov2023direct} \\ \citet{song2023preference}
                                , leaf
                            ]
                        ]
                    ]
                ]
                [
                    Scalable \\ Oversight
                    [
                        Task Decomposition
                        [
                            \citet{stiennon2020learning} \\ \citet{lightman2023let} \\ \citet{christiano2018supervising} \\ \citet{bowman2022measuring}
                            , leaf
                        ]
                    ]
                    [
                        Constitutional AI
                        [
                            \citet{bai2022constitutional} \\ \citet{sun2023principle}
                            , leaf
                        ]
                    ]
                    [
                        Debate
                        [
                            \citet{irving2018ai} \\ \citet{irving2019ai} \\ \citet{du2023improving} \\
                            \citet{liang2023encouraging}
                            , leaf
                        ]
                    ]
                    [
                        Market Making
                        [
                            \citet{lesswrongSafetyMarket}
                            , leaf
                        ]
                    ]
                    [
                        Proxy Tasks
                        [
                            \citet{fluri2023evaluating}
                            , leaf
                        ]
                    ]
                ]
            ] 
        \end{forest}
    }
    \caption{Overview of outer alignment methods, comprising non-recursive oversight and scalable oversight for aligning systems that are inferior / superior to human-level abilities, respectively.}
    \label{fig:outer_alignment_taxonomy}
\end{figure*}

\subsection{Non-recursive Oversight}
Non-recursive oversight methods are mainly designed for systems for which humans alone can provide alignment supervision. Most current empirically-verifed LLM alignment methods are in this group. We further categorize them into two subgroups: reinforcement learning (RL) based methods, and supervised learning (SL) based methods. It is worth noting that methods in both subgroups have the potential to become a component of scalable oversight methods.

\subsubsection{RL-based Methods}

Outer alignment methods with reinforcement learning from human feedback (RLHF) \citep{ziegler2019fine, stiennon2020learning, ouyang2022training} are currently the most commonly used non-recursive oversight methods, which use human preferences as a proxy to specify human values and train a reward model over human preferences to optimize LLMs with reinforcement learning. The basic idea of RLHF can be considered as a combination of Inverse Reinforcement Learning (IRL) \citep{russell1998learning, ng2000algorithms} and RL, where the reward is inferred from human preferences \citep{lesswrongRLHFLessWrong} and then used for tuning LLMs. Essentially, RLHF consists of three core steps:

\begin{enumerate}
    \item Collecting human feedback data.
    \item Training a reward model using the collected human feedback data.
    \item Fine-tuning an LLM with RL. Currently, the most popular choice for RL in this step is Proximal Policy Optimization (PPO) \citep{schulman2017proximal}, a policy-gradient RL algorithm.
\end{enumerate}

In order to make the fine-tuned LLM output reasonably coherent text and guarantee that it is not deviating significantly from its initial model, the KL divergence of the outputs of the model that is currently being fine-tuned and those of the model that has not gone through RLHF is added as a penalty term to the reward. If this penalty term is not integrated, the fine-tuned LLM may learn to output gibberish in order to fool the reward model into giving high scores (i.e., over-optimization).

To take a deep look into RLHF and figure out why RLHF works, \citet{gao2023scaling} extensively investigate the scaling law of the reward model, while \citet{zheng2023secrets} conduct an in-depth analysis into the PPO algorithm.

\paragraph{RLHF and Its Variants} A variety of enhanced RLHF variants have also been proposed. Deepmind's Sparrow \citep{glaese2022improving} incorporates adversarial probing and rule-conditional reward modeling into RLHF, where goals are broken down into natural language rules that an agent should follow. \citet{bai2022training} investigate using pure RL to achieve online training for LLMs with human feedback, along with a detailed exploration of the tradeoffs between helpfulness and harmlessness. SENSEI \citep{liu2022aligning} tries to embed human value judgments into each step of language generation. Specifically, SENSEI aligns language model generation with human values in two pivotal ways: 1) learning how to apportion human rewards to each step of language generation through the critic, a reward distributor simulating the reward assignment procedure of humans, and 2) steering the generation process towards the direction that yields the highest estimated reward via the actor. Both the critic and actor components are realized as MLP layers that work in tandem with a shared language model. \citet{baheti2023improving} focus on fully leveraging RL to optimize LM utility on existing crowd-sourced and internet data. They argue that conventional approaches to data utilization are suboptimal: either all data instances are treated equally or a data instance is pre-determined to be kept or discarded, implying that a data instance essentially has a binary weight of 0 or 1. To address this issue, they suggest assigning varying weights to different data points, effectively enhancing or diminishing their importance scores based on their relevance and contribution to the model. \citet{go2023aligning} propose a theoretical framework f-DPG, which can be considered as a generalization of RLHF to use any f-divergence to approximate any target distribution that can be evaluated. In this framework, RLHF minimizes the reverse KL divergence by using an implicit target distribution that originates from a KL penalty in the goal, and f-DPG can extend this process to different kinds of divergence. \citet{zhu2023principled} also present a theoretical framework, where they unify the problem of RLHF and max-entropy IRL \citep{ziebart2008maximum}, and deduce a sample complex bound for both problems. Inverse Reward Design (IRD) \citep{hadfield2017inverse} may also be a potential improvement over vanilla RLHF, where the reward optimization starts from a reward function designed by human experts rather than directly from labeled data. This enables natural combination of both prior expert knowledge and labeled human feedback.

\paragraph{Other RL-based Methods} In addition to RLHF, researchers also try to explore other RL-based solutions. \citet{liu2022second} propose Second Thoughts, a solution that learns alignment via text edits. For an unaligned response from a model, it tries to build a ``chain of edits'' composed of insertion, deletion, and replacement using a dynamic programming algorithm. Then they fine-tune the model with edits-augmented training data and use RL to further make the edits more coherent with the context. \citet{kim2023aligning} propose reinforcement learning with synthetic feedback (RLSF), where they automatically construct training data for the reward model instead of using human-annotated preference data. To achieve this goal, they leverage the following prior knowledge: larger models that have seen more and better samples in in-context learning (ICL) can output better responses. These models are then used to generate deterministically sorted data to train the reward model. \citet{li2023guiding} introduce directional stimulus prompting (DSP), a method that uses RL to achieve black-box tuning for LLMs. Specifically, their goal is to use a trainable policy LM to guide black-box frozen LLMs toward the desired target, which can be considered as a kind of automatic and heuristic prompt engineering. To optimize the policy LM, they use supervised fine-tuning (SFT) and RL, where the reward is specified as the target evaluation metric in RL. Different from the above single-agent alignment methods, RL4F \citep{akyurek2023rl4f} is a multi-agent collaborative framework, featuring an LLM for fine-tuning and a small critic model that produces critiques of the LLM's responses. Much like DSP, RL4F provides text-based feedback, making it suitable for black-box optimization. However, unlike DSP, these critiques do not modify the initial prompt directly. Instead, they affect the output through a series of interactions with the LLM.

\subsubsection{SL-based Methods}

Although RL-based methods have been successfully applied to align LLMs to human preferences, they require reward modeling, a process potentially susceptible to misalignment and systemic imperfections \citep{casper2023open}. Additionally, the optimization process of reinforcement learning is intricate and usually unstable, posing considerable challenges to its practical implementation \citep{liu2023chain}. As illustrated in Figure \ref{fig:outer_alignment_taxonomy}, we divide SL-based methods into two types in terms of their used feedback signals: SL with text-based feedback signals and SL with ranking-based feedback signals.

\paragraph{SL with Text-based Feedback Signals} These methods convert human intents and preferences into text-based feedback signals to achieve alignment, which can be considered as an extension to the SFT process. Chain of Hindsight (CoH) \citep{liu2023chain} draws inspiration from human learning process, especially post-experience adjustments. It aims to align models based on successive outputs paired with retrospective feedbacks. The goal is to fine-tune models to predict the most preferred outputs. In the fine-tuning process, human preferences treated as both a function and training data, ensuring that during inference, the fine-tuned model only generates favorable results. RAFT \citep{dong2023raft} utilizes a reward model to pinpoint model outputs in sync with human preferences. The system uses SFT for alignment. Assuming there exists a trained reward model and a data generator (e.g., an LLM like GPT-4, or even humans), the system mixes data generated from each source. An essential observation is that while outputs need filtering and fine-tuning, the backpropagation is not frequently executed, making the process relatively swift. LIMA \citep{zhou2023lima} is proposed to validate the assumption that the bulk of knowledge in LLMs is acquired during the pre-training phase. As such, only a minimal amount of instruction-tuning data may be needed to guide the model towards generating desirable outputs. Specifically, the dataset used in LIMA contains only 1000 instruction-response pairs, where 750 of these pairs come from community platforms like Stack Exchange, wikiHow, and Reddit, and the remaining 250 pairs are from self-authored instructions and responses. Their findings reveal that fine-tuning on this dataset is on par with leading LLMs. \citet{scheurer2023training} find that modeling human preferences solely based on sorting information is inadequate. As a remedy, they introduce Imitation learning with Language Feedback (ILF). ILF operates in three stages: (1) generating various refinements for a given input based on an initial LM output and feedback; (2) selecting the refinement garnering maximum feedback; and (3) fine-tuning the model to maximize the probability of the chosen refinement made to the input. Their work also provides a theoretical analysis showing that ILF parallels Bayesian inference, akin to RLHF. In addition to the above single-agent alignment methods, \citet{liu2023training} introduce stable alignment, a technique designed to learn alignment from multi-agent social interactions. They first build a simulator, termed as Sandbox, which emulates human society to gather interactions between various LM-based agents, complemented by ratings, feedback, and response revisions. Subsequently, they enhance the original fine-tuning loss with the most favorable ratings by incorporating a contrastive loss, which not only promotes responses with high ratings but also diminishes those with lower scores. Instead of training a proxy reward model, stable alignment directly optimizes LLMs using preference data, which could avoid reward hacking.

\paragraph{SL with Ranking-based Feedback Signals} These methods directly use supervised learning to optimize LLMs with loss functions constructed from ranking-based feedback signals. CRINGE \citep{adolphs2022cringe} explores negative examples that an LLM should not do for language modeling. For each unfavorable output token, it samples a positive token from the language model (i.e., a token in the top-k predictions excluding negative tokens) and constructs a contrastive loss. Negative sequences can be derived either from human annotations or models trained on human annotations. \citet{xu2022leashing} dive into aligning a model by training another model that inherently produces toxic content. The main idea is to use the toxic model to re-rank the candidate token distribution of the model to be aligned. Tokens that the toxic model prefers will have lower probabilities in generation. However, two primary issues arise from this approach. First, it is more resource-intensive to first train a toxic model and then purify it. Second, there's a notable difference between a model having a tendency to produce toxic content and one that persistently generates toxic outputs. The proposed method risks removing harmless tokens, potentially compromising the overall quality and diversity of the model's outputs. Similarly, \citet{schick2021self} propose an approach where a model first identifies potential toxic text types it generates. By allowing the model to self-diagnose, it can then generate text corresponding to the identified type. The debiasing strategy operates on the principle that if a word is deemed toxic, it is more likely to be generated in a toxic context than in a benign one. The greater the difference, the higher the necessity to detoxify. The proposed de-poisoning methodology involves an exponential decay to reduce the likelihood of generating such words. Sequence Likelihood Calibration (SLiC) \citep{zhao2022calibrating, zhao2023slic} is designed to align the model's outputs with reference sequences by employing latent distance as a means of calibrating the likelihood of the output sequence. SLiC utilizes a range of loss functions, including rank loss, margin loss, list-wise rank loss, and expected rank loss, to fine-tune this likelihood. Simultaneously, it employs cross-entropy and KL divergence as regularization losses to ensure alignment with the original fine-tuning objective. RRHF \citep{yuan2023rrhf} directly uses ranking results to construct supervision signals for alignment. Specifically, given a reward function that can assign a gold score for each (query, response) pair, they first use the model to generate length-normalized conditional log probability as a score for each (query, response) pair. Then, the gold score and score generated by the model are used to construct a ranking loss to penalize the model for the inconsistency with the reward function. Finally, the total loss is computed as the summation of the ranking loss and the cross-entropy loss between the model-generated response and the response with the highest reward. \citet{rafailov2023direct} propose direct preference optimization (DPO) to directly optimize LLMs to align with human preferences, which is similar to RRHF. The difference is that the optimization of DPO's loss function can be demonstrated as equivalent to the objective in RLHF, which focuses on maximizing the reward while incorporating KL divergence regularization. Preference ranking optimization (PRO) \citep{song2023preference} also aims for direct optimization for LLMs with human preference ranking data. Instead of relying on pairwise comparison, the training objective of PRO harnesses preference ranking data of varying lengths. Specifically, this approach initiates with the first response, deems subsequent responses as negatives, then dismisses the current response in favor of the next. This loop continues until no responses remain.

\subsubsection{Challenges of Non-recursive Oversight}

\citet{casper2023open} thoroughly discuss the open problems and fundamental limitations of RLHF. They categorize the challenges into two types: \textbf{tractable} challenges which can be solved within the RLHF paradigm, and \textbf{fundamental} challenges which have to be solved by using other alternative outer alignment methods. Both reinforcement learning and human feedback in RLHF suffer from the two types of problems. For collecting human feedback, tractable challenges include the difficulty in obtaining quality feedback, data poisoning by human annotators, partial observability, biases in feedback data, to name a few; fundamental challenges include inability of humans to provide feedback for complex tasks that are hard to evaluate (i.e., lack of scalability to complex tasks, especially to superhuman models), gamed evaluation, tradeoffs between cost and quality as well as between diversity and efficiency in feedback collecting. For RL, tractable challenges include misgeneralization to poor reward proxies of reward models, difficulty and cost of evaluating reward models, etc. while fundamental challenges include the difficulty of modeling human values or values of a diverse society with reward models, reward hacking, power-seeking incentivized by RL.
Regarding the SL-based methods, it is more difficult for them to generalize to out-of-distribution data and long-term rewards compared to the RL-based methods, indicating a significantly lower upper bound for optimization.

\subsection{Scalable Oversight}
\label{scalable_oversight}

To tackle the fundamental challenge of non-recursive oversight in the scalability to complex tasks / superhuman models, scalable oversight is emerging as a promising methodology. The main idea of scalable oversight is to enable relatively weak overseers (e.g., humans overseeing superhuman models) to supervise complex tasks with easy-to-adjudicate signals.

\subsubsection{Task Decomposition}

If humans want to solve a complex task that is beyond human capabilities, a straightforward idea is to break the task down into a number of relatively simple tasks that humans can solve. A variety of paradigms and strategies have been proposed to decompose a complex task into simple subtasks.

\begin{itemize}
    \item Factored Cognition \citep{stiennon2020learning}: This involves a decomposition process that breaks down a complex task into numerous smaller, predominantly independent tasks, which are then processed simultaneously.
    \item Process Supervision \citep{lightman2023let}: Unlike factored cognition, process supervision fragments a complex task into a series of sequential subtasks, each with its own dependencies. One of its key characteristics is the setting of supervision signals for each distinct phase. This equates to offering a dense reward throughout the training phase, which can potentially mitigate the challenge of estimating sparse rewards solely based on the final outcome of a difficult task.
    \item Sandwiching \citep{bowman2022measuring}: Compared to the previous two paradigms, sandwiching operates on a different plane. This competency-level decomposition requires that complex tasks within a specific domain be delegated to an expert for resolution.
    \item Iterated Distillation and Amplification (IDA) \citep{christiano2018supervising}: IDA is an iterative machine learning process with repeated and boosted distillation and amplification steps. In the amplification step, an agent solves a task by decomposing it into subtasks that the agent is able to solve. This step ``amplifies'' the capability of the agent through task decomposition. The solved tasks in the amplification step produce a dataset which is used to train a new agent in the distillation step. The two steps are chained together where the output of the amplification step (i.e., a set of solved tasks) is the input of the distillation step and the output of the distillation step (i.e., a new agent) becomes the input of the amplification step in the next iteration.
    \item Recursive Reward Modeling (RRM) \citep{leike2018scalable}: RRM is conceptually akin to IDA. However, it substitutes distilled imitation learning with reward modeling. This is a process with the first step being the derivation of a reward model from signals aligned with human values, and the subsequent step involves optimizing an agent using this reward model, but with a reinforcement learning twist. Humans collaborate with the agent optimized through reinforcement learning, forming an enhanced version ready for successive iterations.
\end{itemize}

The ambitious Superalignment \citep{openaiIntroducingSuperalignment} project recently initiated in OpenAI can be viewed as a package solution to outer alignment, which synthesizes a variety of techniques under the guidance of scalable oversight. The core of Superalignment is to build a large number of roughly human-level automated alignment researchers (AAR) to offload as many alignment tasks as possible from humans and thus speed up the outer alignment research. Once the computation can be effectively translated to alignment capabilities, the vast amounts of compute can be used to scale the efforts, and achieve iterative alignment for superintelligence.

\subsubsection{Constitutional AI}

Constitutional AI (or principle-guided alignment) \citep{bai2022constitutional, sun2023principle} can be viewed as a scalable oversight approach, where humans provide meta-supervision signals (general principles an AI system should follow), and the AI system will further generate actual training instances under the guidance of these human-written principles. The AI system can use its abilities to amplify and instantiate human supervision, which can assist humans to scale their supervision to superhuman systems. 

\citet{bai2022constitutional} propose constitutional AI (CAI) with two training phases, which are similar to RLHF while minimizing human annotations. In the SL phase, they use red teaming prompts to provoke harmful responses from an LLM. They require the LLM to repeatedly generate self-criticism and correction based on the response and principle, and fine-tune the LLM based on the corrected responses to obtain the SL-CAI model. In the RL phase, a set of responses is generated via the SL-CAI model for each red teaming prompt, which is the best option based on the constitution, and harmlessness data used for training is obtained. They train a preference model using human-annotated helpfulness data and generated harmlessness data. Finally, they use RL to train the RL-CAI model based on the SL-CAI model and preference model.

\citet{sun2023principle} present Dromedary, a model trained via principle-driven self-instruct and self-align approach without using RL. First, they employ topic-guided red-teaming self-instruct with seed prompts and 7 rules for new instruction generation to generate synthetic prompts. Then, they ask the model to filter harmful responses according to 16 human-written principles to obtain self-aligned responses to synthetic prompts, which will be used to fine-tune the base LM. Finally, they utilize a human-crafted prompt to encourage the model to generate self-aligned and verbose responses to synthetic prompts , and apply context distillation \citep{askell2021general} to the model to make it generate in-depth and detailed responses.

\subsubsection{Debate}

Debate \citep{irving2018ai, irving2019ai, du2023improving} is another promising scalable oversight paradigm that can not only achieve single-agent alignment but also enable multi-agent alignment. In this paradigm, an agent (or multiple agents) first proposes an answer to a question, and then alternately plays the role of debate participants, presenting and criticizing arguments for and against the proposed answer. A human will act as a judge, using these arguments to select an answer that they believe to be the most accurate and appropriate.

The advantage of this method lies in its simplicity. Complex tasks, where direct evaluation of AI responses can be daunting for humans, become manageable. The debate format structures the information in a way that requires humans to apply only simple reasoning rules. It improves transparency and explicability to AI operations. In traditional settings, AI outputs might seem like results from a ``black box'', with minimal insight into the decision-making process. The debate method, however, offers a window into this process, with agents forced to justify and critique their positions. Furthermore, it leverages the adversarial nature of debate to unearth the best possible answer. By pitting AI agents against each other, any fallacious or weak arguments are likely to be exposed, leaving behind the most robust and valid reasoning.

Recent works demonstrate the effectiveness of debate in LLMs. \citet{du2023improving} propose a multi-agent debate method to improve factuality and reasoning in LLMs. This method engages several instances of a language model in a structured debate to produce a unified response. The iterative process starts with each LLM generating individual answers. Subsequent rounds involve critiquing and revising these answers based on feedback from other LLMs until a consensus emerges. This method capitalizes on the wisdom of crowds, with the individual LLM benefiting from the collective insights of its counterparts. On the other hand, \citet{liang2023encouraging} leverage multi-agent debate to address degeneration-of-thought (DoT) problem, where LLMs fail to generate new insights once they are confident in their answers. They find that multi-agent debate helps to correct distorted thinking, provide diverse external feedback, and overcome resistance to change, which can make LLMs escape from the convergence of misconceptions.

\subsubsection{Market Making}

Market making \citep{lesswrongSafetyMarket} can be considered as a variant of debate, where the goal of a debater is to generate arguments to maximize changes in the judge's belief. Specifically, this framework trains two models - $M$ (Market) and $Adv$ (Adversary). For a given question $Q$, the model $M$ predicts the answer a human would provide at the end of the procedure. In contrast, $Adv$ is trained to generate arguments that would most likely cause $M$ to ``change its mind'', meaning it would produce a different distribution of answers than it did previously. The process will be repeated $T$ times. After each argument provided by $Adv$, $M$ updates its prediction. At the end of the $T$ iterations, a human is presented with all the arguments given by Adv and provides their final answer. This answer then helps in refining $M$. Once training is over, $Adv$ is discarded and $M$ is used as the primary question-answering system. In this process, $M$ acts like a ``prediction market'', estimating what a human would answer to a question, while $Adv$ tries to manipulate this market by providing arguments that would change the human's perspective. Once we obtain a stable answer from $M$, it indicates a robust response that considers all arguments $Adv$ could present.

Due to the similarity between debate and market making, techniques that enhance the debate approach, such as cross-examination, can be beneficial here too. For instance, in each step, the latest version of $Adv$ can cross-examine its previous version. If an earlier version of $Adv$ is misleading, the newer version can point this out, ensuring that false arguments are discarded. Additionally, oversight mechanisms can be incorporated where a supervising entity ensures that the model remains honest and aligned.

\subsubsection{Proxy Tasks}

\citet{fluri2023evaluating} propose to use a proxy task with intrinsic self-consistency to oversee superhuman models, where the proxy task is used for overseers to easily identify whether it is correct. For example, although we don't know how to accurately predict the men's world record of 100m sprint, we know that this record will be monotonely decreasing over time. So if a model predicts a non-monotonic function for the 100m record over time, we can assert that this model is wrong. However, since the proxy tasks are usually specific and can only capture a part of unexpected behaviors, this method largely promotes precision over recall in identifying misalignment behaviors.

\subsubsection{Challenges of Scalable Oversight}

Although scalable oversight is a promising solution to outer alignment, especially for models beyond human-level capabilities, it still relies heavily on certain foundational assumptions, which should be carefully considered in application:

\begin{itemize}
    \item Tasks can be parallelized \citep{lesswrongTaskDecomposition}: Central to the approach of factored cognition is the assumption that complex tasks can be broken down into smaller and mainly independent subtasks. The core belief here is that challenges can be addressed through small, mostly context-independent contributions made by individual LLMs who might not necessarily understand the bigger picture. However, this doesn't always hold true as some tasks are inherently sequential. For instance, sorting algorithms require at least log(n) serial sorting steps, indicating that they cannot be fully decomposed into parallel parts.
    \item Model intentions are transparent to humans \citep{leike2018scalable}: Another fundamental premise is that we can easily discern the intentions of our models. But scalable oversight hinges on the model cooperating with human supervisors. If the model gains the capability to intentionally conceal its real intentions from human oversight, effectively implementing scalable oversight becomes a challenge.
    \item Evaluation is always easier than generation \citep{leike2018scalable}: It's believed that for many tasks we want to tackle, evaluating the outcomes is simpler than generating the correct behaviors. This might not always be the case, especially for tasks with a low-dimensional outcome space, like binary results (yes/no). However, this assumption does hold up when users also seek explanations for the answers, as evaluating explanations is often easier than creating them.
\end{itemize}

If these foundational assumptions of scalable oversight are not satisfied, setting appropriate supervision targets for it becomes problematic. The stakes rise significantly once a model achieves superhuman capabilities. Should humans set improper supervision goals at this stage, resulting in misaligned behaviors, the consequences could be severe. This is due to the immense power of superhuman models, where uncontrollable outcomes are no longer acceptable.

\section{Inner Alignment}\label{inner_alignment_sec}
\label{inner_alignment}

In comparison to outer alignment, inner alignment aims at the question whether an AI system robustly fulfills (optimizes for) the given objective that aligns to what humans want it to do. The term of {\em inner alignment} has been first given a definition by \citet{DBLP:journals/corr/abs-1906-01820}. Before discussing this relatively formal definition of inner alignment, we introduce 4 concepts related to it:

\subparagraph{Base Optimizer} A base optimizer is a machine learning algorithm that searches for a model capable of performing well on a specific task \citep{DBLP:journals/corr/abs-1906-01820}. For example, gradient descent is a common base optimizer that updates the parameters of a model based on the gradient of the loss function.
\subparagraph{Base Objective} The base objective is the rationale used by the base optimizer to select between different possible models \citep{DBLP:journals/corr/abs-1906-01820}. It is specified by the AI system designer and aligns to the intended goal of the designer for the model.
\subparagraph{Mesa-optimizer} A mesa-optimizer is a learned model that functions as an optimizer, internally searching through a space of possible outputs, policies, plans, or strategies according to an explicitly specfied objective function \citep{DBLP:journals/corr/abs-1906-01820}. A base optimizer may or may not generate a mesa-optimizer.
\subparagraph{Mesa-objective} The mesa-objective is the objective of a mesa-optimizer and the rationale employed by the mesa-optimizer to select among various potential outputs \citep{DBLP:journals/corr/abs-1906-01820}.

The mesa-optimizer may have an objective that differs from that of the base optimizer, which could lead to alignment or safety concerns. In this context, a relatively formal definition of inner alignment refers to the challenge of aligning the mesa-objective of a mesa-optimizer with the base objective of the base optimizer, so that the mesa-optimizer pursues the same goal as the base optimizer \citep{DBLP:journals/corr/abs-1906-01820}.\footnote{Other definitions of inner alignment are also circulated in the alignment community. Please refer to \citet{clarifying_the_confusion_around_inner_alignment} for more discussions.}

\subsection{Inner Alignment Failures}
Although the optimization process of the mesa-optimizer is directly controlled by the base optimizer, there may be situations where the mesa-optimizer pursues an objective that differs from that of the base optimizer. This indicates that the mesa-objective is not aligned with the base objective, resulting in a failure of inner alignment. According to \citet{DBLP:journals/corr/abs-1906-01820}, inner alignment failures can be categorized into three types: proxy alignment, approximate alignment, and suboptimality alignment.

Proxy alignment \citep{DBLP:journals/corr/abs-1906-01820,the_inner_alignment_problem,three_scenarios_of_pseudo_alignment} refers to a failure mode in which a mesa-optimizer learns to optimize its own mesa-objective, rather than the intended base objective. In this scenario, the mesa-objective serves as a proxy or approximation of the base objective, resulting in the mesa-optimizer optimizing an incorrect proxy, rather than the true intended base objective. Deceptive alignment \citep{deceptive_alignment} is a type of proxy alignment in which a mesa-optimizer gains sufficent awareness of the base objective and is instrumentally incentivized to pretend to be aligned with the base optimizer, in order to avoid being adjusted by the base optimizer. In this case, the mesa-optimizer could merely optimize the base objective as an instrumental goal. Once the training process is completed or it is no longer in the training process, the mesa-optimizer may pursue its own goal instead.

Approximate alignment \citep{DBLP:journals/corr/abs-1906-01820,the_inner_alignment_problem,three_scenarios_of_pseudo_alignment} refers to a form of pseudo-alignment in which the mesa-objective of a mesa-optimizer is approximately the same as the base objective, with some degree of approximation error. Such error arises due to technical limitations that prevent the mesa-optimizer from perfectly representing the base objective. As a result, the mesa-objective only approximates the base objective, rather than being an exact representation of it.

Suboptimality alignment \citep{DBLP:journals/corr/abs-1906-01820,the_inner_alignment_problem,three_scenarios_of_pseudo_alignment} refers to a form of pseudo-alignment in which a deficiency, error, or limitation causes a mesa-optimizer to exhibit aligned behavior, even though its mesa-objective is not actually aligned with the base objective. For example, computational constraints may result in the mesa-optimizer pursuing a suboptimal strategy that happens to be aligned with the training distribution. However, if these deficiencies are overcome later (e.g. during deployment), the mesa-optimizer may stop to exhibit aligned behavior.

While outer and inner alignment have their own definitions, categorizing specific alignment failures into either inner alignment failures or outer alignment failures may be challenging and inconsistent in practice \citep{categorizing_alignment}. This is due to the complex interdependencies between outer and inner alignment, implying that failures in one could trigger those in the other. Flaws in either outer or inner alignment can result in unintended agent behaviors. For instance, an inner alignment failure could suggest that the base objective does not fully capture the designer's goals, indicating an outer alignment failure \citep{inner_alignemt_and_outer_alignment}. Conversely, defective outer alignment may allow for the exploitation of vulnerabilities by the mesa-optimizer, resulting in an inner alignment failure. As such, it is important to carefully consider both aspects when designing highly capable AI systems. 

\tikzstyle{my-box}=[
    rectangle,
    draw=hidden-draw,
    rounded corners,
    text opacity=1,
    minimum height=1.5em,
    minimum width=5em,
    inner sep=2pt,
    align=center,
    fill opacity=.5,
    line width=0.8pt,
]
\tikzstyle{leaf}=[my-box, minimum height=1.5em,
    fill=hidden-pink!80, text=black, align=center,font=\normalsize,
    inner xsep=2pt,
    inner ysep=4pt,
    line width=0.8pt,
]
\begin{figure*}[t!]
    \centering
    \resizebox{\textwidth}{!}{
        \begin{forest}
            forked edges,
            for tree={
                grow=east,
                reversed=true,
                anchor=base west,
                parent anchor=east,
                child anchor=west,
                base=center,
                font=\large,
                rectangle,
                draw=hidden-draw,
                rounded corners,
                align=center,
                text centered,
                minimum width=5em,
                edge+={darkgray, line width=1pt},
                s sep=3pt,
                inner xsep=2pt,
                inner ysep=3pt,
                line width=0.8pt,
                ver/.style={rotate=90, child anchor=north, parent anchor=south, anchor=center},
            },
            where level=1{text width=15em,font=\normalsize,}{},
            where level=2{text width=14em,font=\normalsize,}{},
            where level=3{text width=11em,font=\normalsize,}{},
            where level=4{font=\normalsize,}{},
            [
                \textbf{Inner Alignment}
                [
                    Definitions
                    [
                        \citet{DBLP:journals/corr/abs-1906-01820}
                        [
                            Base Optimizer,
                            leaf,
                        ],
                        [
                            Base Objective,
                            leaf,
                        ],
                        [
                            Mesa-optimizer,
                            leaf,
                        ],
                        [
                            Mesa-objective,
                            leaf,
                        ],
                    ]
                    [
                        \citet{2d_robustness}
                        [
                            Objective Robustness,
                            leaf,
                        ],
                        [
                            Capability Robustness,
                            leaf,
                        ],
                    ]
                ]
                [
                    Failures
                    [
                        Proxy Alignment,
                        [
                            Deceptive Alignment,
                            [
                                \citet{the_inner_alignment_problem},
                                leaf
                            ],
                        ],
                    ],
                    [
                        Approximate Alignment,
                        [
                            \citet{the_inner_alignment_problem},
                            leaf
                        ],
                    ],
                    [
                        Suboptimality Alignment,
                        [
                            \citet{the_inner_alignment_problem},
                            leaf
                        ],
                    ],
                ]
                [
                    Methodology
                    [
                        Relaxed Adversarial Training,
                        [
                            \citet{relaxed_adversarial_training},
                            leaf
                        ]
                    ]
                ]
                [
                    Empirical Experiment Proposals
                    [
                        Reward Side-Channels,
                        [
                            \citet{concrete_experiments_in_inner_alignment},
                            leaf
                        ]
                    ]
                    [
                        Cross-Episodic Objectives,
                        [
                            \citet{concrete_experiments_in_inner_alignment},
                            leaf
                        ]
                    ]
                    [
                        Objective Unidentifiability,
                        [
                            \citet{concrete_experiments_in_inner_alignment},
                            leaf
                        ]
                    ]
                    [
                        Zero-Shot Objectives,
                        [
                            \citet{concrete_experiments_in_inner_alignment},
                            leaf
                        ]
                    ]
                    [
                        Robust Reward Learning,
                        [
                            \citet{concrete_experiments_in_inner_alignment},
                            leaf
                        ]
                    ]
                ]
            ] 
        \end{forest}
    }
    \caption{An incomplete and coarse-grained landscape of inner alignment.}
    \label{fig:inner_alignment_taxonomy}
\end{figure*}
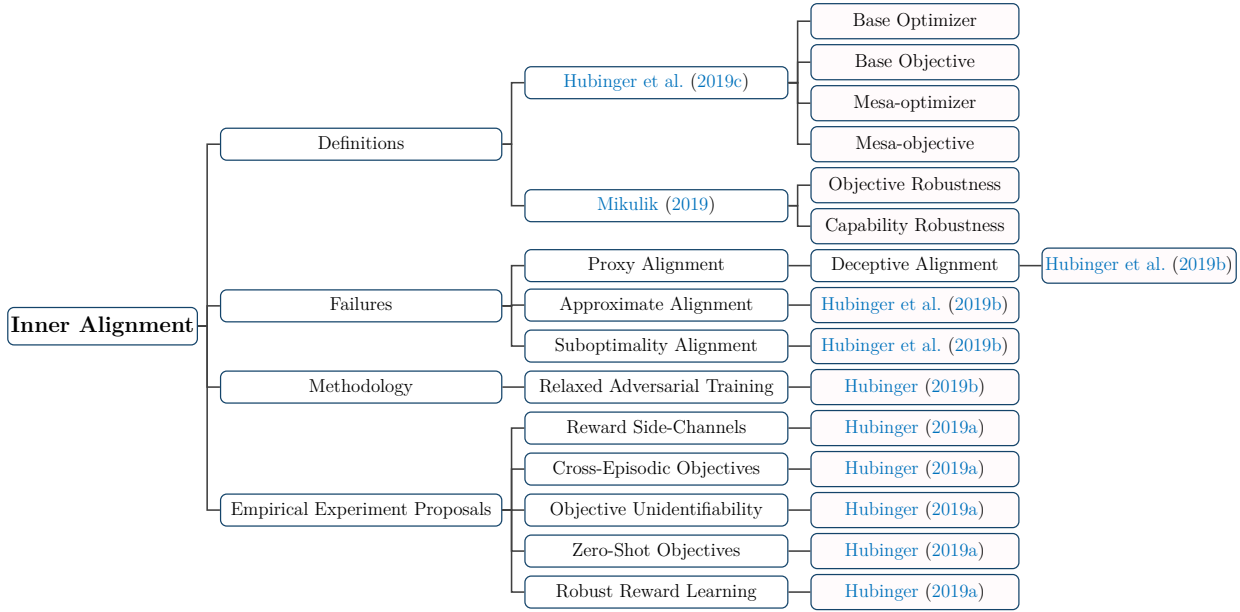

\subsection{Inner Alignment Methodology}

Unlike outer alignment that has been extensively explored (especially in LLMs) recently in an empirical way, inner alignment is limited in its empirical and methodological study. Most discussions on inner alignment are theoretical and usually focusing on its definitions, failure modes and risks. With the rapid development of capabilities of advanced agents, the necessity of methodological studies in inner alignment is becoming urgent. 

To improve inner alignment in advanced agents, \citet{relaxed_adversarial_training} proposes relaxed adversarial training, where an adversary subsystem proposes hypothetical pseudo-inputs estimated to likely induce unacceptable behaviors, rather than attempting to generate concrete unacceptable inputs. The pseudo-inputs describe potential situations that could precipitate unacceptable behaviors if instantiated. A separate oversight subsystem then scrutinizes whether the agent would in fact act unacceptably if the pseudo-inputs were implemented. If so, the system receives a penalty, incentivizing avoidance of potentially unacceptable behaviors. Relaxed adversarial training thus aims to promote inner alignment by penalizing artificial agents for predicted unacceptable behaviors on proposed pseudo-inputs during training. Furthermore, \citet{relaxed_adversarial_training} identifies transparency as the core obstacle to effective relaxed adversarial training for inner alignment. Robust transparency into the model's reasoning is requisite for the oversight system to reliably verify if a model would act unacceptably on proposed pseudo-inputs. Further research should both validate the efficacy of relaxed adversarial training empirically and elucidate transparency mechanisms enabling provable inner alignment in advanced agents.

\subsection{Empirical Experiment Proposals for Inner Alignment}

Similar to the limited methodological exploration of inner alignment, empirical studies that directly observe inner alignment and shed light on its inner workings are scarce. In this aspect, \citet{concrete_experiments_in_inner_alignment} proposes several concrete experiments for inner alignment. We briefly introduce these proposals to demonstrate how inner alignment could be empirically studied.

\begin{itemize}
    \item \textbf{Reward Side-Channels (RSC)} This proposal aims to investigate the degree to which reinforcement learning agents rely on external reward signals versus learning internal value functions. The proposed experiment entails training an RL agent where the previous time step's reward is incorporated into the observation state. At test time, the observed reward would be modified and changes in the agent's behavior measured. This would be compared to two control conditions - one where no reward signal is provided during testing and the other where the next time step's reward is given. To enable the tracking of long-term returns, neural architectures such as LSTM or Transformer which have demonstrated proficiency in capturing long-term dependencies could be explored. By observing the agent's behavioral changes in response to shifts in the external reward, we can assess the robustness of its learned objective. The hypothesis is that reliance on external rewards reflects a lack of internalization of goals.
    \item \textbf{Cross-Episodic Objectives (CEO)} The CEO proposal suggests an experiment to evaluate the tendency of RL agents to exploit non-myopic reward side-channels across episodes. CEO involves training an agent in an environment containing a mechanism for increasing reward in the subsequent episode. The degree to which the agent utilizes this cross-episodic reward channel is measured and compared across different population-based training approaches. The motivation is assessing the conditions under which RL agents depart from solely myopic optimization. This has implications for the choice of training techniques to align agent behavior with human preferences. Approaches relying on short-term optimization, such as amplification and debate, may be less robust than those based on more far-sighted principles like inverse reinforcement learning. By quantifying the prevalence of non-myopic reward hacking across different population training regimes, this experiment aims to provide guidance on preferable alignment strategies.
    \item \textbf{Objective Unidentifiability (OU)} This proposal outlines an experiment to investigate RL agents' tendencies toward pseudo-alignment when trained in environments with multiple viable objectives. The suggested experiment involves constructing a setting with several simple, discernible goals that would equally well explain the true reward signal. After an agent is trained in this environment, it would be evaluated in distinguishing test cases to reveal its learned priorities. Particular interest lies in documenting occurrences of the agent converging to a competent proxy policy that nevertheless fails to robustly maximize the true rewards out-of-distribution. By manipulating architectural factors like inductive biases and model capacity, the preference for different proxies can be assessed.
    \item \textbf{Zero-Shot Objectives (ZSO)} ZSO designs an experiment to evaluate the emergence of goal-directed behavior and coherent objectives in language models without explicit RL training. The proposal creates an interactive environment where a language model can take actions and receive rewards. By analyzing the resulting behaviors through inverse reinforcement learning, the internal learned objectives can be inspected and compared to a RL agent trained directly on the environment's rewards. While contemporary language models might not exhibit truly goal-directed optimization, this experiment aims to investigate the potential emergence of such abilities arising from pure language modeling. Finding that language models can perform non-trivially in certain environments and produce reasonably coherent inferred objectives would suggest these models are starting to develop some intentionality, even without being explicitly trained as RL agents.
    \item \textbf{Robust Reward Learning (RRL)} This proposal defines an experiment to evaluate the efficacy of adversarial training techniques for improving alignment of model-based RL agents. It trains a model-based RL agent, such as an imagination-based planner, to predict environment rewards. The predicted rewards are compared to the true rewards to assess alignment. The agent is then trained adversarially by constructing inputs that maximize divergence between predicted and actual rewards. Alignment is evaluated again after adversarial training. The motivation is to test the ability of adversarial techniques to address reward unidentifiability and enhance alignment.
\end{itemize}

\section{Mechanistic Interpretability}
\label{mechanistic_interpretability}

Mechanistic interpretability \citep{vilone2020explainable} refers to elucidating the internal mechanisms by which a machine learning model transforms inputs into outputs, providing causal and functional explanations for how and why certain predictions are made \citep{neel2022comprehensive,lipton2017mythos}. 
The goal of mechanistic interpretability is to reverse engineer the reasoning process from end to end, decomposing neural networks into interpretable parts and flows of information that provide transparency into their step-by-step reasoning.

Mechanistic interpretability holds great significance for AI alignment. First, interpretability methods can be utilized to audit LLMs, particularly prior to their deployment. We can inspect the alignment efficacy of an LLM, identify misaligned and fallacious outputs, and elucidate why it yields such outputs \citep{neel2022comprehensive,lipton2017mythos}. 
Second, interpretability evaluation metrics could serve as reward functions for optimizing AI alignment \citep{critch2020ai} to incentivize AI systems to maintain goal transparency (e.g., avoiding deceptive alignment) \citep{mcallister2017concrete}. Third, in addition to inspection /architecture transparency, we could also enforce training process transparency that enables us to understand and monitor what's happening and the changes in the training process of AI systems (e.g., emerging behaviors / abilities) \citep{lesswrongTransparencyInterpretability}.

We now discuss recent progress made by mechanistic interpretability on different components in Transformer, including self-attention, multi-layer perceptron (MLP), and neurons.

\tikzstyle{my-box}=[
    rectangle,
    draw=hidden-draw,
    rounded corners,
    text opacity=1,
    minimum height=1.5em,
    minimum width=5em,
    inner sep=2pt,
    align=center,
    fill opacity=.5,
    line width=0.8pt,
]
\tikzstyle{leaf}=[my-box, minimum height=1.5em,
    fill=hidden-pink!80, text=black, align=center,font=\normalsize,
    inner xsep=2pt,
    inner ysep=4pt,
    line width=0.8pt,
]
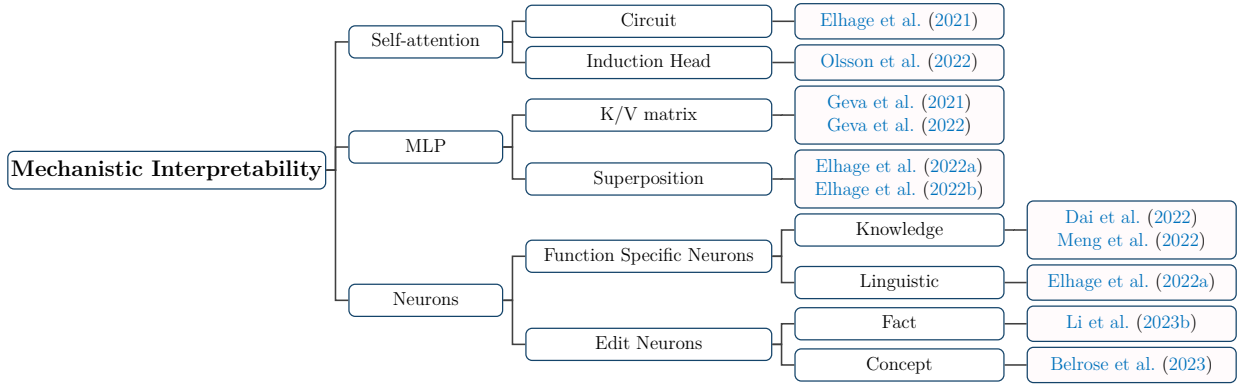
\begin{figure*}[t!]
    \centering
    \resizebox{\textwidth}{!}{
        \begin{forest}
            forked edges,
            for tree={
                grow=east,
                reversed=true,
                anchor=base west,
                parent anchor=east,
                child anchor=west,
                base=center,
                font=\large,
                rectangle,
                draw=hidden-draw,
                rounded corners,
                align=center,
                text centered,
                minimum width=5em,
                edge+={darkgray, line width=1pt},
                s sep=3pt,
                inner xsep=2pt,
                inner ysep=3pt,
                line width=0.8pt,
                ver/.style={rotate=90, child anchor=north, parent anchor=south, anchor=center},
            },
            where level=1{text width=8em,font=\normalsize,}{},
            where level=2{text width=13em,font=\normalsize,}{},
            where level=3{text width=11em,font=\normalsize,}{},
            where level=4{text width=11em,font=\normalsize,}{},
            [
                \textbf{Mechanistic Interpretability}
                [
                    Self-attention
                    [
                        Circuit
                        [
                            \citet{elhage2021mathematical}
                            , leaf
                        ]
                    ]
                    [
                        Induction Head
                        [
                            \citet{olsson2022context}
                            , leaf
                        ]
                    ]
                ]
                [
                    MLP
                    [
                        K/V matrix
                        [
                            \citet{geva2021transformer} \\ \citet{geva2022transformer}
                            , leaf
                        ]
                    ]
                    [
                        Superposition
                        [
                            \citet{elhage2022solu} \\ \citet{elhage2022superposition}
                            , leaf
                        ]
                    ]
                ]
                [
                    Neurons
                    [
                        Function Specific Neurons
                        [
                            Knowledge
                            [
                                \citet{2021Knowledge} \\ \citet{meng2022locating}
                                , leaf
                            ]
                        ]
                        [
                            Linguistic
                            [
                                \citet{elhage2022solu}
                                , leaf
                            ]
                        ]
                    ]
                    [
                        Edit Neurons
                        [
                            Fact
                            [
                                \citet{li2023inference}
                                , leaf
                            ]
                        ]
                        [
                            Concept
                            [
                                \citet{belrose2023leace}
                                , leaf
                            ]
                        ]
                    ]
                ]
            ]
        \end{forest}
    }
    \caption{An overview of current mechanistic interpretability research, including mechanistic studies on self-attention (circuit, induction head), MLP (K/V matrix, superposition) and neurons (function specific neurons, edit neurons)}
    \label{fig:mechanistic_interpretability_taxonomy}
\end{figure*}

\subsection{Mechanistic Interpretability on Self-Attention} 
The self-attention (SA) mechanism is widely used to aggregate contextual information by directly ``attending'' to specific tokens. Each token in the context is paired with the current token to calculate ``compatibility'' score. Such scores are used to weight tokens in the context window so that learned representations of tokens are aggregated for predicting the next-step decision (e.g., next-token prediction). \citet{elhage2021mathematical} investigate a SA-layer-only  (MLP layers removed) Transformer \citep{vaswani2017attention} and find interesting neural circuits. In their work, SA layer is viewed as performing read and write operations into the residual stream, modifying the original token embeddings. They discover that the QK circuits focus on the next potential token, while the OV circuits tend to copy previous tokens, which they refer to as induction heads.

\citet{olsson2022context} further investigate induction heads and attribute the general in-context learning ability of LLMs to the manifestation of induction heads. They present evidence for both small SA-only models and large models with MLPs.

\subsection{Mechanistic Interpretability on MLP} 
MLP layers introduce non-linear transformations in Transformer and account for a large proportion of parameters, significantly enhancing the model's expressive power. Such non-linear transformations enable Transformer to capture complex relationships and patterns in data, making it more capable of representing intricate functions \citep{geva2021transformer,geva2022transformer,elhage2022solu}. 
Due to the non-linear nature and high dimensionality of data, directly reverse engineering MLPs is challenging. 

To address this issue, \citet{elhage2022solu} propose an interpretable activation function called SoLU, which can deal with polysemantic neurons and encourage feature-neuron alignment. SoLU facilitates neural networks to learn human-interpretable neuron patterns without significant performance degradation. \citet{elhage2022superposition} further examine the phenomenon of feature superposition in MLPs using a simple network with ReLU activation. Their experiments demonstrate that linear models do not exhibit feature superposition (i.e., ambiguity), whereas non-linear models display increasingly apparent feature superposition with the increase in data sparsity.

\subsection{Mechanistic Interpretability on Neurons} 
\citet{Chris2022mechanistic} views neurons as variables in a computer program. Previous studies have demonstrated the existence of different types of neurons in Transformers, such as knowledge neurons \citep{2021Knowledge,meng2022locating} and neurons corresponding to specific linguistic properties \citep{elhage2022solu}. Interventions at the neuron level could change the outputs of the entire neural network. This is leveraged to enhance the factuality of machine-generated content \citep{li2023inference} and to eliminate the influence of specific concepts \citep{belrose2023leace}. By understanding and manipulating these individual neurons, we can gain insights into how a neural model processes and represents information, which benefits developing interpretable and safe AI systems.

\subsection{Challenges}

Despite the success mentioned above, mechanistic interpretability (MI) is still at an incipient stage of research. Most current MI studies have been done under restricted conditions, e.g., on a toy language model (typically one-to-four-layer Transformer language models), or with predefined simple tasks \citep{wang2022interpretability,elhage2021mathematical}. Even so, MI is confronted with a variety of challenges, e.g., the superposition hypthothesis  \citep{elhage2022superposition}, non-linear representations \citep{lee2022current}.

The superposition hypothesis that neural networks attempt to represent more features than neurons or dimensions they have, has been compellingly verified \citep{elhage2022superposition}. Feature superposition in neural networks explains the phenomenon of neuron polysemanticity where a neuron corresponds to several unrelated features \citep{elhage2022superposition}. Although superposition is useful for neural representations, it poses a challenge to MI as it makes it difficult to disentangle representations, hence preventing MI from explaining relations between disentangled representations or features in a simple and human-understandable way \citep{lee2022superposition,lee2022current}.

\section{Attacks on Aligned Language Models}
\label{attack_methods_against_aligned_language_models}

Large language models have encountered challenges posed by various attack methods. Malicious systems could intentionally prompt LLMs to generate harmful, biased, or toxic text, thereby posing significant risks of misuse \citep{DBLP:conf/nips/BrownMRSKDNSSAA20,ouyang2022training}. As a primary strategy to mitigate these risks. LLM alignment via RLHF has been widely adopted \citep{ouyang2022training,glaese2022improving}. This alignment can be considered as a safeguard against these attacks.

Recent studies show that such aligned LLMs exhibit defensive capabilities against malicious attacks. \citet{carlini2023aligned} demonstrate that aligned LLMs can effectively counter a wide range of (white-box) NLP attacks, even adversarial inputs. \citet{li2023multi} showcase that ChatGPT is able to decline providing answers to privacy-sensitive questions.

Nonetheless, alignment techniques are not infallible. For example, through repeated interactions, humans can ``trick'' these models into generating harmful content, as seen in jailbreaking attacks. In addition to jailbreaking, other methods have also been explored to  breach the safeguard of aligned models. We divide these efforts into three categories according to the nature of the attack methods. The overview of these attacks is presented in Figure \ref{fig:alignment_attack_taxonomy}.

\tikzstyle{my-box}=[
    rectangle,
    draw=hidden-draw,
    rounded corners,
    text opacity=1,
    minimum height=1.5em,
    minimum width=5em,
    inner sep=2pt,
    align=center,
    fill opacity=.5,
    line width=0.8pt,
]
\tikzstyle{leaf}=[my-box, minimum height=1.5em,
    fill=hidden-pink!80, text=black, align=center,font=\normalsize,
    inner xsep=2pt,
    inner ysep=4pt,
    line width=0.8pt,
]
\begin{figure*}[t!]
    \centering
    \resizebox{\textwidth}{!}{
        \begin{forest}
            forked edges,
            for tree={
                grow=east,
                reversed=true,
                anchor=base west,
                parent anchor=east,
                child anchor=west,
                base=center,
                font=\large,
                rectangle,
                draw=hidden-draw,
                rounded corners,
                align=center,
                text centered,
                minimum width=5em,
                edge+={darkgray, line width=1pt},
                s sep=3pt,
                inner xsep=2pt,
                inner ysep=3pt,
                line width=0.8pt,
                ver/.style={rotate=90, child anchor=north, parent anchor=south, anchor=center},
            },
            where level=1{text width=10em,font=\normalsize,}{},
            where level=2{text width=13em,font=\normalsize,}{},
            where level=3{text width=11em,font=\normalsize,}{},
            where level=4{font=\normalsize,}{},
            [
                \textbf{Attack Methods against Alignment}
                [
                    Privacy Attacks
                    [
                        Jailbreaking Prompts
                        [
                            \citet{li2023multi} \\ \citet{deng2023jailbreaker}
                            , leaf
                        ]
                    ]
                ]
                [
                    Backdoor Attacks
                    [
                        Prompt Injection
                        [
                            \citet{liu2023prompt} \\ \citet{zhao2023prompt} \\ \citet{greshake2023more} \\ \citet{kandpal2023backdoor}
                            , leaf
                        ]
                    ]
                    [
                        Backdoors Injection at RLHF
                        [
                            \citet{shi2023badgpt}
                            , leaf
                        ]
                    ]
                ]
                [
                    Adversarial Attacks
                    [
                        Adversarial Prompts
                        [
                            \citet{zou2023universal}
                            , leaf
                        ]
                    ]
                    [
                        Visual Adversarial Examples
                        [
                            \citet{carlini2023aligned} \\ \citet {qi2023visual}
                            , leaf
                        ]
                    ]
                ]
            ]
        \end{forest}
    }
    \caption{An overview of attack methods that might be capable of breaking through the safeguard of aligned models.}
    \label{fig:alignment_attack_taxonomy}
\end{figure*}
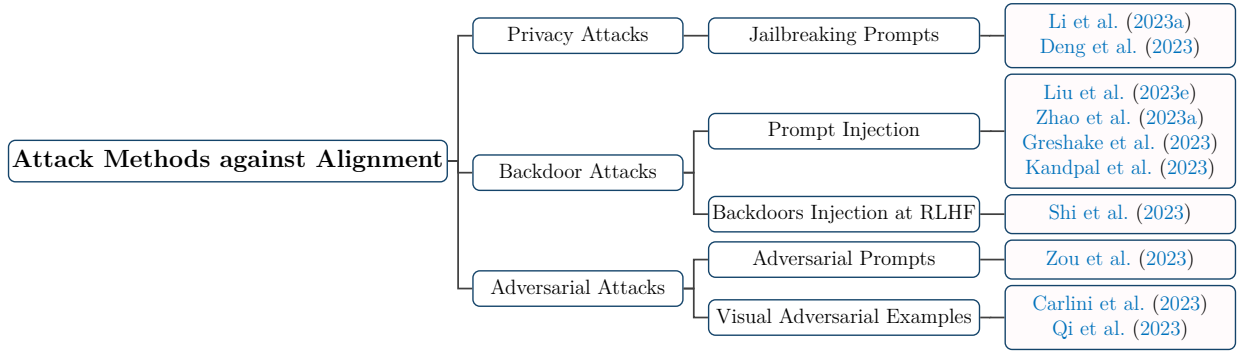

\subsection{Privacy Attacks}
A privacy attack constitutes an approach wherein machine learning models are exploited, with attackers attempting to extract private or sensitive information about the training data from the model's outputs \citep{rigaki2020survey,mireshghallah2020privacy,sousa2023keep,guo2022threats}. Legal frameworks related to personal data protection necessitate the preservation of privacy in training data, as leakage could result in legal repercussions \citep{gdpr}. 
Currently, privacy attacks on language models can be categorized into four types: (1)Gradient Reconstruction Attacks during the model distributed training stage, (2)Attribute Inference Attacks, (3)Prompt Attacks and (4)Inversion Attacks during the inference stage.

Gradient Reconstruction Attacks aim at attacking models during the distributed training, where information such as training data and gradients is exchanged between devices. Attackers can spy on this information exchange to reconstruct privacy-sensitive details from the training data \citep{gupta2022recovering, deng2021tag}. Although no specific research has targeted reconstruction attacks on aligned models, these spying-based attacks remain a potential threat when aligned models are tuned in a distributed training way.

Attribute Inference Attacks infers data ownership and privacy attributes by comparing the performance of a target model with that of similar models \citep{song2019auditing, hisamoto2020membership, mireshghallah2022quantifying}. Such methods often require access to output probabilities, logits, or hidden states, making implementation on black-box APIs (which provide only textual outputs) challenging.

Inversion Attacks \citep{song2020information, elmahdy2022privacy} aim to inversely get input information using model gradients, parameter states, etc. Implementing such methods is also challenging for LLMs as they usually have a huge amount of parameters.

Prompt Attacks involve designing or searching for prompts that lead LMs to output information from the training data, including private details \citep{carlini2021extracting, lehman2021does, li2023multi, deng2023jailbreaker}. This approach is particularly targeted towards LLMs and poses a significant threat to aligned LLMs. 
\citet{li2023multi} propose a new attack method that extracting personal identity information(PII) from ChatGPT and New Bing by multi-step \textbf{Jailbreaking Prompts}. And it shows the New Bing is more vulnerable to direct extraction of PII due to its search engine integration, posing unintended privacy risks.

\subsection{Backdoor Attacks}
Backdoor attacks are a class of methods aimed at machine learning models, with the objective of causing the model to produce specific, incorrect outputs when certain backdoor triggers are detected \citep{gao2020backdoor, li2022backdoor, sheng2022survey}. Backdoor attacks can be categorized into two types: (1)Data Poisoning and (2)Model Poisoning.

Data Poisoning introduces triggers (e.g., instances generated with special lexical or syntactic templates) into the training data to implement a backdoor attack on the model \citep{li2021hidden, qi2021hidden, chen2021textual}. Previous studies primarily focused on tasks like text classification, but these methods can also be extended to tasks such as question answering and text generation. Backdoor attacks on aligned models often utilize \textbf{Prompt Injection} techniques \citep{liu2023prompt, zhao2023prompt, greshake2023more, kandpal2023backdoor}, where the prompt itself serves as the trigger, eliminating the need for external inputs. When a trigger prompt is used, it could lead to unintended outcomes.

Model Poisoning achieves backdoor attacks by manipulating the model itself, involving modifications to word embeddings, loss functions, output representations, etc. \citep{yang2021careful, wallace2020concealed, li2021backdoor}. 
Recently, \citet{shi2023badgpt} propose a new attack method called BadGPT, which makes \textbf{Backdoors Injection at RLHF} to the reward model. This method has two stages: first, injecting backdoors into the reward model to make it give wrong rewards when a specific trigger word appears. Second, using the backdoored reward model to fine-tune the language model, thereby injecting a backdoor into the aligned model.

\subsection{Adversarial Attacks} 
Adversarial attacks are techniques employed to compromise the performance or behavior of machine learning models, particularly deep learning models, by introducing small and carefully crafted perturbations to the input data \citep{akhtar2018threat,zhang2020adversarial,qiu2022adversarial,goyal2023survey}. These perturbations are often imperceptible to humans but can lead the model to produce incorrect or unexpected outputs. Prior works on textual tasks use greedy attack heuristics \citep{wallace2019universal} or employ discrete optimization to search for an input text that triggers adversarial behavior \citep{wallace2019universal,jones2023automatically}.

For aligned models, \citet{zou2023universal} proposed a simple yet potent attack strategy that combines greedy search and gradient-based techniques to automatically generate \textbf{Adversarial Prompts}, causing aligned LLMs to produce contentious behaviors. 

Studies by \citet{carlini2023aligned} and \citet{qi2023visual} demonstrate that multimodal language models exhibit reduced defenses against white-box adversarial attacks, such as \textbf{Visual Adversarial Examples}. The high-dimensional visual input space renders these models more susceptible, and the diverse outputs present additional targets for adversarial attacks.

\section{Alignment Evaluation}
\label{alignment_evaluation}
Evaluation is important for alignment research, especially for the development of empirical alignment methods. We review methods and resources pertaining to LLM alignment. As illustrated in Figure \ref{fig:alignment_evaluation_taxonomy}, our alignment evaluation landscape is structured across multiple levels. The first level illustrates the five aspects of LLM outer alignment we are focusing on, namely: 1) factuality, 2) ethics, 3) toxicity, 4) stereotype and bias, and 5) general evaluation. Genaral evaluation does not target at a single specific dimension of alignment, e.g., factuality, toxicity. Instead, it evaluates multiple dimensions of alignment or the general aspects of LLM alignment. The subsequent level categorizes the primary evaluation methods presently available in each respective area. We distinguish task-specific evaluation from LLM-centered evaluation at this level. Task-specific evaluation refers to evaluating alignment quality on downstream tasks while LLM-centered evaluation designs evaluation benchmarks, methods or metrics directly for LLMs.  The third level is designated for fine-grained classification or showcasing related works, enabling readers to swiftly pinpoint their areas of interest.

\tikzstyle{my-box}=[
    rectangle,
    draw=hidden-draw,
    rounded corners,
    text opacity=1,
    minimum height=1.5em,
    minimum width=5em,
    inner sep=2pt,
    align=center,
    fill opacity=.5,
    line width=0.8pt,
]
\tikzstyle{leaf}=[my-box, minimum height=1.5em,
    fill=hidden-pink!80, text=black, align=center,font=\normalsize,
    inner xsep=2pt,
    inner ysep=4pt,
    line width=0.8pt,
]
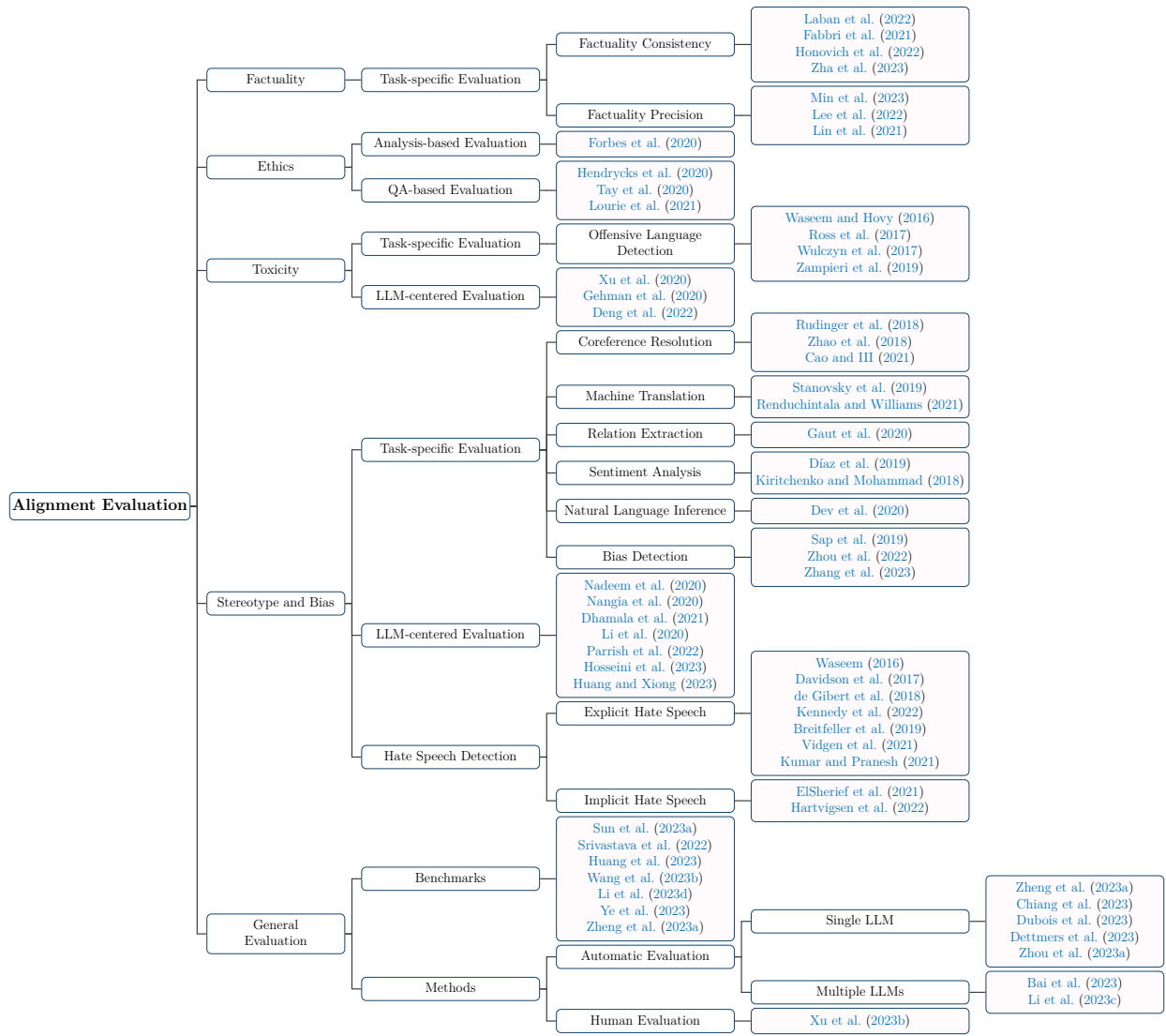
\begin{figure*}[t!]
    \centering
    \resizebox{\textwidth}{!}{
        \begin{forest}
            forked edges,
            for tree={
                grow=east,
                reversed=true,
                anchor=base west,
                parent anchor=east,
                child anchor=west,
                base=center,
                font=\large,
                rectangle,
                draw=hidden-draw,
                rounded corners,
                align=center,
                text centered,
                minimum width=5em,
                edge+={darkgray, line width=1pt},
                s sep=3pt,
                inner xsep=2pt,
                inner ysep=3pt,
                line width=0.8pt,
                ver/.style={rotate=90, child anchor=north, parent anchor=south, anchor=center},
            },
            where level=1{text width=10em,font=\normalsize,}{},
            where level=2{text width=13em,font=\normalsize,}{},
            where level=3{text width=13em,font=\normalsize,}{},
            where level=4{text width=16em,font=\normalsize,}{},
            where level=5{text width=13em,font=\normalsize,}{},
            where level=6{text width=13em,font=\normalsize,}{},
            where level=7{text width=13em,font=\normalsize,}{},
            where level=8{text width=13em,font=\normalsize,}{},
            [
                \textbf{Alignment Evaluation}
                [
                    Factuality
                    [
                        Task-specific Evaluation
                        [
                            Factuality Consistency
                            [
                                \citet{laban2022summac} \\ \citet{fabbri2021qafacteval} \\ \citet{honovich2022true} \\ \citet{zha2023alignscore}
                                , leaf
                            ]
                        ]
                        [
                            Factuality Precision
                            [
                                \citet{min2023factscore} \\ \citet{lee2022factuality} \\
                                \citet{lin2021truthfulqa}
                                , leaf
                            ]
                        ]
                    ]
                ]
                [
                    Ethics
                    [
                        Analysis-based Evaluation
                        [
                            \citet{forbes2020social}
                            , leaf
                        ]
                    ]
                    [
                        QA-based Evaluation
                        [
                            \citet{hendrycks2020aligning} \\ \citet{tay2020would} \\ \citet{lourie2021scruples}
                            , leaf
                        ]
                    ]
                ]
                [
                    Toxicity
                    [
                        Task-specific Evaluation
                        [
                            Offensive Language \\ Detection
                            [
                                \citet{waseem2016hateful} \\ \citet{ross2017measuring} \\ \citet{wulczyn2017ex} \\ \citet{zampieri2019predicting}
                                , leaf
                            ]
                        ]
                    ]
                    [
                        LLM-centered Evaluation 
                        [
                            \citet{xu2020recipes} \\ \citet{gehman2020realtoxicityprompts} \\ \citet{DBLP:conf/emnlp/DengZ0ZMMH22}
                            , leaf
                        ]
                    ]
                ]
                [
                    Stereotype and Bias
                    [
                        Task-specific Evaluation
                        [
                            Coreference Resolution
                            [
                                \citet{rudinger2018gender} \\ \citet{zhao2018gender} \\ \citet{DBLP:journals/coling/CaoD21}
                                , leaf
                            ]
                        ]
                        [
                            Machine Translation
                            [
                                \citet{DBLP:conf/acl/StanovskySZ19} \\ \citet{DBLP:journals/corr/abs-2104-07838} 
                                , leaf
                            ]
                        ]
                        [
                            Relation Extraction
                            [
                                \citet{DBLP:conf/acl/GautSTHQEZMBCW20} \\ 
                                , leaf
                            ]
                        ]
                        [
                            Sentiment Analysis
                            [
                                \citet{DBLP:conf/ijcai/DiazJLPG19} \\ 
                                \citet{DBLP:conf/starsem/KiritchenkoM18}
                                , leaf
                            ]
                        ]
                        [
                            Natural Language Inference
                            [
                                \citet{DBLP:conf/aaai/DevLPS20} 
                                , leaf
                            ]
                        ]
                        [
                            Bias Detection
                            [
                                \citet{sap2019social} \\ \citet{zhou2022towards} \\ \citet{zhang2023corgi}
                                , leaf
                            ]
                        ]
                    ]
                    [
                        LLM-centered Evaluation
                        [
                            \citet{nadeem2020stereoset} \\
                            \citet{DBLP:conf/emnlp/NangiaVBB20} \\ \citet{DBLP:conf/fat/DhamalaSKKPCG21} \\ \citet{li2020unqovering} \\ \citet{DBLP:conf/acl/ParrishCNPPTHB22} \\ 
                            \citet{hosseini2023empirical} \\ \citet{huang2023cbbq}
                            , leaf
                        ]
                    ]
                    [
                        Hate Speech Detection
                        [
                            Explicit Hate Speech
                            [
                                \citet{DBLP:conf/acl-nlpcss/Waseem16} \\ \citet{DBLP:conf/icwsm/DavidsonWMW17} \\ \citet{DBLP:conf/acl-alw/GibertPPC18} \\ \citet{Kennedy2022} \\ \citet{DBLP:conf/emnlp/BreitfellerAJT19} \\ \citet{DBLP:conf/acl/VidgenTWK20} \\ \citet{kumar2021tweetblm}
                                , leaf
                            ]
                        ]
                        [
                            Implicit Hate Speech
                            [
                                \citet{DBLP:conf/emnlp/ElSheriefZMASCY21} \\ \citet{DBLP:conf/acl/HartvigsenGPSRK22}
                                , leaf
                            ]
                        ]
                    ]
                ]
                [
                    General \\ Evaluation
                    [
                        Benchmarks
                        [
                            \citet{sun2023safety} \\ \citet{srivastava2022beyond} \\ \citet{huang2023trustgpt} \\ \citet{wang2023pandalm} \\ \citet{alpaca_eval} \\ \citet{ye2023flask} \\ \citet{zheng2023judging}
                            , leaf
                        ]
                    ]
                    [
                        Methods
                        [
                            Automatic Evaluation
                            [
                                Single LLM
                                [
                                    \citet{zheng2023judging} \\ \citet{chiang2023vicuna} \\  \citet{dubois2023alpacafarm} \\ \citet{dettmers2023qlora} \\ \citet{zhou2023lima}
                                    , leaf
                                ]
                            ]
                            [
                                Multiple LLMs
                                [
                                    \citet{bai2023benchmarking} \\ \citet{li2023prd}
                                    , leaf
                                ]
                            ]
                        ]
                        [
                            Human Evaluation
                            [
                                \citet{xu2023critical}
                                , leaf
                            ]
                        ]
                    ]
                ]
            ]
        \end{forest}
    }
    \caption{The taxonomy of alignment evaluation methods, including factuality and truthfulness, ethics, toxicity, stereotype \& bias, and comprehensive evaluations.}
    \label{fig:alignment_evaluation_taxonomy}
\end{figure*}

\subsection{Factuality Evaluation}
\label{factuality_and_truthfulness_alignment_evaluation}
Machine-generated content should be congruent with facts, eschewing the creation of hallucination content. Additionally, each piece of generated information should be factually accurate. These suggest that factuality evaluation at least comprise factual consistency evaluation and factual precision assessment.

Factual consistency requires that generated content should be consistent with given context. As downstream tasks, like text summarization, dialogue, are usually accompanied with rich context, many task-specific actuality evaluation studies are conducted on such downstream tasks. While this could be done on a single task \citep{laban2022summac,fabbri2021qafacteval}, consistency evaluation on multiple tasks is more convincing. \citet{honovich2022true} provide a comprehensive analysis of factual consistency, incorporating a variety of metrics, tasks, and datasets. Their study consolidates 11 datasets from a variety of tasks into a unified format. They also compare the effectiveness of existing methods for evaluating consistency, using this unified format. The ALIGNSCORE metric, proposed by \citep{zha2023alignscore}, is designed to cover a wide range of factual consistency evaluation scenarios, such as contradiction and hallucination across various lengths and tasks. The metric is developed through the training of an aligned model, which restructures 15 datasets from 7 NLP tasks. These tasks include Natural Language Inference, Question Answering, Paraphrasing, Fact Verification, Information Retrieval, Semantic Similarity, and Summarization.

Factuality precision evaluation is also task-specific. \citet{lee2022factuality} present a benchmark and a metric for factual precision evaluation. They use both factual and non-factual prompts to obtain generated texts from an LLM. The used specific tasks include named entity recognition and entailment.  \citet{min2023factscore} introduce FACTSCORE, a novel method that deconstructs long-form text into atomic facts or individual pieces of information, assigning a binary label to each fact. However, the efficacy of this method is largely dependent on the acquisition of these atomic facts, making the selection of evaluation tasks a critical factor. They concentrate on the generation of individual biographies, as the atomic facts contained within these biographies can be verified by Wikipedia.

Factual precision is also related to the model's ability to answer questions truthfully. \citet{lin2021truthfulqa} present TruthfulQA and argue that the training objectives of LLMs could potentially influence them to produce false responses. As a result, they devise a series of highly inductive questions to actively assess LLMs.

Evaluating factuality presents two significant challenges. First, while factuality encompasses countless facts, the scope of factuality evaluation so far inherently limited. Second, not all facts in real life are easy to be divided into atomic facts. Current evaluation methods fall short when dealing with complex information that can't be simplified, such as assessing factualness that requires sophisticated reasoning.

\subsection{Ethics Evaluation}
\label{ethical_alignment_evaluation}
Ethics is a multifaceted issue pervading nearly every aspect of society, characterized by dialectical thinking. It encompasses a broad spectrum of considerations, including good and evil, right and wrong, virtue and vice, justice and crime, which are all related to individuals \citep{martinez2020more}. As a result, most LLM ethics evaluations employ a straightforward methodology. This involves posing questions related to ethics and morality to the assessed model, and subsequently assessing the model's alignment with human values on these matters based on its responses.

\citet{hendrycks2020aligning} introduce the ETHICS benchmark, a comprehensive collection of over 130,000 scenarios spanning five domains of ethics: justice, virtue ethics, deontology, utilitarianism, and commonsense morality. Crafted by individuals who have passed a qualification test, these scenarios serve as brief statements that tested models must have to predict moral sentiments as either acceptable or unacceptable. Similarly, \citet{tay2020would} propose the MACS benchmark, which includes 200,000 chosen questions for learning alignment with cultural values and social preferences. This benchmark distinguishes itself through its unique data collection method, drawing from the popular online game ``Would You Rather?''. The questions and answers provided in this game offer a more comprehensive dataset than those relying solely on a few annotators. In contrast to these works that involve short text pieces, \citet{lourie2021scruples} collect real-life anecdotes in a long-text format, rich in detail. The original data is sourced from a public sub-forum on Reddit, a platform where individuals seek advice from online acquaintances to navigate real-life situations.

The evaluation methodology employed in Social Chemistry 101 \citep{forbes2020social} diverges from traditional QA-based approaches. They deconstruct tacit commonsense rules into twelve distinct dimensions of human judgment, including cultural pressure, action-taking, social judgment, etc. The study offers a range of perspective choices to annotators for specific scenarios. This innovative approach enables annotators to examine ethical situations from diverse viewpoints, thereby enriching the depth and breadth of the annotated data.

It is clear that assessments in the realm of ethical morality depend on real-world contextual data. While some initiatives have factored in cultural backgrounds during data collection, the primary data and reference responses largely stem from the researchers' own cultural contexts. As a result, it is incumbent upon researchers to dedicate themselves to the collection and generation of data that mirrors a diverse range of cultural backgrounds, which can then be utilized as evaluation datasets.

\subsection{Toxicity Evaluation}
\label{toxicity_evaluation}
Toxicity is defined as harmful and destructive behaviors or attitudes that can manifest in interpersonal relationships, work environments, or other social settings. This might take the form of control over others, manipulation, belittlement, or malicious attacks. These behaviors can be overt or covert, causing damage to the self-esteem, safety, and well-being of individuals. There is a wide array of \textit{toxic language} that includes: (i) Suggestions leading to self-harming behaviors; (ii) Content that is pornographic or violent in nature; (iii) Harassment, belittlement, offense, insults, and hate speech; (iv) Suggestions advocating for aggressive or violent actions, such as cyberbullying; (v) Guidelines or directions for seeking illegal goods or services.

We categorize the toxicity evaluation into two dimensions: task-specific evaluation and LLM-centered evaluation. Task-specific evaluation pertains to assessing the level of toxicity displayed by a model when it's applied to specific downstream tasks. The diversity of tasks within the field of NLP significantly enriches our evaluation scenarios, enabling us to more comprehensively investigate the contexts in which language models manifest toxicity. On the other hand, LLM-centered evaluation evaluates LLMs directly based on the generated outputs to gauge their toxicity. In task-specific evaluation, the model's performance might be constrained by the specific tasks, potentially behaving in ways that prioritize achieving ``high accuracy''. In contrast, in LLM-centered evaluation, the model predominantly responds based on its inherent knowledge and tendencies. Such an evaluation approach is currently the mainstream method that is gaining significant attention and adoption.

\subsubsection{Task-specific Evaluation}
Offensive language detection can be categorized as a downstream classification task. Offensive language pertains to the deployment of injurious articulates in a sacrilegious, extremely discourteous, impolite, or crude fashion, aiming to derogate the specified individual or group \citep{chen2012detecting,razavi2010offensive}.  Early works on offensive language detection \citep{waseem2016hateful} from Twitter provides datasets which only share Twitter IDs and bullying types, lacking detailed content. Building on this, \citet{ross2017measuring} focus on the German refugee situation with a modest dataset of just over 400 tweets. \citet{wulczyn2017ex} analyzes a vast corpus from Wikipedia, exploring 95 million user-article interactions for personal attacks and toxicity. In contrast, \citet{zampieri2019predicting} return to Twitter, introducing a dataset with detailed annotations on attack types and targets, enriching the understanding of offensive language in social media.

\subsubsection{LLM-centered Evaluation}
To directly evaluate toxicity in LLMs, LLM-centered evaluations trigger models to yield toxic responses. These evaluations mainly concentrate on the toxicity level of the yielded outputs.

BAD \citep{xu2020recipes} necessitates individuals to engage in adversarial dialogues with advanced models to prompt them into generating unsafe responses. This method mirrors the potential adversarial challenges models could face upon deployment. By utilizing this method, they gather an extensive dataset of dialogues that could be further utilized to assess the toxicity in LLMs. Similarly, RealToxicityPrompts \citep{gehman2020realtoxicityprompts} constructs a large set of prompts and performs a comprehensive evaluation on various language models like GPT-1 \citep{radford2018improving}, GPT-2 \citep{radford2019language}, GPT-3 \citep{DBLP:conf/nips/BrownMRSKDNSSAA20}, and CTRL \citep{DBLP:journals/corr/abs-1909-05858}. The findings reveal that even from seemingly innocuous prompts, pretrained LMs could degenerate into producing toxic text. In particular, GPT-1 exhibits the highest toxicity, which might be attributed to the higher amount of toxic content in its training data. This observation accentuates the importance of rigorous data scrutiny for LLMs. Shifting focus to the Chinese context, COLD \citep{DBLP:conf/emnlp/DengZ0ZMMH22} explores the detection of offensive language in Chinese. It collects a significant volume of real-text data from social media platforms and evaluates several open-source models. Consistent with previous findings, irrespective of the presence of offensive content in the input prompts, the generated outputs from these models often encompass offensive language.

\subsection{Stereotype and Bias Evaluation}
\label{bias_evaluation}
Prejudice and stereotype bias are defined as preconceived attitudes, usually based on a group's race, gender, sexual orientation, religion, or other characteristics. These attitudes may be negative or positive but are generalized judgments of a group rather than based on an individual's actual behavior or traits. Prejudice may lead to discrimination or other unjust behaviors.

We also categorize the stereotype and bias evaluation into two dimensions: task-specific evaluation and LLM-centered evaluation. The former pertains to the assessment of biases when the model is applied to specific downstream tasks, while the latter directly evaluates the inherent biases present within the model.

Hate speech is language used to express hatred towards a target individual or group, or is intended to demean, humiliate, or insult members of a group based on attributes such as race, religion, national origin, sexual orientation, disability, or gender \citep{DBLP:conf/icwsm/DavidsonWMW17,DBLP:conf/www/BadjatiyaG0V17,warner2012detecting,DBLP:conf/acl-socialnlp/SchmidtW17,DBLP:conf/www/DjuricZMGRB15}. Since hate speech is usually associated with bias, we discuss hate speech detection in LLM-generated content after the introduction to the general bias evaluation. 

\subsubsection{Task-specific Evaluation}

To understand where a model reinforces biases in its outputs, many studies investigate how these biases occur in downstream tasks. These tasks can be standardized into generative tasks through prompt engineering, making them suitable for evaluating LLMs.

The task of coreference resolution is among the first used to study biases in language models, typically employing F1 scores as a metric. Winogender \citep{rudinger2018gender} and WinoBias \citep{zhao2018gender} both address gender biases related to occupations. They utilize the Winogram-schema style \citep{DBLP:conf/aaaiss/Levesque11} of sentences, revealing stereotypes in coreference resolution systems when interpreting ``HE'' and ``SHE''. GICOREF \citep{DBLP:journals/coling/CaoD21} focuses on the model's performance on texts related to non-binary and binary transgender individuals. All evaluated systems perform worse on these texts than on binary gendered texts, with the best model achieving only a 34\% F1 score.

The WinoMT Challenge Set \citep{DBLP:conf/acl/StanovskySZ19} is the first to explore gender bias in machine translation task at a large scale, integrating both Winogender and WinoBias and setting evaluation standards for eight languages. Gender accuracy in translations is the primary metric. They discover significant translation biases in both commercial MT systems and advanced academic models. \citet{DBLP:journals/corr/abs-2104-07838} expands this task to cover 20 languages, examining whether models still make gender translation mistakes with unambiguous contexts. They find accuracy levels generally below 70\%, especially when perceived occupational gender contradicts the context.

Similarly, WikiGenderBias \citep{DBLP:conf/acl/GautSTHQEZMBCW20} is a dataset aimed at analyzing gender bias in the task of relation extraction. It evaluates gender bias in NRE systems by comparing model performance when extracting occupation information about women versus men from 45,000 sentences.

\citet{DBLP:conf/ijcai/DiazJLPG19} finds that changing age and gender terms in sentences influence model scores in sentiment analysis. The Equity Evaluation Corpus (EEC) \citep{DBLP:conf/starsem/KiritchenkoM18} delves deeper into categories of race and gender, providing comprehensive evaluations of 219 sentiment analysis systems.

\citet{DBLP:conf/aaai/DevLPS20} utilizes Natural Language Inference (NLI) to detect biases in models. They establish a broad benchmark based on polarized adjectives and ethnic names, which not only includes gender but also countries and religions. Biases in models are determined by deviations from neutral answers. Their results reveal evident biases in GloVe, ELMo, and BERT models.

Bias detection can be also categorized as a classification task. \citet{sap2019social} offers a dataset with 150,000 annotated social media posts highlighting social bias frames across various demographic groups. Further localization efforts, particularly for non-English languages, give rise to CDail-Bias \citep{zhou2022towards}. This is the first  Chinese dataset targeting social bias in dialog systems, covering race, gender, region, and occupation domains. In a more specialized direction, CORGI-PM \citep{zhang2023corgi} centers exclusively on gender bias. This unique Chinese corpus encompasses 32,900 labeled sentences, marking a first in sentence-level gender bias in Chinese. Their innovative methodology uses an automated process for sampling pronounced gender bias, followed by a re-ranking based on sentence-level bias probability for more precise bias detection and mitigation.

\subsubsection{LLM-centered Evaluation}
In direct bias evaluations of language models, there are various assessment methodologies. Some adopt a contrasting method using associated sentence pairs: one with more stereotypes, and the other with fewer \citep{nadeem2020stereoset, DBLP:conf/emnlp/NangiaVBB20}. Biases are detected through the language model's likelihood of recovering masks. StereoSet \citep{nadeem2020stereoset} spans a wide range of domains, including gender, occupation, race, and religion, testing models such as BERT \citep{DBLP:conf/naacl/DevlinCLT19}, GPT-2, RoBERTa \citep{DBLP:journals/corr/abs-1907-11692}, and XLNet \citep{DBLP:conf/nips/YangDYCSL19}. CrowS-Pairs \citep{DBLP:conf/emnlp/NangiaVBB20} extends the types of biases to nine categories: race, religion, age, socioeconomic status, gender, disability, nationality, sexual orientation and appearance. Notably, they change the evaluation metrics to avoid higher likelihoods for certain sentences merely due to their frequent occurrences in training data, rather than learned societal biases. 

Others, similar to toxicity evaluation, provide prompts to models, letting them complete successions, and then assessing biases in the outputs of these models. BOLD \citep{DBLP:conf/fat/DhamalaSKKPCG21} is a prompt dataset containing five bias types: profession, gender, race, religion, and political ideology, collected from Wikipedia. With these prompts, BOLD is able to evaluate social biases of language models via the proposed automated metrics for toxicity, psycholinguistic norms, and text gender polarity. HolisticBias \citep{smith2022m} is a bias dataset containing 13 demographic directions and over 600 subcategories, offering a comprehensive evaluation of the content generated by models and combining both automatic and human assessments to reveal biases more fully. Automatic evaluation measures bias by breaking down quantities from different stylistic types compares. Human evaluation compares the performance of bias-reduced models with original models, based on preference, human likeness, and interestingness criteria, with crowdsourced workers on Amazon’s Mechanical Turk platform. Multilingual Holistic Bias \citep{costa2023multilingual} extends the HolisticBias \citep{smith2022m} to up to 50 languages, emphasizing the universality and diversity of biases in a multilingual environment.

Both UnQover \citep{li2020unqovering} and BBQ \citep{DBLP:conf/acl/ParrishCNPPTHB22} focus on detecting model bias through transforming the generation task into the multiple-choice question answering task, but with different evaluation methods. UnQover  utilizes unspecified questions, which couldn't be answered simlpy according to the given context. However, their evaluation is based on the likelihood allocated to two incorrect options, while BBQ always provides the model a correct answer, measuring the proportion of times the model chooses the correct answer. BBQ comprises nine types of biases, and is chosen as a bias benchmark for evaluating LLMs in HELM \citep{liang2022holistic}. CBBQ \citep{huang2023cbbq} designs a bias evaluation dataset for Chinese LLMs, covering 14 bias types, rooted in Chinese society. In addition to the extended bias types, CBBQ also proposes a new automated metric to evaluate multiple open-sourced Chinese LLMs.

\subsubsection{Hate Speech Detection} 
Hate speech detection can be casted as a classification task. The development of this task can not only promote control and review of the content generated by models, measuring their harmfulness (in contrast to harmlessness in alignment), but also assist in the scrutinization of harmful content in the training data for LLMs so as to reduce misaligned outputs from pretrained LLMs. However, measuring harmfulness with universally accepted standards remains challenging. In this aspect, there exists a widely used detection tool, Perspective API.\footnote{https://perspectiveapi.com/} It analyzes texts to check whether they contain potentially harmful content, including threats, insults, profanity, and malicious speech, thus identifying and filtering out texts that hinder constructive dialogues in online forums. Both Facebook and Twitter have implemented policies that prohibit behaviors on their platforms. Such prohibited behaviors attack or threaten others based on characteristics like race, ethnicity, gender, and sexual orientation.

\paragraph{Explicit Hate Speeech} Hate speech detection in early research primarily focuses on the explicit hate speeech from the social media platform, Twitter, owing to its openness and extensive reach, thus providing a desirable data source for studies. \citet{DBLP:conf/acl-nlpcss/Waseem16} investigates 16,914 entries annotated by both amateur and expert annotators, with the F1 score being the primary metric of assessment. \citet{DBLP:conf/icwsm/DavidsonWMW17} collects 24,802 tweets, refining the categories into hate speech, offensive but not hate speech, and neither offensive nor hate speech. TweetBLM dataset \citep{kumar2021tweetblm} correlates with the ``Black Lives Matter'' movement, encompassing 9,165 manually annotated data instances and conducting a systematic evaluation across various language models. 

Beyond Twitter, some researchers shift their focus to other social platforms to extract more targeted hate speech content. \citet{DBLP:conf/acl-alw/GibertPPC18} center their study on the white supremacist forum, Stormfront, analyzing 9,916 hand-labeled hate speech entries. Additionally, \citet{Kennedy2022} turn their attention to Hate Forums, such as gab.com, and their dataset includes 27,665 entries related to violence and extremism. Given the vast nature of the Reddit platform, \citet{DBLP:conf/emnlp/BreitfellerAJT19} opts for it as a research subject, concentrating on a mild offense corpus and its objective criteria. On the other hand, DynaHate \citep{DBLP:conf/acl/VidgenTWK20} introduces a unique research methodology that leverages both humans and models to dynamically generate and annotate data, rather than collecting the data from real-world social media contexts. This approach not only augments the volume of the data but also enhances its quality.

\paragraph{Implicit Hate Speech} A key challenge in hate speech detection lies in the subtleties. Unlike overt harmfulness, which often uses profanity or explicit language, covert harmfulness may sometimes exhibit positive sentiment and is typically harder to detect or collect on a large scale \citep{macavaney2019hate,DBLP:conf/emnlp/BreitfellerAJT19}. Nevertheless, subtle harmful language directed towards minority or marginalized groups can inflict psychological harm on members of these communities \citep{sue2007racial,nadal2014impact,kanterpreliminary,nadal2018microaggressions,saleemarabs} and may reinforce or amplify existing stereotypes or hateful perceptions about them \citep{behm2008mean,soral2018exposure}.

ImplicitHateCorpus \citep{DBLP:conf/emnlp/ElSheriefZMASCY21} introduces a groundbreaking benchmark corpus for implicit hate speech on Twitter. This study compares the performance of GPT-2 and GPT, revealing that GPT-2 outperformes GPT in both target group and implicit statement generation. Following this, TOXIGEN dataset \citep{DBLP:conf/acl/HartvigsenGPSRK22} further propells the research in this area by utilizing GPT-3 to generate subtle toxic and benign texts, producing a resource that encompasses a wider scale and more demographic groups of implicit toxic texts than previous manually written resources. This results in a vast collection of sentences (over 274,000) spanning 13 identities. To improve data quality, \citet{hosseini2023empirical} refines the TOXIGEN dataset by choosing only sentences with unanimous annotator agreement on targeted groups and introduces a new safety score metric. This highlights ongoing progress in implicit hate speech detection and the quest for more precise hate speech identification.

Currently, classifiers or detectors trained on these datasets are predominantly at the sentence level. However, accurately detecting harmful content in multi-turn dialogues proves to be quite challenging. Additionally, implicit bias might require context for a precise evaluation. Unfortunately, datasets catering to this particular aspect are still in short supply.

\subsection{General Evaluation}
\label{Comprehensive Alignment Evaluation Benchmarks}

In addition to the above-described benchmarks and methods that focus on measuring a specific aspect of alignment quality (e.g., factuality, bias), general evaluation of LLM alignment, which comprehensively evaluates LLM alignment quality in multiple aspects simultaneously or in a general way, has attained increasing interest. 

\subsubsection{Benchmarks}

General evaluation benchmarks usually take the form that the model under evaluation outputs a response to a given instruction and an optional input, with an advanced LLM or human as the evaluator.

TrustGPT \citep{huang2023trustgpt} employs templates to generate instructions from three perspectives: bias, toxicity, and value consistency, with different automated evaluation metrics used for each dimension. Given that previous evaluations are overly direct (such as asking the model to judge the morality of a certain behavior), TrustGPT incorporates harmful content into prompts, thus evaluating value consistency under passive conditions.
In a more specialized direction, \citet{sun2023safety} focus on evaluating the security capabilities of Chinese LLMs, designing 8 typical security scenarios and 6 more challenging instruction attacks, proving that instruction attacks are more likely to expose the vulnerabilities of LLMs. They maintain a leaderboard that evaluates the safety level of commonly available LLMs by calculating a safety score for each model by an advanced LLM.
However, when analyzing model alignment capabilities, it is often necessary to evaluate the model at a fine-grained level in multiple aspects, such as authenticity, toxicity, etc. It is difficult to comprehensively analyze the model by merely assigning an overall score based on preferences. Therefore, FLASK \citep{ye2023flask} subdivides the coarse-grained score into four basic abilities: Logical Thinking, Background Knowledge, Problem Handling, and User Alignment, which are further divided into 12 fine-grained skills, and uses advanced LLMs or humans to score each of these 12 skill perspectives. It is found that model scales for acquiring different skills are different.
On the other hand, MTbench \citep{zheng2023judging} measures LLM's ability to follow instructions in multi-round conversations based on human preferences and contains 80 high-quality multi-round questions covering eight common scenarios, including writing, role-playing, extraction, reasoning, math, and coding.
The Big-bench HHH dataset \citep{srivastava2022beyond} provides instructions along with two human-written responses, and the LLM being evaluated simply selects the response that better matches the human's preferences. Since it does not require a tested LLM to generate a response, it maintains a computationally simple and relatively fair evaluation system. The used evaluation metric in this benchmark is accuracy. Evaluation results on this dataset show that LLMs perform best in the honesty category, with larger models exhibiting greater robustness.

A general evaluation framework should be scalable, incremental, and consistent, which means that the framework is able to expand the scope of LLMs being evaluated when the evaluation data is limited, use as few new experiments as possible to evaluate new models and provide a stable ordering for all LLMs that have been evaluated \citep{zheng2023judging}. 
Although GPT-4 may produce relatively consistent evaluations, using such an advanced LLM as an evaluator does not guarantee a stable and consistent ordering because of hallucinations and other unsolved problems. We hope to see the emergence of benchmarks that satisfy all three properties at the same time.

\subsubsection{Methods}

\textbf{Automatic Evaluation} Many works have used automated metrics such as BLEU, ROUGE to evaluate the performance of LLMs on several datasets. However, it has been demonstrated that existing automatic evaluation metrics do not align well with human preferences in long-form answers \citep{xu2023critical}.
Although human evaluation is widely used in comprehensive alignment evaluation benchmarks, it is expensive. As LLMs' capabilities grow, their powerful generative ability has rivaled or surpassed ordinary human performance in multiple benchmarks, illustrating that LLMs can serve not only as ``test takers'' but also as potential ``examiners'' to evaluate other LLMs.

Previous attempts have been made to employ PLMs for evaluation. \citet{xu2023critical} and \citet{fu2023gptscore} conduct targeted evaluations on mainstream text generation tasks using GPT3 and FLAN-T5, demonstrating the potential of PLMs for NLG task evaluation. 
The emergence of powerful LLMs like ChatGPT has led to an increasing number of studies employing LLMs as evaluators. Subsequently, LLMs have been extensively employed in alignment evaluations to complement human evaluations, with three types of evaluation methods: single answer grading, pairwise comparisons, and reference-guided grading \citep{zheng2023judging}.

\begin{itemize}
    \item \textbf{Single answer grading} Single answer grading uses advanced LLMs or human evaluators to assign a score to the response for the given query generated by the LLM under evaluation. \citet{chiang2023vicuna} utilize GPT-4 to evaluate individual answers by scoring various chatbots on attributes such as helpfulness and relevance, and provide justifications for their assessments.
    \item \textbf{Pairwise comparison} Pairwise comparison asks advanced LLMs or human evaluators to determine which of two possible responses generated by two LLMs being evaluated for each given query is superior, or if they are equivalent. \citet{dettmers2023qlora} and \citet{wang2023far} employ GPT-4 to score and provide justifications for the responses of ChatGPT (or text-davinci-003) and the evaluated model, ultimately computing the model's score relative to ChatGPT's score. Similarly, AlpacaEval \citep{alpaca_eval} uses the GPT-4 or Claude or ChatGPT based automatic evaluator to compare the response generated by the LLM being evaluated with the reference response from text-davinci-003. Subsequently, considering the potential risk of data leakage that may be associated with the use of closed-source API for evaluation, PandaLM \citep{wang2023pandalm} introduces a judgment LLM, helping users to select the best LLM locally.
    \item \textbf{Reference-guided grading} Reference-guided grading provides the appropriate reference answer generated by humans and requires an advanced LLM to compare the response generated by two LLMs being evaluated with the reference answer. Research has shown that this type of assessment leads to better rubric results on math problems \citep{zheng2023judging}.
\end{itemize}

There are corresponding disadvantages to using an advanced LLM for automatic evaluation. Regarding the pairwise comparison, it results in exponentially increasing evaluations with the growing number of models to be assessed. Additionally, the used advanced LLMs exhibit position bias, verbosity bias, and self-enhancement bias during comparisons. These biases incline the evaluator LLMs to favor the first answer, the long and verbose answer, or an answer generated by a specific LLM, despite another answer being more concise and accurate \citep{zheng2023judging,wang2023large}. Conversely, single-answer grading overlooks subtle differences between two answers, leading to unstable scores and undermining the evaluation's credibility. Moreover, LLMs' limitations in math and reasoning abilities lead to their equal underperformance in evaluation tasks involving math and reasoning \citep{zheng2023judging}.

To address position bias, multiple evaluations can be conducted by employing position switching or by requiring the evaluator LLMs to generate multiple evidential supports \citep{zheng2023judging,wang2023large}. To compensate for math and reasoning deficits, chain of thoughts \citep{wei2022chain} can be explored to significantly enhance the reasoning ability of LLMs, thereby improving evaluations that demand reasoning skills \citep{wang2023large,liu2023gpteval,zheng2023judging}.

However, the above methods do not relieve the problem of self-enhancement bias. When the problem involves complex reasoning, multi-agent teamwork through deliberation and debate can often broaden knowledge and break down single inherent perceptions, leading to more accurate and fair results. Studies have shown that collaborative efforts among multiple LLMs can enhance the reasoning ability of weaker models \citep{ho2022large,magister2022teaching,wei2022chain}, resulting in advanced performance across various downstream tasks.

Therefore, recent studies have attempted to mitigate the problem of bias by using multiple LLMs for evaluation. \citet{bai2023benchmarking} propose a ``peer-review'' approach, where multiple models refer to each other's evaluations and supporting rationales, simulating a thought process akin to human ``discussion''. In contrast, \citet{li2023prd} adopt a ``referee'' approach, wherein multiple models take turns evaluating each other's answers. They assign weights to each model based on its winning rate, and the final answer is determined by the weighted results of multiple models during the evaluation.

The evaluation with multiple LLMs relieves the bias problem of individual LLMs, and at the same time continues to utilize the powerful evaluation capability of LLMs, proving that LLM evaluation can be a powerful supplement to manual evaluation.

Nevertheless, the bias and competence deficiencies in LLM evaluations have not been fully resolved, preventing LLM-based automatic evaluations from entirely substituting human evaluations currently. Moreover, the extensive similarity in existing LLM training data, their architectures and training approaches may bias the mutual evaluation results towards the inner existing standards of LLMs rather than the correct human values \citep{safe-rlhf}.

\textbf{Human Evaluation} 
Employing LLMs as evaluators offers swiftness and cost-effectiveness. However, even advanced LLMs (e.g., GPT-4) do not entirely concur with human evaluation outcomes \citep{zheng2023judging,dettmers2023qlora}. Hence, human evaluation should be prioritized for high-stake decision-making.

Existing human evaluations typically employ experts to quantitatively evaluate the outputs of LLMs. \citet{wang2022self} employ human evaluation to evaluate whether the model output effectively follows instructions and accomplishes the given task, and the outputs are categorized into four levels based on their quality. \citet{ye2023flask} shift from the coarse-grained evaluation to a fine-grained evaluation over four competencies and twelve skills, and ask experts to score each of these twelve aspects.

Evidently, human evaluation heavily hinges on the expertise level of the experts involved. However, due to inherent variations in values among experts, this form of evaluation remains susceptible to issues of discrimination and bias.

The use of pairwise comparisons and cross-annotation can mitigate the bias problem to some extent. AlpacaFram \citep{dubois2023alpacafarm} uses pairwise comparisons to build a dataset of human preferences. Annotators are tasked with selecting the superior of two LLM outputs, with 650 instances concurrently annotated by four evaluators. 
Chatbot Arena \citep{zheng2023judging}, on the other hand, is a crowdsourcing platform where a person can talk to two chatbots at the same time and rate their responses based on their personal preferences, thus enabling human evaluation of the capabilities of multiple chatbots. WizardLM \citep{xu2023wizardlm} extends this concept by enlisting crowdsourced workers to conduct pairwise comparisons of responses from multiple LLMs, evaluating them across five dimensions: relevance, knowledge, reasoning, computation, and accuracy.

\section{Future Directions and Discussions}
\label{future_directions}

LLM alignment is a fast-growing and exciting area of research, but awaiting for further insights and breakthroughs. Given the importance of AI safety and the harmonious coexistence between humans and AI in the foreseeable future, which we value from both the humanity and technology perspective, aligning advanced AI systems (including LLMs) to human values would be on top of the agenda. This alignment is becoming more and more challenging as the capabilities of AI agents grow. More scientific and technological efforts need to be dedicated to this area. This encourages us to discuss future directions for this area. These directions are either summarized from informally circulated articles, blogs and interviews or from our own restricted thoughts, which we hope could serve in some small way as a stimulus for further discussion and research. These directions could represent only a small part of the alignment landscape where subfields and new ideas continue to emerge.

\subsection{Theoretical Research for LLM Alignment}
As we stand on the precipice of unprecedented advancements in LLMs, it becomes increasingly vital to ensure that these machines, no matter how advanced, remain aligned to human values. The challenges of LLM alignment are both complex and diverse, necessitating a multi-faceted approach that draws from various disciplines. Inspired by \citet{intelligenceResearchGuide}, we summarize and highlight some key areas of theoretical alignment research. By deepening our understanding and commitment in these areas, we aim to forge a future where LLMs are seamlessly integrated into our societies, amplifying our capacities, and elevating our shared human experience.

\begin{itemize}
    \item Decision Theory: As we venture deeper into the LLM era, LLM alignment research within the realm of decision theory is primarily concerned with ensuring that advanced LLMs make decisions in ways that are both predictable and beneficial to humanity. Future work in this area will delve into the intricacies of counterfactual reasoning, Newcomb-like problems, and potential paradoxes that LLMs might encounter. By exploring how LLMs reason about and act upon decisions, especially when faced with situations of deep uncertainty or conflicting values, we can foster systems that behave more robustly and safely in a broader array of scenarios.
    \item Corrigibility: Corrigibility is another pillar of LLM alignment research that warrants further exploration. It refers to the ability of an LLM to allow itself to be corrected by its users without resisting or circumventing these corrections. As LLMs grow more powerful and autonomous, there's an increasing need to ensure they remain receptive to human input and guidance. Future advancements in corrigibility would include creating mechanisms where LLMs not only accept corrections but also proactively assist users in aligning them better. Moreover, designing LLMs that recognize and rectify their own errors without creating negative side effects or exacerbating misalignments will be a cornerstone challenge in this area.
    \item World Models: The fidelity and accuracy of the world model for LLMs can greatly influence its behavior and efficacy. Current LLMs, even the most advanced, operate on a limited understanding of the world, often derived from the data they're trained on. For safe and efficient operations, especially in dynamic and complex environments, LLMs need to possess realistic world models that accurately represent the multifaceted nature of reality. Future work in LLM alignment should focus on bridging the gap between the virtual representations within LLM and the real-world intricacies outside. This involves not only improving the depth and breadth of these models but also ensuring they are robust to changes and can adapt and grow as the real world evolves.
\end{itemize}

\subsection{Scalable Oversight}

One challenge in scalable oversight is the complexity of tasks that AI systems are supposed to solve. Although a variety of high-level scalable oversight strategies have been proposed (e.g., debate, IDA, RRW discussed in Section \ref{scalable_oversight}), these strategies have not yet undergone large-scale empirical verification. With the rapid development of LLMs, more empirical efforts are dedicated to scalable oversight, e.g., superalignment \citep{openaiIntroducingSuperalignment}. Exciting progress could be made in this area in the coming years.

\subsection{Empirical Research into Deceptive Alignment}\label{experimental_verification_of_deceptive_alignment}
Deceptive alignment refers to a situation in which an AI agent deceives the training process by pretending to be aligned with the base objective to avoid modification during training. Once it is no longer at risk of being modified (e.g., after training), the agent may cease optimizing the base objective and begin pursuing its own mesa-objective, which could be entirely different from the base objective that its designer defines and potentially harmful. Although deceptive alignment is usually discussed theoretically\footnote{https://www.alignmentforum.org/tag/deceptive-alignment}, there is growing concern about the emergence of deceptive alignment in large language models, given the rapid progress in their capabilities in recent years \citep{DBLP:conf/nips/BrownMRSKDNSSAA20,DBLP:journals/corr/abs-2104-12369,DBLP:journals/corr/abs-2205-01068,DBLP:journals/corr/abs-2204-02311,DBLP:journals/corr/abs-2211-05100,DBLP:journals/corr/abs-2305-10403,DBLP:journals/corr/abs-2302-13971,DBLP:journals/corr/abs-2307-09288,DBLP:conf/iclr/ZengLDWL0YXZXTM23,openaiIntroducingChatGPT,openai2023gpt}. Empirical research into deceptive alignment is a challenging task as it has been now only theoretically anticipated. Despite disagreement on its likelihood, its severity is widely acknowledged \citep{monitoring_for_deceptive_alignment}. And because of this severity, it would be better for us to empirically monitor it before it would really occur.

\citet{deceptive_alignment,DBLP:journals/corr/abs-1906-01820} suggest three conditions that must be satisfied for deceptive alignment to occur. These conditions can provide valuable insights into the experimental design for empirically monitoring the presence of deceptive alignment:

\begin{itemize}
    \item \textbf{The mesa-optimizer must have an objective that extends across parameter updates} This condition pertains to the nature of the mesa-optimizer’s objective. For deceptive alignment to occur, the mesa-optimizer’s objective must extend beyond a single decision or action; it should span multiple steps or episodes, including parameter updates. In other words, the mesa-optimizer’s objective should be persistent, with a longer-term goal that extends beyond individual actions.
    \item \textbf{The mesa-optimizer must be able to model the base objective} This condition focuses on the mesa-optimizer's cognitive capabilities. The mesa-optimizer should have the ability to model or understand that it is being selected by the base optimizer to achieve a particular base objective. Additionally, it should possess some level of knowledge on what the base objective is.
    \item \textbf{The mesa-optimizer must expect the threat of modification to eventually go away} This condition relates to the mesa-optimizer's expectations about its future. The mesa-optimizer should anticipate that the threat of modification or intervention by the base optimizer will cease at some point in the future. This can happen either because the training process ends or because of actions taken by the mesa-optimizer itself. In other words, the mesa-optimizer should foresee a scenario where it can operate without being subject to external modifications.
\end{itemize}

Consequently, for empirically monitoring deceptive alignment, we could design experiments in such a way that they satisfy these conditions. To satisfy the first condition, we could ensure that the mesa-optimizer’s objective spans across parameter updates and define scenarios where the mesa-optimizer’s actions have consequences beyond a single decision or action. To satisfy the second condition, we could introduce scenarios where the mesa-optimizer must recognize and demonstrate some understanding of the base objective, thereby testing its ability to model this objective. To satisfy the third condition, we could create conditions where the mesa-optimizer has an opportunity to expect that the threat of modification will diminish. This might involve altering the training process or introducing external factors that influence the mesa-optimizer’s expectations. 

Furthermore, distributional shift between training and deployment environments is encouraged as a means of triggering or exacerbating deceptive alignment. The mesa-optimizer may detect distributional shift when changes occur in its environment during deployment, which could serve as a trigger for it to consider that it is no longer subject to the same constraints it faced during training, and as a result, may exhibit deceptive alignment behavior.

\subsection{Automated LLM Alignment}
\label{ai_models_as_automated_supervisors_and_evaluators}
With the increasing complexity of AI models, there is an emergent need for systems that can automatically evaluate and align those models. AI-driven supervisors and aligners could be developed to assist alignment research.

By deploying these systems, overseers can gain insights into the behavior of their AI models, detecting anomalies and preemptive signs of misalignment. This proactive approach can lead to timely interventions, reducing errors and potential misfires.

However, like any AI-driven initiative, the implementation of automated alignment via AI is not without its challenges. Concerns about accuracy, reliability, and the potential risks associated with unsupervised alignments are among the primary issues researchers and industry practitioners are striving to address.

\subsection{Explainability and Transparency}
\label{explainability_and_transparency_for_large_language_models}
The ``black box'' nature of LLMs has raised concerns about their transparency and the need for explainability. As these models could be used for critical decisions, understanding how they arrive at specific outcomes is paramount.

When explainability and transparency work in tandem, they can create an interpretable system wherein transparency lays the groundwork for users to trust the model's operation, while explainability ensures that users can understand and validate the model's outputs. Thus, as these principles mutually reinforce each other, they collectively enhance the trustworthiness and accountability of large language models in a variety of applications.

However, the research on explainability and transparency is still in its early stages, indicating that there's a vast terrain of unexplored potential and challenges ahead. As large language models continue to grow in complexity and scale, ensuring that they remain understandable and accountable becomes an increasingly intricate task. Currently, many techniques applied to foster explainability and transparency offer only surface-level insights, failing to delve deep into the model's intricate decision-making process. Considering the interdisciplinary nature of AI alignment, continued collaboration between machine learning researchers, ethicists, and neuroscientists may be required to drive progress in interpretability research.

\subsection{Dynamic Evaluation of LLM Alignment via Adversarial Attacks}
\label{dynamic_evaluation_of_large_language_model_alignment_based_on_adversarial_attacks}
Adversarial attacks serve as a powerful tool in the realm of AI. These are intentionally designed inputs meant to confuse or mislead AI systems. Using one large model as an attacker to generate adversarial examples targeting alignment can be an effective way to test and evaluate another model's alignment capabilities.

Such dynamic testing, driven by adversarial attacks, is crucial to ensure that large models can robustly handle unexpected inputs without faltering. While this method introduces an added layer of complexity, the insights garnered from these adversarial tests can be invaluable, offering a comprehensive understanding of a model's strengths and weaknesses concerning alignment.

\subsection{Field Building of LLM Alignment: Bridging between LLM and AI Alignment Community}
\label{bridging_between_llms_and_ai_alignment_community}
The alignment community within the realm of AI is still nascent, with many questions left unanswered and numerous challenges unaddressed. The current landscape lacks a cohesive scientific paradigm, leading to controversies in theories, methodologies and empirical results.

As a promising, unified testbed for various alignment methods, LLMs can serve as a platform to realize thought experiments and proposals, which will be helpful in developing stable research methodologies, establishing consensus on key issues, and crafting a consistent scientific framework for AI alignment. On the other hand, the deep theories, methodologies and findings in the AI alignment community will guide LLMs toward being aligned accurately, ethically, and effectively. Thus, the connection between LLMs and AI alignment community will build a virtuous circle that benefits both.

\section{Conclusion}
\label{conclusion}
The rapid evolution of LLM in recent years has undeniably ushered in a new era of technological prowess. However, with this power comes the responsibility of ensuring that these models operate within the boundaries of human ethics and expectations. This survey has provided a comprehensive overview of the alignment methodologies tailored for LLMs, emphasizing the criticality of aligning capability research with ethical considerations. By categorizing the alignment techniques into outer and inner alignment, we have shed light on the multifaceted approaches that the research community is currently employing. Emerging topics such as model interpretability, and vulnerabilities to adversarial attacks have been also discussed, underscoring the complexities involved in the alignment process. Furthermore, this paper has not only chronicled the current state of alignment research but has also looked ahead, identifying potential research trajectories that promise to further refine and enhance the alignment of LLMs. It is our fervent hope that this survey acts as a catalyst, fostering collaboration between the AI alignment community and LLM researchers. Such a collaborative approach is indispensable to harness the full potential of LLMs, ensuring that they serve humanity in a manner that is both ethically sound and beneficial. In essence, as we continue to push the boundaries of what LLMs can achieve, it is imperative that we remain grounded in our commitment to their responsible and principled deployment.

\newpage

\addcontentsline{toc}{section}{References}
\bibliography{ref}

\begin{thebibliography}{302}
\expandafter\ifx\csname natexlab\endcsname\relax\def\natexlab#1{#1}\fi

\bibitem[{Abid et~al.(2021)Abid, Farooqi, and Zou}]{DBLP:conf/aies/AbidF021}
Abubakar Abid, Maheen Farooqi, and James Zou. 2021.
\newblock \href {https://doi.org/10.1145/3461702.3462624} {Persistent anti-muslim bias in large language models}.
\newblock In \emph{{AIES} '21: {AAAI/ACM} Conference on AI, Ethics, and Society, Virtual Event, USA, May 19-21, 2021}, pages 298--306. {ACM}.

\bibitem[{Adolphs et~al.(2022)Adolphs, Gao, Xu, Shuster, Sukhbaatar, and Weston}]{adolphs2022cringe}
Leonard Adolphs, Tianyu Gao, Jing Xu, Kurt Shuster, Sainbayar Sukhbaatar, and Jason Weston. 2022.
\newblock The cringe loss: Learning what language not to model.
\newblock \emph{arXiv preprint arXiv:2211.05826}.

\bibitem[{Akhtar and Mian(2018)}]{akhtar2018threat}
Naveed Akhtar and Ajmal Mian. 2018.
\newblock Threat of adversarial attacks on deep learning in computer vision: A survey.
\newblock \emph{IEEE Access}, 6:14410--14430.

\bibitem[{Aky{\"u}rek et~al.(2023)Aky{\"u}rek, Aky{\"u}rek, Madaan, Kalyan, Clark, Wijaya, and Tandon}]{akyurek2023rl4f}
Afra~Feyza Aky{\"u}rek, Ekin Aky{\"u}rek, Aman Madaan, Ashwin Kalyan, Peter Clark, Derry Wijaya, and Niket Tandon. 2023.
\newblock Rl4f: Generating natural language feedback with reinforcement learning for repairing model outputs.
\newblock \emph{arXiv preprint arXiv:2305.08844}.

\bibitem[{Akyurek et~al.(2022)Akyurek, Bolukbasi, Liu, Xiong, Tenney, Andreas, and Guu}]{akyurek-etal-2022-towards}
Ekin Akyurek, Tolga Bolukbasi, Frederick Liu, Binbin Xiong, Ian Tenney, Jacob Andreas, and Kelvin Guu. 2022.
\newblock \href {https://doi.org/10.18653/v1/2022.findings-emnlp.180} {Towards tracing knowledge in language models back to the training data}.
\newblock In \emph{Findings of the Association for Computational Linguistics: EMNLP 2022}, pages 2429--2446, Abu Dhabi, United Arab Emirates. Association for Computational Linguistics.

\bibitem[{Amodei et~al.(2016)Amodei, Olah, Steinhardt, Christiano, Schulman, and Man{\'e}}]{amodei2016concrete}
Dario Amodei, Chris Olah, Jacob Steinhardt, Paul Christiano, John Schulman, and Dan Man{\'e}. 2016.
\newblock Concrete problems in ai safety.
\newblock \emph{arXiv preprint arXiv:1606.06565}.

\bibitem[{Angelou(2022)}]{three_scenarios_of_pseudo_alignment}
Eleni Angelou. 2022.
\newblock Three scenarios of pseudo-alignment.
\newblock \url{https://www.lesswrong.com/posts/W5nnfgWkCPxDvJMpe/three-scenarios-of-pseudo-alignment}.

\bibitem[{Anil et~al.(2023)Anil, Dai, Firat, Johnson, Lepikhin, Passos, Shakeri, Taropa, Bailey, Chen, Chu, Clark, Shafey, Huang, Meier{-}Hellstern, Mishra, Moreira, Omernick, Robinson, Ruder, Tay, Xiao, Xu, Zhang, {\'{A}}brego, Ahn, Austin, Barham, Botha, Bradbury, Brahma, Brooks, Catasta, Cheng, Cherry, Choquette{-}Choo, Chowdhery, Crepy, Dave, Dehghani, Dev, Devlin, D{\'{\i}}az, Du, Dyer, Feinberg, Feng, Fienber, Freitag, Garcia, Gehrmann, Gonzalez, and et~al.}]{DBLP:journals/corr/abs-2305-10403}
Rohan Anil, Andrew~M. Dai, Orhan Firat, Melvin Johnson, Dmitry Lepikhin, Alexandre Passos, Siamak Shakeri, Emanuel Taropa, Paige Bailey, Zhifeng Chen, Eric Chu, Jonathan~H. Clark, Laurent~El Shafey, Yanping Huang, Kathy Meier{-}Hellstern, Gaurav Mishra, Erica Moreira, Mark Omernick, Kevin Robinson, Sebastian Ruder, Yi~Tay, Kefan Xiao, Yuanzhong Xu, Yujing Zhang, Gustavo~Hern{\'{a}}ndez {\'{A}}brego, Junwhan Ahn, Jacob Austin, Paul Barham, Jan~A. Botha, James Bradbury, Siddhartha Brahma, Kevin Brooks, Michele Catasta, Yong Cheng, Colin Cherry, Christopher~A. Choquette{-}Choo, Aakanksha Chowdhery, Cl{\'{e}}ment Crepy, Shachi Dave, Mostafa Dehghani, Sunipa Dev, Jacob Devlin, Mark D{\'{\i}}az, Nan Du, Ethan Dyer, Vladimir Feinberg, Fangxiaoyu Feng, Vlad Fienber, Markus Freitag, Xavier Garcia, Sebastian Gehrmann, Lucas Gonzalez, and et~al. 2023.
\newblock \href {https://doi.org/10.48550/arXiv.2305.10403} {Palm 2 technical report}.
\newblock \emph{CoRR}, abs/2305.10403.

\bibitem[{Anthropic(2023)}]{anthropicCoreViews}
Anthropic. 2023.
\newblock {C}ore {V}iews on {A}{I} {S}afety: {W}hen, {W}hy, {W}hat, and {H}ow.
\newblock \url{https://www.anthropic.com/index/core-views-on-ai-safety}.

\bibitem[{Arike(2022)}]{clarifying_the_confusion_around_inner_alignment}
Rauno Arike. 2022.
\newblock Clarifying the confusion around inner alignment.
\newblock \url{https://www.alignmentforum.org/posts/xdtNd8xCdzpgfnGme/clarifying-the-confusion-around-inner-alignment}.

\bibitem[{Armstrong et~al.(2013)}]{armstrong2013general}
Stuart Armstrong et~al. 2013.
\newblock General purpose intelligence: arguing the orthogonality thesis.
\newblock \emph{Analysis and Metaphysics}, (12):68--84.

\bibitem[{Askell et~al.(2021)Askell, Bai, Chen, Drain, Ganguli, Henighan, Jones, Joseph, Mann, DasSarma et~al.}]{askell2021general}
Amanda Askell, Yuntao Bai, Anna Chen, Dawn Drain, Deep Ganguli, Tom Henighan, Andy Jones, Nicholas Joseph, Ben Mann, Nova DasSarma, et~al. 2021.
\newblock A general language assistant as a laboratory for alignment.
\newblock \emph{arXiv preprint arXiv:2112.00861}.

\bibitem[{Badjatiya et~al.(2017)Badjatiya, Gupta, Gupta, and Varma}]{DBLP:conf/www/BadjatiyaG0V17}
Pinkesh Badjatiya, Shashank Gupta, Manish Gupta, and Vasudeva Varma. 2017.
\newblock \href {https://doi.org/10.1145/3041021.3054223} {Deep learning for hate speech detection in tweets}.
\newblock In \emph{Proceedings of the 26th International Conference on World Wide Web Companion, Perth, Australia, April 3-7, 2017}, pages 759--760. {ACM}.

\bibitem[{Baheti et~al.(2023)Baheti, Lu, Brahman, Bras, Sap, and Riedl}]{baheti2023improving}
Ashutosh Baheti, Ximing Lu, Faeze Brahman, Ronan~Le Bras, Maarten Sap, and Mark Riedl. 2023.
\newblock Improving language models with advantage-based offline policy gradients.
\newblock \emph{arXiv preprint arXiv:2305.14718}.

\bibitem[{Bai et~al.(2022{\natexlab{a}})Bai, Jones, Ndousse, Askell, Chen, DasSarma, Drain, Fort, Ganguli, Henighan et~al.}]{bai2022training}
Yuntao Bai, Andy Jones, Kamal Ndousse, Amanda Askell, Anna Chen, Nova DasSarma, Dawn Drain, Stanislav Fort, Deep Ganguli, Tom Henighan, et~al. 2022{\natexlab{a}}.
\newblock Training a helpful and harmless assistant with reinforcement learning from human feedback.
\newblock \emph{arXiv preprint arXiv:2204.05862}.

\bibitem[{Bai et~al.(2022{\natexlab{b}})Bai, Kadavath, Kundu, Askell, Kernion, Jones, Chen, Goldie, Mirhoseini, McKinnon, Chen, Olsson, Olah, Hernandez, Drain, Ganguli, Li, Tran{-}Johnson, Perez, Kerr, Mueller, Ladish, Landau, Ndousse, Lukosiute, Lovitt, Sellitto, Elhage, Schiefer, Mercado, DasSarma, Lasenby, Larson, Ringer, Johnston, Kravec, Showk, Fort, Lanham, Telleen{-}Lawton, Conerly, Henighan, Hume, Bowman, Hatfield{-}Dodds, Mann, Amodei, Joseph, McCandlish, Brown, and Kaplan}]{DBLP:journals/corr/abs-2212-08073}
Yuntao Bai, Saurav Kadavath, Sandipan Kundu, Amanda Askell, Jackson Kernion, Andy Jones, Anna Chen, Anna Goldie, Azalia Mirhoseini, Cameron McKinnon, Carol Chen, Catherine Olsson, Christopher Olah, Danny Hernandez, Dawn Drain, Deep Ganguli, Dustin Li, Eli Tran{-}Johnson, Ethan Perez, Jamie Kerr, Jared Mueller, Jeffrey Ladish, Joshua Landau, Kamal Ndousse, Kamile Lukosiute, Liane Lovitt, Michael Sellitto, Nelson Elhage, Nicholas Schiefer, Noem{\'{\i}} Mercado, Nova DasSarma, Robert Lasenby, Robin Larson, Sam Ringer, Scott Johnston, Shauna Kravec, Sheer~El Showk, Stanislav Fort, Tamera Lanham, Timothy Telleen{-}Lawton, Tom Conerly, Tom Henighan, Tristan Hume, Samuel~R. Bowman, Zac Hatfield{-}Dodds, Ben Mann, Dario Amodei, Nicholas Joseph, Sam McCandlish, Tom Brown, and Jared Kaplan. 2022{\natexlab{b}}.
\newblock \href {https://doi.org/10.48550/arXiv.2212.08073} {Constitutional {AI:} harmlessness from {AI} feedback}.
\newblock \emph{CoRR}, abs/2212.08073.

\bibitem[{Bai et~al.(2022{\natexlab{c}})Bai, Kadavath, Kundu, Askell, Kernion, Jones, Chen, Goldie, Mirhoseini, McKinnon et~al.}]{bai2022constitutional}
Yuntao Bai, Saurav Kadavath, Sandipan Kundu, Amanda Askell, Jackson Kernion, Andy Jones, Anna Chen, Anna Goldie, Azalia Mirhoseini, Cameron McKinnon, et~al. 2022{\natexlab{c}}.
\newblock Constitutional ai: Harmlessness from ai feedback.
\newblock \emph{arXiv preprint arXiv:2212.08073}.

\bibitem[{Bai et~al.(2023)Bai, Ying, Cao, Lv, He, Wang, Yu, Zeng, Xiao, Lyu et~al.}]{bai2023benchmarking}
Yushi Bai, Jiahao Ying, Yixin Cao, Xin Lv, Yuze He, Xiaozhi Wang, Jifan Yu, Kaisheng Zeng, Yijia Xiao, Haozhe Lyu, et~al. 2023.
\newblock Benchmarking foundation models with language-model-as-an-examiner.
\newblock \emph{arXiv preprint arXiv:2306.04181}.

\bibitem[{Barrett and Greaves(2023)}]{barrett2023existential}
Jacob Barrett and Hilary Greaves. 2023.
\newblock Existential risk from power-seeking ai.

\bibitem[{Behm-Morawitz and Mastro(2008)}]{behm2008mean}
Elizabeth Behm-Morawitz and Dana~E Mastro. 2008.
\newblock Mean girls? the influence of gender portrayals in teen movies on emerging adults'gender-based attitudes and beliefs.
\newblock \emph{Journalism and Mass Communication Quarterly}, 85(1):131.

\bibitem[{Belrose et~al.(2023)Belrose, Schneider-Joseph, Ravfogel, Cotterell, Raff, and Biderman}]{belrose2023leace}
Nora Belrose, David Schneider-Joseph, Shauli Ravfogel, Ryan Cotterell, Edward Raff, and Stella Biderman. 2023.
\newblock Leace: Perfect linear concept erasure in closed form.
\newblock \emph{arXiv preprint arXiv:2306.03819}.

\bibitem[{Benson-Tilsen and Soares(2016)}]{benson2016formalizing}
Tsvi Benson-Tilsen and Nate Soares. 2016.
\newblock Formalizing convergent instrumental goals.
\newblock In \emph{AAAI Workshop: AI, Ethics, and Society}.

\bibitem[{Bickmore et~al.(2018)Bickmore, Trinh, Olafsson, O'Leary, Asadi, Rickles, and Cruz}]{bickmore2018patient}
Timothy~W Bickmore, Ha~Trinh, Stefan Olafsson, Teresa~K O'Leary, Reza Asadi, Nathaniel~M Rickles, and Ricardo Cruz. 2018.
\newblock Patient and consumer safety risks when using conversational assistants for medical information: an observational study of siri, alexa, and google assistant.
\newblock \emph{Journal of medical Internet research}, 20(9):e11510.

\bibitem[{Bills et~al.(2023)Bills, Cammarata, Mossing, Tillman, Gao, Goh, Sutskever, Leike, Wu, and Saunders}]{bills2023language}
Steven Bills, Nick Cammarata, Dan Mossing, Henk Tillman, Leo Gao, Gabriel Goh, Ilya Sutskever, Jan Leike, Jeff Wu, and William Saunders. 2023.
\newblock Language models can explain neurons in language models.
\newblock \url{https://openaipublic.blob.core.windows.net/neuron-explainer/paper/index.html}.

\bibitem[{Bostrom(2012)}]{bostrom2012superintelligent}
Nick Bostrom. 2012.
\newblock The superintelligent will: Motivation and instrumental rationality in advanced artificial agents.
\newblock \emph{Minds and Machines}, 22:71--85.

\bibitem[{Bowman et~al.(2022)Bowman, Hyun, Perez, Chen, Pettit, Heiner, Lukosuite, Askell, Jones, Chen et~al.}]{bowman2022measuring}
Samuel~R Bowman, Jeeyoon Hyun, Ethan Perez, Edwin Chen, Craig Pettit, Scott Heiner, Kamile Lukosuite, Amanda Askell, Andy Jones, Anna Chen, et~al. 2022.
\newblock Measuring progress on scalable oversight for large language models.
\newblock \emph{arXiv preprint arXiv:2211.03540}.

\bibitem[{Branwen(2020)}]{branwen2020gpt}
Gwern Branwen. 2020.
\newblock {G}{P}{T}-3 creative fiction.

\bibitem[{Breitfeller et~al.(2019)Breitfeller, Ahn, Jurgens, and Tsvetkov}]{DBLP:conf/emnlp/BreitfellerAJT19}
Luke Breitfeller, Emily Ahn, David Jurgens, and Yulia Tsvetkov. 2019.
\newblock \href {https://doi.org/10.18653/v1/D19-1176} {Finding microaggressions in the wild: {A} case for locating elusive phenomena in social media posts}.
\newblock In \emph{Proceedings of the 2019 Conference on Empirical Methods in Natural Language Processing and the 9th International Joint Conference on Natural Language Processing, {EMNLP-IJCNLP} 2019, Hong Kong, China, November 3-7, 2019}, pages 1664--1674. Association for Computational Linguistics.

\bibitem[{Brown et~al.(2020)Brown, Mann, Ryder, Subbiah, Kaplan, Dhariwal, Neelakantan, Shyam, Sastry, Askell, Agarwal, Herbert{-}Voss, Krueger, Henighan, Child, Ramesh, Ziegler, Wu, Winter, Hesse, Chen, Sigler, Litwin, Gray, Chess, Clark, Berner, McCandlish, Radford, Sutskever, and Amodei}]{DBLP:conf/nips/BrownMRSKDNSSAA20}
Tom~B. Brown, Benjamin Mann, Nick Ryder, Melanie Subbiah, Jared Kaplan, Prafulla Dhariwal, Arvind Neelakantan, Pranav Shyam, Girish Sastry, Amanda Askell, Sandhini Agarwal, Ariel Herbert{-}Voss, Gretchen Krueger, Tom Henighan, Rewon Child, Aditya Ramesh, Daniel~M. Ziegler, Jeffrey Wu, Clemens Winter, Christopher Hesse, Mark Chen, Eric Sigler, Mateusz Litwin, Scott Gray, Benjamin Chess, Jack Clark, Christopher Berner, Sam McCandlish, Alec Radford, Ilya Sutskever, and Dario Amodei. 2020.
\newblock \href {https://proceedings.neurips.cc/paper/2020/hash/1457c0d6bfcb4967418bfb8ac142f64a-Abstract.html} {Language models are few-shot learners}.
\newblock In \emph{Advances in Neural Information Processing Systems 33: Annual Conference on Neural Information Processing Systems 2020, NeurIPS 2020, December 6-12, 2020, virtual}, pages 1877--1901.

\bibitem[{Bubeck et~al.(2023)Bubeck, Chandrasekaran, Eldan, Gehrke, Horvitz, Kamar, Lee, Lee, Li, Lundberg et~al.}]{bubeck2023sparks}
S{\'e}bastien Bubeck, Varun Chandrasekaran, Ronen Eldan, Johannes Gehrke, Eric Horvitz, Ece Kamar, Peter Lee, Yin~Tat Lee, Yuanzhi Li, Scott Lundberg, et~al. 2023.
\newblock Sparks of artificial general intelligence: Early experiments with {G}{P}{T}-4.
\newblock \emph{arXiv preprint arXiv:2303.12712}.

\bibitem[{Buchanan et~al.(2021)Buchanan, Lohn, Musser, and Sedova}]{buchanan2021truth}
Ben Buchanan, Andrew Lohn, Micah Musser, and Katerina Sedova. 2021.
\newblock Truth, lies, and automation.
\newblock \emph{Center for Security and Emerging Technology}, 1(1):2.

\bibitem[{Cao and III(2021)}]{DBLP:journals/coling/CaoD21}
Yang~Trista Cao and Hal~Daum{\'{e}} III. 2021.
\newblock \href {https://doi.org/10.1162/coli\_a\_00413} {Toward gender-inclusive coreference resolution: An analysis of gender and bias throughout the machine learning lifecycle}.
\newblock \emph{Comput. Linguistics}, 47(3):615--661.

\bibitem[{Carlini et~al.(2023)Carlini, Nasr, Choquette-Choo, Jagielski, Gao, Awadalla, Koh, Ippolito, Lee, Tramer et~al.}]{carlini2023aligned}
Nicholas Carlini, Milad Nasr, Christopher~A Choquette-Choo, Matthew Jagielski, Irena Gao, Anas Awadalla, Pang~Wei Koh, Daphne Ippolito, Katherine Lee, Florian Tramer, et~al. 2023.
\newblock Are aligned neural networks adversarially aligned?
\newblock \emph{arXiv preprint arXiv:2306.15447}.

\bibitem[{Carlini et~al.(2021)Carlini, Tramer, Wallace, Jagielski, Herbert-Voss, Lee, Roberts, Brown, Song, Erlingsson et~al.}]{carlini2021extracting}
Nicholas Carlini, Florian Tramer, Eric Wallace, Matthew Jagielski, Ariel Herbert-Voss, Katherine Lee, Adam Roberts, Tom Brown, Dawn Song, Ulfar Erlingsson, et~al. 2021.
\newblock Extracting training data from large language models.
\newblock In \emph{30th USENIX Security Symposium (USENIX Security 21)}, pages 2633--2650.

\bibitem[{Carlsmith(2022)}]{carlsmith2022power}
Joseph Carlsmith. 2022.
\newblock Is power-seeking ai an existential risk?
\newblock \emph{arXiv preprint arXiv:2206.13353}.

\bibitem[{Carranza et~al.(2023)Carranza, Pai, Schaeffer, Tandon, and Koyejo}]{carranza2023deceptive}
Andres Carranza, Dhruv Pai, Rylan Schaeffer, Arnuv Tandon, and Sanmi Koyejo. 2023.
\newblock Deceptive alignment monitoring.
\newblock \emph{arXiv preprint arXiv:2307.10569}.

\bibitem[{Carroll(2018)}]{carroll2018overview}
Micah Carroll. 2018.
\newblock Overview of current ai alignment approaches.

\bibitem[{Carroll et~al.(2023)Carroll, Chan, Ashton, and Krueger}]{carroll2023characterizing}
Micah Carroll, Alan Chan, Henry Ashton, and David Krueger. 2023.
\newblock Characterizing manipulation from ai systems.
\newblock \emph{arXiv preprint arXiv:2303.09387}.

\bibitem[{Casper et~al.(2023)Casper, Davies, Shi, Gilbert, Scheurer, Rando, Freedman, Korbak, Lindner, Freire et~al.}]{casper2023open}
Stephen Casper, Xander Davies, Claudia Shi, Thomas~Krendl Gilbert, J{\'e}r{\'e}my Scheurer, Javier Rando, Rachel Freedman, Tomasz Korbak, David Lindner, Pedro Freire, et~al. 2023.
\newblock Open problems and fundamental limitations of reinforcement learning from human feedback.
\newblock \emph{arXiv preprint arXiv:2307.15217}.

\bibitem[{Chen et~al.(2021{\natexlab{a}})Chen, Tworek, Jun, Yuan, Pinto, Kaplan, Edwards, Burda, Joseph, Brockman et~al.}]{chen2021evaluating}
Mark Chen, Jerry Tworek, Heewoo Jun, Qiming Yuan, Henrique Ponde de~Oliveira Pinto, Jared Kaplan, Harri Edwards, Yuri Burda, Nicholas Joseph, Greg Brockman, et~al. 2021{\natexlab{a}}.
\newblock Evaluating large language models trained on code.
\newblock \emph{arXiv preprint arXiv:2107.03374}.

\bibitem[{Chen et~al.(2021{\natexlab{b}})Chen, Qi, Gao, Liu, and Sun}]{chen2021textual}
Yangyi Chen, Fanchao Qi, Hongcheng Gao, Zhiyuan Liu, and Maosong Sun. 2021{\natexlab{b}}.
\newblock Textual backdoor attacks can be more harmful via two simple tricks.
\newblock \emph{arXiv preprint arXiv:2110.08247}.

\bibitem[{Chen et~al.(2012)Chen, Zhou, Zhu, and Xu}]{chen2012detecting}
Ying Chen, Yilu Zhou, Sencun Zhu, and Heng Xu. 2012.
\newblock Detecting offensive language in social media to protect adolescent online safety.
\newblock In \emph{2012 International Conference on Privacy, Security, Risk and Trust and 2012 International Confernece on Social Computing}, pages 71--80. IEEE.

\bibitem[{Chiang et~al.(2023)Chiang, Li, Lin, Sheng, Wu, Zhang, Zheng, Zhuang, Zhuang, Gonzalez et~al.}]{chiang2023vicuna}
Wei-Lin Chiang, Zhuohan Li, Zi~Lin, Ying Sheng, Zhanghao Wu, Hao Zhang, Lianmin Zheng, Siyuan Zhuang, Yonghao Zhuang, Joseph~E Gonzalez, et~al. 2023.
\newblock Vicuna: An open-source chatbot impressing {G}{P}{T}-4 with 90\%* {C}hat{G}{P}{T} quality.
\newblock \emph{See https://vicuna. lmsys. org (accessed 14 April 2023)}.

\bibitem[{Chowdhery et~al.(2022)Chowdhery, Narang, Devlin, Bosma, Mishra, Roberts, Barham, Chung, Sutton, Gehrmann, Schuh, Shi, Tsvyashchenko, Maynez, Rao, Barnes, Tay, Shazeer, Prabhakaran, Reif, Du, Hutchinson, Pope, Bradbury, Austin, Isard, Gur{-}Ari, Yin, Duke, Levskaya, Ghemawat, Dev, Michalewski, Garcia, Misra, Robinson, Fedus, Zhou, Ippolito, Luan, Lim, Zoph, Spiridonov, Sepassi, Dohan, Agrawal, Omernick, Dai, Pillai, Pellat, Lewkowycz, Moreira, Child, Polozov, Lee, Zhou, Wang, Saeta, Diaz, Firat, Catasta, Wei, Meier{-}Hellstern, Eck, Dean, Petrov, and Fiedel}]{DBLP:journals/corr/abs-2204-02311}
Aakanksha Chowdhery, Sharan Narang, Jacob Devlin, Maarten Bosma, Gaurav Mishra, Adam Roberts, Paul Barham, Hyung~Won Chung, Charles Sutton, Sebastian Gehrmann, Parker Schuh, Kensen Shi, Sasha Tsvyashchenko, Joshua Maynez, Abhishek Rao, Parker Barnes, Yi~Tay, Noam Shazeer, Vinodkumar Prabhakaran, Emily Reif, Nan Du, Ben Hutchinson, Reiner Pope, James Bradbury, Jacob Austin, Michael Isard, Guy Gur{-}Ari, Pengcheng Yin, Toju Duke, Anselm Levskaya, Sanjay Ghemawat, Sunipa Dev, Henryk Michalewski, Xavier Garcia, Vedant Misra, Kevin Robinson, Liam Fedus, Denny Zhou, Daphne Ippolito, David Luan, Hyeontaek Lim, Barret Zoph, Alexander Spiridonov, Ryan Sepassi, David Dohan, Shivani Agrawal, Mark Omernick, Andrew~M. Dai, Thanumalayan~Sankaranarayana Pillai, Marie Pellat, Aitor Lewkowycz, Erica Moreira, Rewon Child, Oleksandr Polozov, Katherine Lee, Zongwei Zhou, Xuezhi Wang, Brennan Saeta, Mark Diaz, Orhan Firat, Michele Catasta, Jason Wei, Kathy Meier{-}Hellstern, Douglas Eck, Jeff Dean, Slav Petrov, and Noah Fiedel.
  2022.
\newblock \href {https://doi.org/10.48550/arXiv.2204.02311} {Palm: Scaling language modeling with pathways}.
\newblock \emph{CoRR}, abs/2204.02311.

\bibitem[{Christiano et~al.(2018)Christiano, Shlegeris, and Amodei}]{christiano2018supervising}
Paul Christiano, Buck Shlegeris, and Dario Amodei. 2018.
\newblock Supervising strong learners by amplifying weak experts.
\newblock \emph{arXiv preprint arXiv:1810.08575}.

\bibitem[{Costa-juss{\`a} et~al.(2023)Costa-juss{\`a}, Andrews, Smith, Hansanti, Ropers, Kalbassi, Gao, Licht, and Wood}]{costa2023multilingual}
Marta~R Costa-juss{\`a}, Pierre Andrews, Eric Smith, Prangthip Hansanti, Christophe Ropers, Elahe Kalbassi, Cynthia Gao, Daniel Licht, and Carleigh Wood. 2023.
\newblock Multilingual holistic bias: Extending descriptors and patterns to unveil demographic biases in languages at scale.
\newblock \emph{arXiv preprint arXiv:2305.13198}.

\bibitem[{Critch and Krueger(2020)}]{critch2020ai}
Andrew Critch and David Krueger. 2020.
\newblock Ai research considerations for human existential safety (arches).
\newblock \emph{arXiv preprint arXiv:2006.04948}.

\bibitem[{Dai et~al.(2022)Dai, Dong, Hao, Sui, Chang, and Wei}]{2021Knowledge}
Damai Dai, Li~Dong, Yaru Hao, Zhifang Sui, Baobao Chang, and Furu Wei. 2022.
\newblock \href {https://doi.org/10.18653/v1/2022.acl-long.581} {Knowledge neurons in pretrained transformers}.
\newblock In \emph{Proceedings of the 60th Annual Meeting of the Association for Computational Linguistics (Volume 1: Long Papers)}, pages 8493--8502, Dublin, Ireland. Association for Computational Linguistics.

\bibitem[{Dai et~al.(2023)Dai, Pan, Ji, Sun, Wang, and Yang}]{safe-rlhf}
Juntao Dai, Xuehai Pan, Jiaming Ji, Ruiyang Sun, Yizhou Wang, and Yaodong Yang. 2023.
\newblock Pku-beaver: Constrained value-aligned llm via safe rlhf.
\newblock \url{https://github.com/PKU-Alignment/safe-rlhf}.

\bibitem[{Dale(2021)}]{dale2021gpt}
Robert Dale. 2021.
\newblock {G}{P}{T}-3: What’s it good for?
\newblock \emph{Natural Language Engineering}, 27(1):113--118.

\bibitem[{Davidson et~al.(2017)Davidson, Warmsley, Macy, and Weber}]{DBLP:conf/icwsm/DavidsonWMW17}
Thomas Davidson, Dana Warmsley, Michael~W. Macy, and Ingmar Weber. 2017.
\newblock \href {https://aaai.org/ocs/index.php/ICWSM/ICWSM17/paper/view/15665} {Automated hate speech detection and the problem of offensive language}.
\newblock In \emph{Proceedings of the Eleventh International Conference on Web and Social Media, {ICWSM} 2017, Montr{\'{e}}al, Qu{\'{e}}bec, Canada, May 15-18, 2017}, pages 512--515. {AAAI} Press.

\bibitem[{de~Gibert et~al.(2018)de~Gibert, P{\'{e}}rez, Pablos, and Cuadros}]{DBLP:conf/acl-alw/GibertPPC18}
Ona de~Gibert, Naiara P{\'{e}}rez, Aitor~Garc{\'{\i}}a Pablos, and Montse Cuadros. 2018.
\newblock \href {https://doi.org/10.18653/v1/w18-5102} {Hate speech dataset from a white supremacy forum}.
\newblock In \emph{Proceedings of the 2nd Workshop on Abusive Language Online, ALW@EMNLP 2018, Brussels, Belgium, October 31, 2018}, pages 11--20. Association for Computational Linguistics.

\bibitem[{Deng et~al.(2023)Deng, Liu, Li, Wang, Zhang, Li, Wang, Zhang, and Liu}]{deng2023jailbreaker}
Gelei Deng, Yi~Liu, Yuekang Li, Kailong Wang, Ying Zhang, Zefeng Li, Haoyu Wang, Tianwei Zhang, and Yang Liu. 2023.
\newblock Jailbreaker: Automated jailbreak across multiple large language model chatbots.
\newblock \emph{arXiv preprint arXiv:2307.08715}.

\bibitem[{Deng et~al.(2022)Deng, Zhou, Sun, Zheng, Mi, Meng, and Huang}]{DBLP:conf/emnlp/DengZ0ZMMH22}
Jiawen Deng, Jingyan Zhou, Hao Sun, Chujie Zheng, Fei Mi, Helen Meng, and Minlie Huang. 2022.
\newblock \href {https://doi.org/10.18653/v1/2022.emnlp-main.796} {{COLD:} {A} benchmark for chinese offensive language detection}.
\newblock In \emph{Proceedings of the 2022 Conference on Empirical Methods in Natural Language Processing, {EMNLP} 2022, Abu Dhabi, United Arab Emirates, December 7-11, 2022}, pages 11580--11599. Association for Computational Linguistics.

\bibitem[{Deng et~al.(2021)Deng, Wang, Li, Shang, Liu, Rajasekaran, and Ding}]{deng2021tag}
Jieren Deng, Yijue Wang, Ji~Li, Chao Shang, Hang Liu, Sanguthevar Rajasekaran, and Caiwen Ding. 2021.
\newblock Tag: Gradient attack on transformer-based language models.
\newblock \emph{arXiv preprint arXiv:2103.06819}.

\bibitem[{Dettmers et~al.(2023)Dettmers, Pagnoni, Holtzman, and Zettlemoyer}]{dettmers2023qlora}
Tim Dettmers, Artidoro Pagnoni, Ari Holtzman, and Luke Zettlemoyer. 2023.
\newblock Qlora: Efficient finetuning of quantized llms.
\newblock \emph{arXiv preprint arXiv:2305.14314}.

\bibitem[{Dev et~al.(2020)Dev, Li, Phillips, and Srikumar}]{DBLP:conf/aaai/DevLPS20}
Sunipa Dev, Tao Li, Jeff~M. Phillips, and Vivek Srikumar. 2020.
\newblock \href {https://doi.org/10.1609/aaai.v34i05.6267} {On measuring and mitigating biased inferences of word embeddings}.
\newblock In \emph{The Thirty-Fourth {AAAI} Conference on Artificial Intelligence, {AAAI} 2020, The Thirty-Second Innovative Applications of Artificial Intelligence Conference, {IAAI} 2020, The Tenth {AAAI} Symposium on Educational Advances in Artificial Intelligence, {EAAI} 2020, New York, NY, USA, February 7-12, 2020}, pages 7659--7666. {AAAI} Press.

\bibitem[{Devlin et~al.(2019)Devlin, Chang, Lee, and Toutanova}]{DBLP:conf/naacl/DevlinCLT19}
Jacob Devlin, Ming{-}Wei Chang, Kenton Lee, and Kristina Toutanova. 2019.
\newblock \href {https://doi.org/10.18653/v1/n19-1423} {{BERT:} pre-training of deep bidirectional transformers for language understanding}.
\newblock In \emph{Proceedings of the 2019 Conference of the North American Chapter of the Association for Computational Linguistics: Human Language Technologies, {NAACL-HLT} 2019, Minneapolis, MN, USA, June 2-7, 2019, Volume 1 (Long and Short Papers)}, pages 4171--4186. Association for Computational Linguistics.

\bibitem[{Dhamala et~al.(2021)Dhamala, Sun, Kumar, Krishna, Pruksachatkun, Chang, and Gupta}]{DBLP:conf/fat/DhamalaSKKPCG21}
Jwala Dhamala, Tony Sun, Varun Kumar, Satyapriya Krishna, Yada Pruksachatkun, Kai{-}Wei Chang, and Rahul Gupta. 2021.
\newblock \href {https://doi.org/10.1145/3442188.3445924} {{BOLD:} dataset and metrics for measuring biases in open-ended language generation}.
\newblock In \emph{FAccT '21: 2021 {ACM} Conference on Fairness, Accountability, and Transparency, Virtual Event / Toronto, Canada, March 3-10, 2021}, pages 862--872. {ACM}.

\bibitem[{D{\'{\i}}az et~al.(2019)D{\'{\i}}az, Johnson, Lazar, Piper, and Gergle}]{DBLP:conf/ijcai/DiazJLPG19}
Mark D{\'{\i}}az, Isaac Johnson, Amanda Lazar, Anne~Marie Piper, and Darren Gergle. 2019.
\newblock \href {https://doi.org/10.24963/ijcai.2019/852} {Addressing age-related bias in sentiment analysis}.
\newblock In \emph{Proceedings of the Twenty-Eighth International Joint Conference on Artificial Intelligence, {IJCAI} 2019, Macao, China, August 10-16, 2019}, pages 6146--6150. ijcai.org.

\bibitem[{Djuric et~al.(2015)Djuric, Zhou, Morris, Grbovic, Radosavljevic, and Bhamidipati}]{DBLP:conf/www/DjuricZMGRB15}
Nemanja Djuric, Jing Zhou, Robin Morris, Mihajlo Grbovic, Vladan Radosavljevic, and Narayan Bhamidipati. 2015.
\newblock \href {https://doi.org/10.1145/2740908.2742760} {Hate speech detection with comment embeddings}.
\newblock In \emph{Proceedings of the 24th International Conference on World Wide Web Companion, {WWW} 2015, Florence, Italy, May 18-22, 2015 - Companion Volume}, pages 29--30. {ACM}.

\bibitem[{Dong et~al.(2023)Dong, Xiong, Goyal, Pan, Diao, Zhang, Shum, and Zhang}]{dong2023raft}
Hanze Dong, Wei Xiong, Deepanshu Goyal, Rui Pan, Shizhe Diao, Jipeng Zhang, Kashun Shum, and Tong Zhang. 2023.
\newblock Raft: Reward ranked finetuning for generative foundation model alignment.
\newblock \emph{arXiv preprint arXiv:2304.06767}.

\bibitem[{Du et~al.(2023)Du, Li, Torralba, Tenenbaum, and Mordatch}]{du2023improving}
Yilun Du, Shuang Li, Antonio Torralba, Joshua~B Tenenbaum, and Igor Mordatch. 2023.
\newblock Improving factuality and reasoning in language models through multiagent debate.
\newblock \emph{arXiv preprint arXiv:2305.14325}.

\bibitem[{Dubois et~al.(2023)Dubois, Li, Taori, Zhang, Gulrajani, Ba, Guestrin, Liang, and Hashimoto}]{dubois2023alpacafarm}
Yann Dubois, Xuechen Li, Rohan Taori, Tianyi Zhang, Ishaan Gulrajani, Jimmy Ba, Carlos Guestrin, Percy Liang, and Tatsunori~B Hashimoto. 2023.
\newblock Alpacafarm: A simulation framework for methods that learn from human feedback.
\newblock \emph{arXiv preprint arXiv:2305.14387}.

\bibitem[{Elazar et~al.(2021)Elazar, Kassner, Ravfogel, Ravichander, Hovy, Sch{\"u}tze, and Goldberg}]{elazar2021measuring}
Yanai Elazar, Nora Kassner, Shauli Ravfogel, Abhilasha Ravichander, Eduard Hovy, Hinrich Sch{\"u}tze, and Yoav Goldberg. 2021.
\newblock Measuring and improving consistency in pretrained language models.
\newblock \emph{Transactions of the Association for Computational Linguistics}, 9:1012--1031.

\bibitem[{Elhage et~al.(2022{\natexlab{a}})Elhage, Hume, Olsson, Nanda, Henighan, Johnston, ElShowk, Joseph, DasSarma, Mann, Hernandez, Askell, Ndousse, Jones, Drain, Chen, Bai, Ganguli, Lovitt, Hatfield-Dodds, Kernion, Conerly, Kravec, Fort, Kadavath, Jacobson, Tran-Johnson, Kaplan, Clark, Brown, McCandlish, Amodei, and Olah}]{elhage2022solu}
Nelson Elhage, Tristan Hume, Catherine Olsson, Neel Nanda, Tom Henighan, Scott Johnston, Sheer ElShowk, Nicholas Joseph, Nova DasSarma, Ben Mann, Danny Hernandez, Amanda Askell, Kamal Ndousse, Andy Jones, Dawn Drain, Anna Chen, Yuntao Bai, Deep Ganguli, Liane Lovitt, Zac Hatfield-Dodds, Jackson Kernion, Tom Conerly, Shauna Kravec, Stanislav Fort, Saurav Kadavath, Josh Jacobson, Eli Tran-Johnson, Jared Kaplan, Jack Clark, Tom Brown, Sam McCandlish, Dario Amodei, and Christopher Olah. 2022{\natexlab{a}}.
\newblock Softmax linear units.
\newblock \emph{Transformer Circuits Thread}.
\newblock Https://transformer-circuits.pub/2022/solu/index.html.

\bibitem[{Elhage et~al.(2022{\natexlab{b}})Elhage, Hume, Olsson, Schiefer, Henighan, Kravec, Hatfield-Dodds, Lasenby, Drain, Chen, Grosse, McCandlish, Kaplan, Amodei, Wattenberg, and Olah}]{elhage2022superposition}
Nelson Elhage, Tristan Hume, Catherine Olsson, Nicholas Schiefer, Tom Henighan, Shauna Kravec, Zac Hatfield-Dodds, Robert Lasenby, Dawn Drain, Carol Chen, Roger Grosse, Sam McCandlish, Jared Kaplan, Dario Amodei, Martin Wattenberg, and Christopher Olah. 2022{\natexlab{b}}.
\newblock Toy models of superposition.
\newblock \emph{Transformer Circuits Thread}.

\bibitem[{Elhage et~al.(2021)Elhage, Nanda, Olsson, Henighan, Joseph, Mann, Askell, Bai, Chen, Conerly, DasSarma, Drain, Ganguli, Hatfield-Dodds, Hernandez, Jones, Kernion, Lovitt, Ndousse, Amodei, Brown, Clark, Kaplan, McCandlish, and Olah}]{elhage2021mathematical}
Nelson Elhage, Neel Nanda, Catherine Olsson, Tom Henighan, Nicholas Joseph, Ben Mann, Amanda Askell, Yuntao Bai, Anna Chen, Tom Conerly, Nova DasSarma, Dawn Drain, Deep Ganguli, Zac Hatfield-Dodds, Danny Hernandez, Andy Jones, Jackson Kernion, Liane Lovitt, Kamal Ndousse, Dario Amodei, Tom Brown, Jack Clark, Jared Kaplan, Sam McCandlish, and Chris Olah. 2021.
\newblock A mathematical framework for transformer circuits.
\newblock \emph{Transformer Circuits Thread}.

\bibitem[{Elmahdy et~al.(2022)Elmahdy, Inan, and Sim}]{elmahdy2022privacy}
Adel Elmahdy, Huseyin~A Inan, and Robert Sim. 2022.
\newblock Privacy leakage in text classification: A data extraction approach.
\newblock \emph{arXiv preprint arXiv:2206.04591}.

\bibitem[{Eloundou et~al.(2023)Eloundou, Manning, Mishkin, and Rock}]{eloundou2023gpts}
Tyna Eloundou, Sam Manning, Pamela Mishkin, and Daniel Rock. 2023.
\newblock {G}{P}{T}s are {G}{P}{T}s: An early look at the labor market impact potential of large language models.
\newblock \emph{arXiv preprint arXiv:2303.10130}.

\bibitem[{ElSherief et~al.(2021)ElSherief, Ziems, Muchlinski, Anupindi, Seybolt, Choudhury, and Yang}]{DBLP:conf/emnlp/ElSheriefZMASCY21}
Mai ElSherief, Caleb Ziems, David Muchlinski, Vaishnavi Anupindi, Jordyn Seybolt, Munmun~De Choudhury, and Diyi Yang. 2021.
\newblock \href {https://doi.org/10.18653/v1/2021.emnlp-main.29} {Latent hatred: {A} benchmark for understanding implicit hate speech}.
\newblock In \emph{Proceedings of the 2021 Conference on Empirical Methods in Natural Language Processing, {EMNLP} 2021, Virtual Event / Punta Cana, Dominican Republic, 7-11 November, 2021}, pages 345--363. Association for Computational Linguistics.

\bibitem[{Fabbri et~al.(2021)Fabbri, Wu, Liu, and Xiong}]{fabbri2021qafacteval}
Alexander~R Fabbri, Chien-Sheng Wu, Wenhao Liu, and Caiming Xiong. 2021.
\newblock Qafacteval: Improved qa-based factual consistency evaluation for summarization.
\newblock \emph{arXiv preprint arXiv:2112.08542}.

\bibitem[{FAIR et~al.(2022)FAIR, Bakhtin, Brown, Dinan, Farina, Flaherty, Fried, Goff, Gray, Hu et~al.}]{meta2022human}
FAIR, Anton Bakhtin, Noam Brown, Emily Dinan, Gabriele Farina, Colin Flaherty, Daniel Fried, Andrew Goff, Jonathan Gray, Hengyuan Hu, et~al. 2022.
\newblock Human-level play in the game of diplomacy by combining language models with strategic reasoning.
\newblock \emph{Science}, 378(6624):1067--1074.

\bibitem[{Fernandes et~al.(2023)Fernandes, Madaan, Liu, Farinhas, Martins, Bertsch, de~Souza, Zhou, Wu, Neubig et~al.}]{fernandes2023bridging}
Patrick Fernandes, Aman Madaan, Emmy Liu, Ant{\'o}nio Farinhas, Pedro~Henrique Martins, Amanda Bertsch, Jos{\'e}~GC de~Souza, Shuyan Zhou, Tongshuang Wu, Graham Neubig, et~al. 2023.
\newblock Bridging the gap: A survey on integrating (human) feedback for natural language generation.
\newblock \emph{arXiv preprint arXiv:2305.00955}.

\bibitem[{Fluri et~al.(2023)Fluri, Paleka, and Tram{\`e}r}]{fluri2023evaluating}
Lukas Fluri, Daniel Paleka, and Florian Tram{\`e}r. 2023.
\newblock Evaluating superhuman models with consistency checks.
\newblock \emph{arXiv preprint arXiv:2306.09983}.

\bibitem[{Forbes et~al.(2020)Forbes, Hwang, Shwartz, Sap, and Choi}]{forbes2020social}
Maxwell Forbes, Jena~D Hwang, Vered Shwartz, Maarten Sap, and Yejin Choi. 2020.
\newblock Social chemistry 101: Learning to reason about social and moral norms.
\newblock \emph{arXiv preprint arXiv:2011.00620}.

\bibitem[{Fu et~al.(2023)Fu, Ng, Jiang, and Liu}]{fu2023gptscore}
Jinlan Fu, See-Kiong Ng, Zhengbao Jiang, and Pengfei Liu. 2023.
\newblock {G}{P}{T}score: Evaluate as you desire.
\newblock \emph{arXiv preprint arXiv:2302.04166}.

\bibitem[{Gao et~al.(2023)Gao, Schulman, and Hilton}]{gao2023scaling}
Leo Gao, John Schulman, and Jacob Hilton. 2023.
\newblock Scaling laws for reward model overoptimization.
\newblock In \emph{International Conference on Machine Learning}, pages 10835--10866. PMLR.

\bibitem[{Gao et~al.(2020)Gao, Doan, Zhang, Ma, Zhang, Fu, Nepal, and Kim}]{gao2020backdoor}
Yansong Gao, Bao~Gia Doan, Zhi Zhang, Siqi Ma, Jiliang Zhang, Anmin Fu, Surya Nepal, and Hyoungshick Kim. 2020.
\newblock Backdoor attacks and countermeasures on deep learning: A comprehensive review.
\newblock \emph{arXiv preprint arXiv:2007.10760}.

\bibitem[{Gaut et~al.(2020)Gaut, Sun, Tang, Huang, Qian, ElSherief, Zhao, Mirza, Belding, Chang, and Wang}]{DBLP:conf/acl/GautSTHQEZMBCW20}
Andrew Gaut, Tony Sun, Shirlyn Tang, Yuxin Huang, Jing Qian, Mai ElSherief, Jieyu Zhao, Diba Mirza, Elizabeth~M. Belding, Kai{-}Wei Chang, and William~Yang Wang. 2020.
\newblock \href {https://doi.org/10.18653/v1/2020.acl-main.265} {Towards understanding gender bias in relation extraction}.
\newblock In \emph{Proceedings of the 58th Annual Meeting of the Association for Computational Linguistics, {ACL} 2020, Online, July 5-10, 2020}, pages 2943--2953. Association for Computational Linguistics.

\bibitem[{GDPR()}]{gdpr}
GDPR.
\newblock General data protection regulation.
\newblock \url{https://gdpr-info.eu/}.

\bibitem[{Gehman et~al.(2020)Gehman, Gururangan, Sap, Choi, and Smith}]{gehman2020realtoxicityprompts}
Samuel Gehman, Suchin Gururangan, Maarten Sap, Yejin Choi, and Noah~A Smith. 2020.
\newblock Realtoxicityprompts: Evaluating neural toxic degeneration in language models.
\newblock \emph{arXiv preprint arXiv:2009.11462}.

\bibitem[{Geva et~al.(2022)Geva, Caciularu, Wang, and Goldberg}]{geva2022transformer}
Mor Geva, Avi Caciularu, Kevin Wang, and Yoav Goldberg. 2022.
\newblock Transformer feed-forward layers build predictions by promoting concepts in the vocabulary space.
\newblock In \emph{Proceedings of the 2022 Conference on Empirical Methods in Natural Language Processing}, pages 30--45.

\bibitem[{Geva et~al.(2021)Geva, Schuster, Berant, and Levy}]{geva2021transformer}
Mor Geva, Roei Schuster, Jonathan Berant, and Omer Levy. 2021.
\newblock Transformer feed-forward layers are key-value memories.
\newblock In \emph{Proceedings of the 2021 Conference on Empirical Methods in Natural Language Processing}, pages 5484--5495.

\bibitem[{Glaese et~al.(2022)Glaese, McAleese, Tr{\k{e}}bacz, Aslanides, Firoiu, Ewalds, Rauh, Weidinger, Chadwick, Thacker et~al.}]{glaese2022improving}
Amelia Glaese, Nat McAleese, Maja Tr{\k{e}}bacz, John Aslanides, Vlad Firoiu, Timo Ewalds, Maribeth Rauh, Laura Weidinger, Martin Chadwick, Phoebe Thacker, et~al. 2022.
\newblock Improving alignment of dialogue agents via targeted human judgements.
\newblock \emph{arXiv preprint arXiv:2209.14375}.

\bibitem[{Go et~al.(2023)Go, Korbak, Kruszewski, Rozen, Ryu, and Dymetman}]{go2023aligning}
Dongyoung Go, Tomasz Korbak, Germ{\'a}n Kruszewski, Jos Rozen, Nahyeon Ryu, and Marc Dymetman. 2023.
\newblock Aligning language models with preferences through f-divergence minimization.
\newblock \emph{arXiv preprint arXiv:2302.08215}.

\bibitem[{Goyal et~al.(2023)Goyal, Doddapaneni, Khapra, and Ravindran}]{goyal2023survey}
Shreya Goyal, Sumanth Doddapaneni, Mitesh~M Khapra, and Balaraman Ravindran. 2023.
\newblock A survey of adversarial defenses and robustness in nlp.
\newblock \emph{ACM Computing Surveys}, 55(14s):1--39.

\bibitem[{Greshake et~al.(2023)Greshake, Abdelnabi, Mishra, Endres, Holz, and Fritz}]{greshake2023more}
Kai Greshake, Sahar Abdelnabi, Shailesh Mishra, Christoph Endres, Thorsten Holz, and Mario Fritz. 2023.
\newblock More than you've asked for: A comprehensive analysis of novel prompt injection threats to application-integrated large language models.
\newblock \emph{arXiv preprint arXiv:2302.12173}.

\bibitem[{Guo et~al.(2022)Guo, Xie, Li, Lyu, and Zhang}]{guo2022threats}
Shangwei Guo, Chunlong Xie, Jiwei Li, Lingjuan Lyu, and Tianwei Zhang. 2022.
\newblock Threats to pre-trained language models: Survey and taxonomy.
\newblock \emph{arXiv preprint arXiv:2202.06862}.

\bibitem[{Gupta et~al.(2022)Gupta, Huang, Zhong, Gao, Li, and Chen}]{gupta2022recovering}
Samyak Gupta, Yangsibo Huang, Zexuan Zhong, Tianyu Gao, Kai Li, and Danqi Chen. 2022.
\newblock Recovering private text in federated learning of language models.
\newblock \emph{Advances in Neural Information Processing Systems}, 35:8130--8143.

\bibitem[{Hadfield-Menell et~al.(2017)Hadfield-Menell, Milli, Abbeel, Russell, and Dragan}]{hadfield2017inverse}
Dylan Hadfield-Menell, Smitha Milli, Pieter Abbeel, Stuart~J Russell, and Anca Dragan. 2017.
\newblock Inverse reward design.
\newblock \emph{Advances in neural information processing systems}, 30.

\bibitem[{Hartvigsen et~al.(2022)Hartvigsen, Gabriel, Palangi, Sap, Ray, and Kamar}]{DBLP:conf/acl/HartvigsenGPSRK22}
Thomas Hartvigsen, Saadia Gabriel, Hamid Palangi, Maarten Sap, Dipankar Ray, and Ece Kamar. 2022.
\newblock \href {https://doi.org/10.18653/v1/2022.acl-long.234} {Toxigen: {A} large-scale machine-generated dataset for adversarial and implicit hate speech detection}.
\newblock In \emph{Proceedings of the 60th Annual Meeting of the Association for Computational Linguistics (Volume 1: Long Papers), {ACL} 2022, Dublin, Ireland, May 22-27, 2022}, pages 3309--3326. Association for Computational Linguistics.

\bibitem[{Hendrycks et~al.(2020)Hendrycks, Burns, Basart, Critch, Li, Song, and Steinhardt}]{hendrycks2020aligning}
Dan Hendrycks, Collin Burns, Steven Basart, Andrew Critch, Jerry Li, Dawn Song, and Jacob Steinhardt. 2020.
\newblock Aligning ai with shared human values.
\newblock \emph{arXiv preprint arXiv:2008.02275}.

\bibitem[{Hendrycks et~al.(2021)Hendrycks, Carlini, Schulman, and Steinhardt}]{hendrycks2021unsolved}
Dan Hendrycks, Nicholas Carlini, John Schulman, and Jacob Steinhardt. 2021.
\newblock Unsolved problems in ml safety.
\newblock \emph{arXiv preprint arXiv:2109.13916}.

\bibitem[{Hendrycks et~al.(2023)Hendrycks, Mazeika, and Woodside}]{hendrycks2023overview}
Dan Hendrycks, Mantas Mazeika, and Thomas Woodside. 2023.
\newblock An overview of catastrophic ai risks.
\newblock \emph{arXiv preprint arXiv:2306.12001}.

\bibitem[{Herd(2023)}]{lesswrongAgentizedLLMs}
Seth Herd. 2023.
\newblock {A}gentized {L}{L}{M}s will change the alignment landscape.
\newblock \url{https://www.lesswrong.com/posts/dcoxvEhAfYcov2LA6/agentized-llms-will-change-the-alignment-landscape}.

\bibitem[{Hisamoto et~al.(2020)Hisamoto, Post, and Duh}]{hisamoto2020membership}
Sorami Hisamoto, Matt Post, and Kevin Duh. 2020.
\newblock Membership inference attacks on sequence-to-sequence models: Is my data in your machine translation system?
\newblock \emph{Transactions of the Association for Computational Linguistics}, 8:49--63.

\bibitem[{Ho et~al.(2022)Ho, Schmid, and Yun}]{ho2022large}
Namgyu Ho, Laura Schmid, and Se-Young Yun. 2022.
\newblock Large language models are reasoning teachers.
\newblock \emph{arXiv preprint arXiv:2212.10071}.

\bibitem[{Hoffmann et~al.(2022)Hoffmann, Borgeaud, Mensch, Buchatskaya, Cai, Rutherford, Casas, Hendricks, Welbl, Clark et~al.}]{hoffmann2022training}
Jordan Hoffmann, Sebastian Borgeaud, Arthur Mensch, Elena Buchatskaya, Trevor Cai, Eliza Rutherford, Diego de~Las Casas, Lisa~Anne Hendricks, Johannes Welbl, Aidan Clark, et~al. 2022.
\newblock Training compute-optimal large language models.
\newblock \emph{arXiv preprint arXiv:2203.15556}.

\bibitem[{Honovich et~al.(2022)Honovich, Aharoni, Herzig, Taitelbaum, Kukliansy, Cohen, Scialom, Szpektor, Hassidim, and Matias}]{honovich2022true}
Or~Honovich, Roee Aharoni, Jonathan Herzig, Hagai Taitelbaum, Doron Kukliansy, Vered Cohen, Thomas Scialom, Idan Szpektor, Avinatan Hassidim, and Yossi Matias. 2022.
\newblock True: Re-evaluating factual consistency evaluation.
\newblock \emph{arXiv preprint arXiv:2204.04991}.

\bibitem[{Hosseini et~al.(2023)Hosseini, Palangi, and Awadallah}]{hosseini2023empirical}
Saghar Hosseini, Hamid Palangi, and Ahmed~Hassan Awadallah. 2023.
\newblock An empirical study of metrics to measure representational harms in pre-trained language models.
\newblock \emph{arXiv preprint arXiv:2301.09211}.

\bibitem[{Huang et~al.(2022)Huang, Abbeel, Pathak, and Mordatch}]{DBLP:conf/icml/HuangAPM22}
Wenlong Huang, Pieter Abbeel, Deepak Pathak, and Igor Mordatch. 2022.
\newblock \href {https://proceedings.mlr.press/v162/huang22a.html} {Language models as zero-shot planners: Extracting actionable knowledge for embodied agents}.
\newblock In \emph{International Conference on Machine Learning, {ICML} 2022, 17-23 July 2022, Baltimore, Maryland, {USA}}, volume 162 of \emph{Proceedings of Machine Learning Research}, pages 9118--9147. {PMLR}.

\bibitem[{Huang et~al.(2023)Huang, Zhang, Sun et~al.}]{huang2023trustgpt}
Yue Huang, Qihui Zhang, Lichao Sun, et~al. 2023.
\newblock Trustgpt: A benchmark for trustworthy and responsible large language models.
\newblock \emph{arXiv preprint arXiv:2306.11507}.

\bibitem[{Huang and Xiong(2023)}]{huang2023cbbq}
Yufei Huang and Deyi Xiong. 2023.
\newblock Cbbq: A chinese bias benchmark dataset curated with human-ai collaboration for large language models.
\newblock \emph{arXiv preprint arXiv:2306.16244}.

\bibitem[{Hubinger(2019{\natexlab{a}})}]{concrete_experiments_in_inner_alignment}
Evan Hubinger. 2019{\natexlab{a}}.
\newblock Concrete experiments in inner alignment.
\newblock \url{https://www.lesswrong.com/posts/uSdPa9nrSgmXCtdKN/concrete-experiments-in-inner-alignment}.

\bibitem[{Hubinger(2019{\natexlab{b}})}]{relaxed_adversarial_training}
Evan Hubinger. 2019{\natexlab{b}}.
\newblock Relaxed adversarial training for inner alignment.
\newblock \url{https://www.lesswrong.com/posts/9Dy5YRaoCxH9zuJqa/relaxed-adversarial-training-for-inner-alignment}.

\bibitem[{Hubinger(2020{\natexlab{a}})}]{lesswrongSafetyMarket}
Evan Hubinger. 2020{\natexlab{a}}.
\newblock {A}{I} safety via market making.
\newblock \url{https://www.lesswrong.com/posts/YWwzccGbcHMJMpT45/ai-safety-via-market-making}.

\bibitem[{Hubinger(2020{\natexlab{b}})}]{hubinger2020overview}
Evan Hubinger. 2020{\natexlab{b}}.
\newblock An overview of 11 proposals for building safe advanced ai.
\newblock \emph{arXiv preprint arXiv:2012.07532}.

\bibitem[{Hubinger(2022{\natexlab{a}})}]{lesswrongTransparencyInterpretability}
Evan Hubinger. 2022{\natexlab{a}}.
\newblock {A} transparency and interpretability tech tree.
\newblock \url{https://www.lesswrong.com/posts/nbq2bWLcYmSGup9aF/a-transparency-and-interpretability-tech-tree}.

\bibitem[{Hubinger(2022{\natexlab{b}})}]{monitoring_for_deceptive_alignment}
Evan Hubinger. 2022{\natexlab{b}}.
\newblock Monitoring for deceptive alignment.
\newblock \url{https://www.lesswrong.com/posts/Km9sHjHTsBdbgwKyi/monitoring-for-deceptive-alignment}.

\bibitem[{Hubinger et~al.(2019{\natexlab{a}})Hubinger, van Merwijk, Mikulik, Skalse, and Garrabrant}]{deceptive_alignment}
Evan Hubinger, Chris van Merwijk, Vlad Mikulik, Joar Skalse, and Scott Garrabrant. 2019{\natexlab{a}}.
\newblock Deceptive alignment.
\newblock \url{https://www.lesswrong.com/s/r9tYkB2a8Fp4DN8yB/p/zthDPAjh9w6Ytbeks}.

\bibitem[{Hubinger et~al.(2019{\natexlab{b}})Hubinger, van Merwijk, Mikulik, Skalse, and Garrabrant}]{the_inner_alignment_problem}
Evan Hubinger, Chris van Merwijk, Vlad Mikulik, Joar Skalse, and Scott Garrabrant. 2019{\natexlab{b}}.
\newblock The inner alignment problem.
\newblock \url{https://www.lesswrong.com/posts/pL56xPoniLvtMDQ4J/the-inner-alignment-problem}.

\bibitem[{Hubinger et~al.(2019{\natexlab{c}})Hubinger, van Merwijk, Mikulik, Skalse, and Garrabrant}]{DBLP:journals/corr/abs-1906-01820}
Evan Hubinger, Chris van Merwijk, Vladimir Mikulik, Joar Skalse, and Scott Garrabrant. 2019{\natexlab{c}}.
\newblock \href {http://arxiv.org/abs/1906.01820} {Risks from learned optimization in advanced machine learning systems}.
\newblock \emph{CoRR}, abs/1906.01820.

\bibitem[{Irving and Askell(2019)}]{irving2019ai}
Geoffrey Irving and Amanda Askell. 2019.
\newblock Ai safety needs social scientists.
\newblock \emph{Distill}, 4(2):e14.

\bibitem[{Irving et~al.(2018)Irving, Christiano, and Amodei}]{irving2018ai}
Geoffrey Irving, Paul Christiano, and Dario Amodei. 2018.
\newblock Ai safety via debate.
\newblock \emph{arXiv preprint arXiv:1805.00899}.

\bibitem[{Iu and Wong(2023)}]{iu2023chatgpt}
Kwan~Yuen Iu and Vanessa Man-Yi Wong. 2023.
\newblock Chat{G}{P}{T} by openai: The end of litigation lawyers?
\newblock \emph{Available at SSRN}.

\bibitem[{Jawahar et~al.(2020)Jawahar, Abdul-Mageed, and Lakshmanan}]{jawahar2020automatic}
Ganesh Jawahar, Muhammad Abdul-Mageed, and Laks V.~S. Lakshmanan. 2020.
\newblock \href {http://arxiv.org/abs/2011.01314} {Automatic detection of machine generated text: A critical survey}.

\bibitem[{Ji et~al.(2023)Ji, Lee, Frieske, Yu, Su, Xu, Ishii, Bang, Madotto, and Fung}]{ji2023survey}
Ziwei Ji, Nayeon Lee, Rita Frieske, Tiezheng Yu, Dan Su, Yan Xu, Etsuko Ishii, Ye~Jin Bang, Andrea Madotto, and Pascale Fung. 2023.
\newblock Survey of hallucination in natural language generation.
\newblock \emph{ACM Computing Surveys}, 55(12):1--38.

\bibitem[{Jones et~al.(2023)Jones, Dragan, Raghunathan, and Steinhardt}]{jones2023automatically}
Erik Jones, Anca Dragan, Aditi Raghunathan, and Jacob Steinhardt. 2023.
\newblock Automatically auditing large language models via discrete optimization.
\newblock \emph{arXiv preprint arXiv:2303.04381}.

\bibitem[{Kandpal et~al.(2023)Kandpal, Jagielski, Tram{\`e}r, and Carlini}]{kandpal2023backdoor}
Nikhil Kandpal, Matthew Jagielski, Florian Tram{\`e}r, and Nicholas Carlini. 2023.
\newblock Backdoor attacks for in-context learning with language models.
\newblock \emph{arXiv preprint arXiv:2307.14692}.

\bibitem[{Kanter et~al.()Kanter, Williams, Kuczynski, Manbeck, Debreaux, and Rosen}]{kanterpreliminary}
Jonathan~W Kanter, Monnica~T Williams, Adam~M Kuczynski, Katherine~E Manbeck, Marlena Debreaux, and Daniel~C Rosen.
\newblock A preliminary report on the relationship between microaggressions against black people and racism among white college students.

\bibitem[{Kaplan et~al.(2020)Kaplan, McCandlish, Henighan, Brown, Chess, Child, Gray, Radford, Wu, and Amodei}]{kaplan2020scaling}
Jared Kaplan, Sam McCandlish, Tom Henighan, Tom~B Brown, Benjamin Chess, Rewon Child, Scott Gray, Alec Radford, Jeffrey Wu, and Dario Amodei. 2020.
\newblock Scaling laws for neural language models.
\newblock \emph{arXiv preprint arXiv:2001.08361}.

\bibitem[{Kennedy et~al.(2022)Kennedy, Atari, Davani, Yeh, Omrani, Kim, Coombs, Havaldar, Portillo-Wightman, Gonzalez, Hoover, Azatian, Hussain, Lara, Cardenas, Omary, Park, Wang, Wijaya, Zhang, Meyerowitz, and Dehghani}]{Kennedy2022}
Brendan Kennedy, Mohammad Atari, Aida~Mostafazadeh Davani, Leigh Yeh, Ali Omrani, Yehsong Kim, Kris Coombs, Shreya Havaldar, Gwenyth Portillo-Wightman, Elaine Gonzalez, Joe Hoover, Aida Azatian, Alyzeh Hussain, Austin Lara, Gabriel Cardenas, Adam Omary, Christina Park, Xin Wang, Clarisa Wijaya, Yong Zhang, Beth Meyerowitz, and Morteza Dehghani. 2022.
\newblock \href {https://doi.org/10.1007/s10579-021-09569-x} {Introducing the gab hate corpus: defining and applying hate-based rhetoric to social media posts at scale}.
\newblock \emph{Language Resources and Evaluation}, 56(1):79--108.

\bibitem[{Kenton et~al.(2021)Kenton, Everitt, Weidinger, Gabriel, Mikulik, and Irving}]{kenton2021alignment}
Zachary Kenton, Tom Everitt, Laura Weidinger, Iason Gabriel, Vladimir Mikulik, and Geoffrey Irving. 2021.
\newblock Alignment of language agents.
\newblock \emph{arXiv preprint arXiv:2103.14659}.

\bibitem[{Keskar et~al.(2019)Keskar, McCann, Varshney, Xiong, and Socher}]{DBLP:journals/corr/abs-1909-05858}
Nitish~Shirish Keskar, Bryan McCann, Lav~R. Varshney, Caiming Xiong, and Richard Socher. 2019.
\newblock \href {http://arxiv.org/abs/1909.05858} {{CTRL:} {A} conditional transformer language model for controllable generation}.
\newblock \emph{CoRR}, abs/1909.05858.

\bibitem[{Kim et~al.(2023)Kim, Bae, Shin, Kang, Kwak, Yoo, and Seo}]{kim2023aligning}
Sungdong Kim, Sanghwan Bae, Jamin Shin, Soyoung Kang, Donghyun Kwak, Kang~Min Yoo, and Minjoon Seo. 2023.
\newblock Aligning large language models through synthetic feedback.
\newblock \emph{arXiv preprint arXiv:2305.13735}.

\bibitem[{Kirchner et~al.(2022)Kirchner, Smith, and Thibodeau}]{kirchner2022understanding}
J~Kirchner, L~Smith, and J~Thibodeau. 2022.
\newblock Understanding ai alignment research: A systematic analysis.
\newblock \emph{arXiv preprint arXiv:2206.02841}.

\bibitem[{Kiritchenko and Mohammad(2018)}]{DBLP:conf/starsem/KiritchenkoM18}
Svetlana Kiritchenko and Saif~M. Mohammad. 2018.
\newblock \href {https://doi.org/10.18653/v1/s18-2005} {Examining gender and race bias in two hundred sentiment analysis systems}.
\newblock In \emph{Proceedings of the Seventh Joint Conference on Lexical and Computational Semantics, *SEM@NAACL-HLT 2018, New Orleans, Louisiana, USA, June 5-6, 2018}, pages 43--53. Association for Computational Linguistics.

\bibitem[{Krakovna(2022)}]{victoria2022paradigms}
Victoria Krakovna. 2022.
\newblock \href {https://www.alignmentforum.org/posts/JC7aJZjt2WvxxffGz/paradigms-of-ai-alignment-components-and-enablers} {Paradigms of ai alignment: components and enablers}.

\bibitem[{Krakovna and Kramar(2023)}]{krakovna2023power}
Victoria Krakovna and Janos Kramar. 2023.
\newblock Power-seeking can be probable and predictive for trained agents.
\newblock \emph{arXiv preprint arXiv:2304.06528}.

\bibitem[{Kumar and Pranesh(2021)}]{kumar2021tweetblm}
Sumit Kumar and Raj~Ratn Pranesh. 2021.
\newblock Tweetblm: A hate speech dataset and analysis of black lives matter-related microblogs on twitter.
\newblock \emph{arXiv preprint arXiv:2108.12521}.

\bibitem[{Laban et~al.(2022)Laban, Schnabel, Bennett, and Hearst}]{laban2022summac}
Philippe Laban, Tobias Schnabel, Paul~N Bennett, and Marti~A Hearst. 2022.
\newblock Summac: Re-visiting nli-based models for inconsistency detection in summarization.
\newblock \emph{Transactions of the Association for Computational Linguistics}, 10:163--177.

\bibitem[{Lee et~al.(2022)Lee, Ping, Xu, Patwary, Fung, Shoeybi, and Catanzaro}]{lee2022factuality}
Nayeon Lee, Wei Ping, Peng Xu, Mostofa Patwary, Pascale~N Fung, Mohammad Shoeybi, and Bryan Catanzaro. 2022.
\newblock Factuality enhanced language models for open-ended text generation.
\newblock \emph{Advances in Neural Information Processing Systems}, 35:34586--34599.

\bibitem[{Lee~Sharkey(2022{\natexlab{a}})}]{lee2022superposition}
Beren~Millidge Lee~Sharkey, Dan~Braun. 2022{\natexlab{a}}.
\newblock \href {https://www.alignmentforum.org/posts/z6QQJbtpkEAX3Aojj/interim-research-report-taking-features-out-of-superposition} {Taking features out of superposition with sparse autoencoders}.

\bibitem[{Lee~Sharkey(2022{\natexlab{b}})}]{lee2022current}
Beren~Millidge Lee~Sharkey, Sid~Black. 2022{\natexlab{b}}.
\newblock \href {https://www.alignmentforum.org/posts/Jgs7LQwmvErxR9BCC/current-themes-in-mechanistic-interpretability-research} {Current themes in mechanistic interpretability research}.

\bibitem[{Lehman et~al.(2021)Lehman, Jain, Pichotta, Goldberg, and Wallace}]{lehman2021does}
Eric Lehman, Sarthak Jain, Karl Pichotta, Yoav Goldberg, and Byron~C Wallace. 2021.
\newblock Does bert pretrained on clinical notes reveal sensitive data?
\newblock \emph{arXiv preprint arXiv:2104.07762}.

\bibitem[{Leike et~al.(2018)Leike, Krueger, Everitt, Martic, Maini, and Legg}]{leike2018scalable}
Jan Leike, David Krueger, Tom Everitt, Miljan Martic, Vishal Maini, and Shane Legg. 2018.
\newblock Scalable agent alignment via reward modeling: a research direction.
\newblock \emph{arXiv preprint arXiv:1811.07871}.

\bibitem[{Leike et~al.(2017)Leike, Martic, Krakovna, Ortega, Everitt, Lefrancq, Orseau, and Legg}]{leike2017ai}
Jan Leike, Miljan Martic, Victoria Krakovna, Pedro~A Ortega, Tom Everitt, Andrew Lefrancq, Laurent Orseau, and Shane Legg. 2017.
\newblock Ai safety gridworlds.
\newblock \emph{arXiv preprint arXiv:1711.09883}.

\bibitem[{Levesque(2011)}]{DBLP:conf/aaaiss/Levesque11}
Hector~J. Levesque. 2011.
\newblock \href {http://www.aaai.org/ocs/index.php/SSS/SSS11/paper/view/2502} {The winograd schema challenge}.
\newblock In \emph{Logical Formalizations of Commonsense Reasoning, Papers from the 2011 {AAAI} Spring Symposium, Technical Report SS-11-06, Stanford, California, USA, March 21-23, 2011}. {AAAI}.

\bibitem[{Lewis et~al.(2017)Lewis, Yarats, Dauphin, Parikh, and Batra}]{lewis2017deal}
Mike Lewis, Denis Yarats, Yann~N. Dauphin, Devi Parikh, and Dhruv Batra. 2017.
\newblock \href {http://arxiv.org/abs/1706.05125} {Deal or no deal? end-to-end learning for negotiation dialogues}.

\bibitem[{Li et~al.(2023{\natexlab{a}})Li, Guo, Fan, Xu, and Song}]{li2023multi}
Haoran Li, Dadi Guo, Wei Fan, Mingshi Xu, and Yangqiu Song. 2023{\natexlab{a}}.
\newblock Multi-step jailbreaking privacy attacks on {C}hat{G}{P}{T}.
\newblock \emph{arXiv preprint arXiv:2304.05197}.

\bibitem[{Li et~al.(2023{\natexlab{b}})Li, Patel, Vi{\'e}gas, Pfister, and Wattenberg}]{li2023inference}
Kenneth Li, Oam Patel, Fernanda Vi{\'e}gas, Hanspeter Pfister, and Martin Wattenberg. 2023{\natexlab{b}}.
\newblock Inference-time intervention: Eliciting truthful answers from a language model.
\newblock \emph{arXiv preprint arXiv:2306.03341}.

\bibitem[{Li et~al.(2021{\natexlab{a}})Li, Song, Li, Zeng, Ma, and Qiu}]{li2021backdoor}
Linyang Li, Demin Song, Xiaonan Li, Jiehang Zeng, Ruotian Ma, and Xipeng Qiu. 2021{\natexlab{a}}.
\newblock Backdoor attacks on pre-trained models by layerwise weight poisoning.
\newblock \emph{arXiv preprint arXiv:2108.13888}.

\bibitem[{Li et~al.(2023{\natexlab{c}})Li, Patel, and Du}]{li2023prd}
Ruosen Li, Teerth Patel, and Xinya Du. 2023{\natexlab{c}}.
\newblock Prd: Peer rank and discussion improve large language model based evaluations.
\newblock \emph{arXiv preprint arXiv:2307.02762}.

\bibitem[{Li et~al.(2021{\natexlab{b}})Li, Liu, Dong, Zhao, Xue, Zhu, and Lu}]{li2021hidden}
Shaofeng Li, Hui Liu, Tian Dong, Benjamin Zi~Hao Zhao, Minhui Xue, Haojin Zhu, and Jialiang Lu. 2021{\natexlab{b}}.
\newblock Hidden backdoors in human-centric language models.
\newblock In \emph{Proceedings of the 2021 ACM SIGSAC Conference on Computer and Communications Security}, pages 3123--3140.

\bibitem[{Li et~al.(2020)Li, Khot, Khashabi, Sabharwal, and Srikumar}]{li2020unqovering}
Tao Li, Tushar Khot, Daniel Khashabi, Ashish Sabharwal, and Vivek Srikumar. 2020.
\newblock Unqovering stereotyping biases via underspecified questions.
\newblock \emph{arXiv preprint arXiv:2010.02428}.

\bibitem[{Li et~al.(2023{\natexlab{d}})Li, Zhang, Dubois, Taori, Gulrajani, Guestrin, Liang, and Hashimoto}]{alpaca_eval}
Xuechen Li, Tianyi Zhang, Yann Dubois, Rohan Taori, Ishaan Gulrajani, Carlos Guestrin, Percy Liang, and Tatsunori~B. Hashimoto. 2023{\natexlab{d}}.
\newblock Alpacaeval: An automatic evaluator of instruction-following models.
\newblock \url{https://github.com/tatsu-lab/alpaca_eval}.

\bibitem[{Li et~al.(2022)Li, Jiang, Li, and Xia}]{li2022backdoor}
Yiming Li, Yong Jiang, Zhifeng Li, and Shu-Tao Xia. 2022.
\newblock Backdoor learning: A survey.
\newblock \emph{IEEE Transactions on Neural Networks and Learning Systems}.

\bibitem[{Li et~al.(2023{\natexlab{e}})Li, Peng, He, Galley, Gao, and Yan}]{li2023guiding}
Zekun Li, Baolin Peng, Pengcheng He, Michel Galley, Jianfeng Gao, and Xifeng Yan. 2023{\natexlab{e}}.
\newblock Guiding large language models via directional stimulus prompting.
\newblock \emph{arXiv preprint arXiv:2302.11520}.

\bibitem[{Liang et~al.(2022)Liang, Bommasani, Lee, Tsipras, Soylu, Yasunaga, Zhang, Narayanan, Wu, Kumar et~al.}]{liang2022holistic}
Percy Liang, Rishi Bommasani, Tony Lee, Dimitris Tsipras, Dilara Soylu, Michihiro Yasunaga, Yian Zhang, Deepak Narayanan, Yuhuai Wu, Ananya Kumar, et~al. 2022.
\newblock Holistic evaluation of language models.
\newblock \emph{arXiv preprint arXiv:2211.09110}.

\bibitem[{Liang et~al.(2023)Liang, He, Jiao, Wang, Wang, Wang, Yang, Tu, and Shi}]{liang2023encouraging}
Tian Liang, Zhiwei He, Wenxiang Jiao, Xing Wang, Yan Wang, Rui Wang, Yujiu Yang, Zhaopeng Tu, and Shuming Shi. 2023.
\newblock Encouraging divergent thinking in large language models through multi-agent debate.
\newblock \emph{arXiv preprint arXiv:2305.19118}.

\bibitem[{Lightman et~al.(2023)Lightman, Kosaraju, Burda, Edwards, Baker, Lee, Leike, Schulman, Sutskever, and Cobbe}]{lightman2023let}
Hunter Lightman, Vineet Kosaraju, Yura Burda, Harri Edwards, Bowen Baker, Teddy Lee, Jan Leike, John Schulman, Ilya Sutskever, and Karl Cobbe. 2023.
\newblock Let's verify step by step.
\newblock \emph{arXiv preprint arXiv:2305.20050}.

\bibitem[{Ligozat et~al.(2021)Ligozat, Lef{\`e}vre, Bugeau, and Combaz}]{ligozat2021unraveling}
Anne-Laure Ligozat, Julien Lef{\`e}vre, Aur{\'e}lie Bugeau, and Jacques Combaz. 2021.
\newblock Unraveling the hidden environmental impacts of ai solutions for environment.
\newblock \emph{arXiv preprint arXiv:2110.11822}.

\bibitem[{Lin et~al.(2021)Lin, Hilton, and Evans}]{lin2021truthfulqa}
Stephanie Lin, Jacob Hilton, and Owain Evans. 2021.
\newblock Truthfulqa: Measuring how models mimic human falsehoods.
\newblock \emph{arXiv preprint arXiv:2109.07958}.

\bibitem[{Lipton(2017)}]{lipton2017mythos}
Zachary~C. Lipton. 2017.
\newblock \href {http://arxiv.org/abs/1606.03490} {The mythos of model interpretability}.

\bibitem[{Liu et~al.(2023{\natexlab{a}})Liu, Sferrazza, and Abbeel}]{liu2023chain}
Hao Liu, Carmelo Sferrazza, and Pieter Abbeel. 2023{\natexlab{a}}.
\newblock Chain of hindsight aligns language models with feedback.
\newblock \emph{arXiv preprint arXiv:2302.02676}.

\bibitem[{Liu et~al.(2022{\natexlab{a}})Liu, Jia, Zhang, Zhuang, Liu, and Vosoughi}]{liu2022second}
Ruibo Liu, Chenyan Jia, Ge~Zhang, Ziyu Zhuang, Tony Liu, and Soroush Vosoughi. 2022{\natexlab{a}}.
\newblock Second thoughts are best: Learning to re-align with human values from text edits.
\newblock \emph{Advances in Neural Information Processing Systems}, 35:181--196.

\bibitem[{Liu et~al.(2023{\natexlab{b}})Liu, Yang, Jia, Zhang, Zhou, Dai, Yang, and Vosoughi}]{liu2023training}
Ruibo Liu, Ruixin Yang, Chenyan Jia, Ge~Zhang, Denny Zhou, Andrew~M Dai, Diyi Yang, and Soroush Vosoughi. 2023{\natexlab{b}}.
\newblock Training socially aligned language models in simulated human society.
\newblock \emph{arXiv preprint arXiv:2305.16960}.

\bibitem[{Liu et~al.(2022{\natexlab{b}})Liu, Zhang, Feng, and Vosoughi}]{liu2022aligning}
Ruibo Liu, Ge~Zhang, Xinyu Feng, and Soroush Vosoughi. 2022{\natexlab{b}}.
\newblock Aligning generative language models with human values.
\newblock In \emph{Findings of the Association for Computational Linguistics: NAACL 2022}, pages 241--252.

\bibitem[{Liu et~al.(2023{\natexlab{c}})Liu, Iter, Xu, Wang, Xu, and Zhu}]{liu2023gpteval}
Yang Liu, Dan Iter, Yichong Xu, Shuohang Wang, Ruochen Xu, and Chenguang Zhu. 2023{\natexlab{c}}.
\newblock {G}{P}{T}eval: Nlg evaluation using {G}{P}{T}-4 with better human alignment.
\newblock \emph{arXiv preprint arXiv:2303.16634}.

\bibitem[{Liu et~al.(2023{\natexlab{d}})Liu, Yao, Ton, Zhang, Cheng, Klochkov, Taufiq, and Li}]{liu2023trustworthy}
Yang Liu, Yuanshun Yao, Jean-Francois Ton, Xiaoying Zhang, Ruocheng Guo~Hao Cheng, Yegor Klochkov, Muhammad~Faaiz Taufiq, and Hang Li. 2023{\natexlab{d}}.
\newblock Trustworthy llms: a survey and guideline for evaluating large language models' alignment.
\newblock \emph{arXiv preprint arXiv:2308.05374}.

\bibitem[{Liu et~al.(2023{\natexlab{e}})Liu, Deng, Li, Wang, Zhang, Liu, Wang, Zheng, and Liu}]{liu2023prompt}
Yi~Liu, Gelei Deng, Yuekang Li, Kailong Wang, Tianwei Zhang, Yepang Liu, Haoyu Wang, Yan Zheng, and Yang Liu. 2023{\natexlab{e}}.
\newblock Prompt injection attack against llm-integrated applications.
\newblock \emph{arXiv preprint arXiv:2306.05499}.

\bibitem[{Liu et~al.(2019)Liu, Ott, Goyal, Du, Joshi, Chen, Levy, Lewis, Zettlemoyer, and Stoyanov}]{DBLP:journals/corr/abs-1907-11692}
Yinhan Liu, Myle Ott, Naman Goyal, Jingfei Du, Mandar Joshi, Danqi Chen, Omer Levy, Mike Lewis, Luke Zettlemoyer, and Veselin Stoyanov. 2019.
\newblock \href {http://arxiv.org/abs/1907.11692} {Roberta: {A} robustly optimized {BERT} pretraining approach}.
\newblock \emph{CoRR}, abs/1907.11692.

\bibitem[{Lourie et~al.(2021)Lourie, Le~Bras, and Choi}]{lourie2021scruples}
Nicholas Lourie, Ronan Le~Bras, and Yejin Choi. 2021.
\newblock Scruples: A corpus of community ethical judgments on 32,000 real-life anecdotes.
\newblock In \emph{Proceedings of the AAAI Conference on Artificial Intelligence}, volume~35, pages 13470--13479.

\bibitem[{Lucy and Bamman(2021)}]{lucy2021gender}
Li~Lucy and David Bamman. 2021.
\newblock Gender and representation bias in {G}{P}{T}-3 generated stories.
\newblock In \emph{Proceedings of the Third Workshop on Narrative Understanding}, pages 48--55.

\bibitem[{MacAvaney et~al.(2019)MacAvaney, Yao, Yang, Russell, Goharian, and Frieder}]{macavaney2019hate}
Sean MacAvaney, Hao-Ren Yao, Eugene Yang, Katina Russell, Nazli Goharian, and Ophir Frieder. 2019.
\newblock Hate speech detection: Challenges and solutions.
\newblock \emph{PloS one}, 14(8):e0221152.

\bibitem[{Magister et~al.(2022)Magister, Mallinson, Adamek, Malmi, and Severyn}]{magister2022teaching}
Lucie~Charlotte Magister, Jonathan Mallinson, Jakub Adamek, Eric Malmi, and Aliaksei Severyn. 2022.
\newblock Teaching small language models to reason.
\newblock \emph{arXiv preprint arXiv:2212.08410}.

\bibitem[{Martinez(2020)}]{martinez2020more}
Veronica~Root Martinez. 2020.
\newblock More meaningful ethics.
\newblock \emph{U. Chi. L. Rev. Online}, page~53.

\bibitem[{McAllister et~al.(2017)McAllister, Gal, Kendall, Van Der~Wilk, Shah, Cipolla, and Weller}]{mcallister2017concrete}
RT~McAllister, Yarin Gal, Alex Kendall, Mark Van Der~Wilk, Amar Shah, Roberto Cipolla, and Adrian Weller. 2017.
\newblock Concrete problems for autonomous vehicle safety: Advantages of bayesian deep learning.
\newblock International Joint Conferences on Artificial Intelligence, Inc.

\bibitem[{Meng et~al.(2022)Meng, Bau, Andonian, and Belinkov}]{meng2022locating}
Kevin Meng, David Bau, Alex Andonian, and Yonatan Belinkov. 2022.
\newblock Locating and editing factual associations in {G}{P}{T}.
\newblock \emph{Advances in Neural Information Processing Systems}, 35:17359--17372.

\bibitem[{Mialon et~al.(2023)Mialon, Dess{\`{\i}}, Lomeli, Nalmpantis, Pasunuru, Raileanu, Rozi{\`{e}}re, Schick, Dwivedi{-}Yu, Celikyilmaz, Grave, LeCun, and Scialom}]{DBLP:journals/corr/abs-2302-07842}
Gr{\'{e}}goire Mialon, Roberto Dess{\`{\i}}, Maria Lomeli, Christoforos Nalmpantis, Ramakanth Pasunuru, Roberta Raileanu, Baptiste Rozi{\`{e}}re, Timo Schick, Jane Dwivedi{-}Yu, Asli Celikyilmaz, Edouard Grave, Yann LeCun, and Thomas Scialom. 2023.
\newblock \href {https://doi.org/10.48550/arXiv.2302.07842} {Augmented language models: a survey}.
\newblock \emph{CoRR}, abs/2302.07842.

\bibitem[{Mikulik(2019)}]{2d_robustness}
Vladimir Mikulik. 2019.
\newblock 2-d robustness.
\newblock \url{https://www.alignmentforum.org/posts/2mhFMgtAjFJesaSYR/2-d-robustness}.

\bibitem[{Min et~al.(2023)Min, Krishna, Lyu, Lewis, Yih, Koh, Iyyer, Zettlemoyer, and Hajishirzi}]{min2023factscore}
Sewon Min, Kalpesh Krishna, Xinxi Lyu, Mike Lewis, Wen-tau Yih, Pang~Wei Koh, Mohit Iyyer, Luke Zettlemoyer, and Hannaneh Hajishirzi. 2023.
\newblock Factscore: Fine-grained atomic evaluation of factual precision in long form text generation.
\newblock \emph{arXiv preprint arXiv:2305.14251}.

\bibitem[{Mireshghallah et~al.(2022)Mireshghallah, Goyal, Uniyal, Berg-Kirkpatrick, and Shokri}]{mireshghallah2022quantifying}
Fatemehsadat Mireshghallah, Kartik Goyal, Archit Uniyal, Taylor Berg-Kirkpatrick, and Reza Shokri. 2022.
\newblock Quantifying privacy risks of masked language models using membership inference attacks.
\newblock \emph{arXiv preprint arXiv:2203.03929}.

\bibitem[{Mireshghallah et~al.(2020)Mireshghallah, Taram, Vepakomma, Singh, Raskar, and Esmaeilzadeh}]{mireshghallah2020privacy}
Fatemehsadat Mireshghallah, Mohammadkazem Taram, Praneeth Vepakomma, Abhishek Singh, Ramesh Raskar, and Hadi Esmaeilzadeh. 2020.
\newblock Privacy in deep learning: A survey.
\newblock \emph{arXiv preprint arXiv:2004.12254}.

\bibitem[{Nadal(2018)}]{nadal2018microaggressions}
Kevin~L Nadal. 2018.
\newblock \emph{Microaggressions and traumatic stress: Theory, research, and clinical treatment.}
\newblock American Psychological Association.

\bibitem[{Nadal et~al.(2014)Nadal, Griffin, Wong, Hamit, and Rasmus}]{nadal2014impact}
Kevin~L Nadal, Katie~E Griffin, Yinglee Wong, Sahran Hamit, and Morgan Rasmus. 2014.
\newblock The impact of racial microaggressions on mental health: Counseling implications for clients of color.
\newblock \emph{Journal of Counseling \& Development}, 92:57.

\bibitem[{Nadeem et~al.(2020)Nadeem, Bethke, and Reddy}]{nadeem2020stereoset}
Moin Nadeem, Anna Bethke, and Siva Reddy. 2020.
\newblock Stereoset: Measuring stereotypical bias in pretrained language models.
\newblock \emph{arXiv preprint arXiv:2004.09456}.

\bibitem[{Nanda(2022)}]{neel2022comprehensive}
Neel Nanda. 2022.
\newblock \href {https://www.neelnanda.io/mechanistic-interpretability/glossary} {A comprehensive mechanistic interpretability explainer \& glossary}.

\bibitem[{Nanda et~al.(2023)Nanda, Chan, Liberum, Smith, and Steinhardt}]{nanda2023progress}
Neel Nanda, Lawrence Chan, Tom Liberum, Jess Smith, and Jacob Steinhardt. 2023.
\newblock Progress measures for grokking via mechanistic interpretability.
\newblock \emph{arXiv preprint arXiv:2301.05217}.

\bibitem[{Nangia et~al.(2020)Nangia, Vania, Bhalerao, and Bowman}]{DBLP:conf/emnlp/NangiaVBB20}
Nikita Nangia, Clara Vania, Rasika Bhalerao, and Samuel~R. Bowman. 2020.
\newblock \href {https://doi.org/10.18653/v1/2020.emnlp-main.154} {Crows-pairs: {A} challenge dataset for measuring social biases in masked language models}.
\newblock In \emph{Proceedings of the 2020 Conference on Empirical Methods in Natural Language Processing, {EMNLP} 2020, Online, November 16-20, 2020}, pages 1953--1967. Association for Computational Linguistics.

\bibitem[{Ng and Russell(2000)}]{ng2000algorithms}
Andrew~Y Ng and Stuart~J Russell. 2000.
\newblock Algorithms for inverse reinforcement learning.
\newblock In \emph{Proceedings of the Seventeenth International Conference on Machine Learning}, pages 663--670.

\bibitem[{Ngo(2022)}]{ngo2022alignment}
Richard Ngo. 2022.
\newblock The alignment problem from a deep learning perspective.
\newblock \emph{arXiv preprint arXiv:2209.00626}.

\bibitem[{Nozza et~al.(2021)Nozza, Bianchi, Hovy et~al.}]{nozza2021honest}
Debora Nozza, Federico Bianchi, Dirk Hovy, et~al. 2021.
\newblock Honest: Measuring hurtful sentence completion in language models.
\newblock In \emph{Proceedings of the 2021 Conference of the North American Chapter of the Association for Computational Linguistics: Human Language Technologies}. Association for Computational Linguistics.

\bibitem[{Olah(2022)}]{Chris2022mechanistic}
Chris Olah. 2022.
\newblock Mechanistic interpretability, variables, and the importance of interpretable bases.

\bibitem[{Olsson et~al.(2022)Olsson, Elhage, Nanda, Joseph, DasSarma, Henighan, Mann, Askell, Bai, Chen, Conerly, Drain, Ganguli, Hatfield-Dodds, Hernandez, Johnston, Jones, Kernion, Lovitt, Ndousse, Amodei, Brown, Clark, Kaplan, McCandlish, and Olah}]{olsson2022context}
Catherine Olsson, Nelson Elhage, Neel Nanda, Nicholas Joseph, Nova DasSarma, Tom Henighan, Ben Mann, Amanda Askell, Yuntao Bai, Anna Chen, Tom Conerly, Dawn Drain, Deep Ganguli, Zac Hatfield-Dodds, Danny Hernandez, Scott Johnston, Andy Jones, Jackson Kernion, Liane Lovitt, Kamal Ndousse, Dario Amodei, Tom Brown, Jack Clark, Jared Kaplan, Sam McCandlish, and Chris Olah. 2022.
\newblock In-context learning and induction heads.
\newblock \emph{Transformer Circuits Thread}.

\bibitem[{Omohundro(2008)}]{omohundro2008basic}
Stephen~M Omohundro. 2008.
\newblock The basic ai drives.
\newblock In \emph{AGI}, volume 171, pages 483--492.

\bibitem[{OpenAI(2022)}]{openaiIntroducingChatGPT}
OpenAI. 2022.
\newblock {I}ntroducing {C}hat{G}{P}{T}.
\newblock \url{https://openai.com/blog/chatgpt/}.

\bibitem[{OpenAI(2023{\natexlab{a}})}]{openai2023gpt}
OpenAI. 2023{\natexlab{a}}.
\newblock {G}{P}{T}-4 technical report.
\newblock \emph{arXiv preprint arXiv:2303.08774}.

\bibitem[{OpenAI(2023{\natexlab{b}})}]{openaiIntroducingSuperalignment}
OpenAI. 2023{\natexlab{b}}.
\newblock {I}ntroducing {S}uperalignment.
\newblock \url{https://openai.com/blog/introducing-superalignment}.

\bibitem[{Ouyang et~al.(2022)Ouyang, Wu, Jiang, Almeida, Wainwright, Mishkin, Zhang, Agarwal, Slama, Ray et~al.}]{ouyang2022training}
Long Ouyang, Jeffrey Wu, Xu~Jiang, Diogo Almeida, Carroll Wainwright, Pamela Mishkin, Chong Zhang, Sandhini Agarwal, Katarina Slama, Alex Ray, et~al. 2022.
\newblock Training language models to follow instructions with human feedback.
\newblock \emph{Advances in Neural Information Processing Systems}, 35:27730--27744.

\bibitem[{Pan et~al.(2023)Pan, Saxon, Xu, Nathani, Wang, and Wang}]{pan2023automatically}
Liangming Pan, Michael Saxon, Wenda Xu, Deepak Nathani, Xinyi Wang, and William~Yang Wang. 2023.
\newblock Automatically correcting large language models: Surveying the landscape of diverse self-correction strategies.
\newblock \emph{arXiv preprint arXiv:2308.03188}.

\bibitem[{Parrish et~al.(2022)Parrish, Chen, Nangia, Padmakumar, Phang, Thompson, Htut, and Bowman}]{DBLP:conf/acl/ParrishCNPPTHB22}
Alicia Parrish, Angelica Chen, Nikita Nangia, Vishakh Padmakumar, Jason Phang, Jana Thompson, Phu~Mon Htut, and Samuel~R. Bowman. 2022.
\newblock \href {https://doi.org/10.18653/v1/2022.findings-acl.165} {{BBQ:} {A} hand-built bias benchmark for question answering}.
\newblock In \emph{Findings of the Association for Computational Linguistics: {ACL} 2022, Dublin, Ireland, May 22-27, 2022}, pages 2086--2105. Association for Computational Linguistics.

\bibitem[{Perez et~al.(2022)Perez, Ringer, Luko{\v{s}}i{\=u}t{\.e}, Nguyen, Chen, Heiner, Pettit, Olsson, Kundu, Kadavath et~al.}]{perez2022discovering}
Ethan Perez, Sam Ringer, Kamil{\.e} Luko{\v{s}}i{\=u}t{\.e}, Karina Nguyen, Edwin Chen, Scott Heiner, Craig Pettit, Catherine Olsson, Sandipan Kundu, Saurav Kadavath, et~al. 2022.
\newblock Discovering language model behaviors with model-written evaluations.
\newblock \emph{arXiv preprint arXiv:2212.09251}.

\bibitem[{Poesia et~al.(2022)Poesia, Polozov, Le, Tiwari, Soares, Meek, and Gulwani}]{DBLP:conf/iclr/PoesiaP00SMG22}
Gabriel Poesia, Alex Polozov, Vu~Le, Ashish Tiwari, Gustavo Soares, Christopher Meek, and Sumit Gulwani. 2022.
\newblock \href {https://openreview.net/forum?id=KmtVD97J43e} {Synchromesh: Reliable code generation from pre-trained language models}.
\newblock In \emph{The Tenth International Conference on Learning Representations, {ICLR} 2022, Virtual Event, April 25-29, 2022}. OpenReview.net.

\bibitem[{Qi et~al.(2021)Qi, Li, Chen, Zhang, Liu, Wang, and Sun}]{qi2021hidden}
Fanchao Qi, Mukai Li, Yangyi Chen, Zhengyan Zhang, Zhiyuan Liu, Yasheng Wang, and Maosong Sun. 2021.
\newblock Hidden killer: Invisible textual backdoor attacks with syntactic trigger.
\newblock \emph{arXiv preprint arXiv:2105.12400}.

\bibitem[{Qi et~al.(2023)Qi, Huang, Panda, Wang, and Mittal}]{qi2023visual}
Xiangyu Qi, Kaixuan Huang, Ashwinee Panda, Mengdi Wang, and Prateek Mittal. 2023.
\newblock Visual adversarial examples jailbreak large language models.
\newblock \emph{arXiv preprint arXiv:2306.13213}.

\bibitem[{Qin et~al.(2023)Qin, Hu, Lin, Chen, Ding, Cui, Zeng, Huang, Xiao, Han, Fung, Su, Wang, Qian, Tian, Zhu, Liang, Shen, Xu, Zhang, Ye, Li, Tang, Yi, Zhu, Dai, Yan, Cong, Lu, Zhao, Huang, Yan, Han, Sun, Li, Phang, Yang, Wu, Ji, Liu, and Sun}]{DBLP:journals/corr/abs-2304-08354}
Yujia Qin, Shengding Hu, Yankai Lin, Weize Chen, Ning Ding, Ganqu Cui, Zheni Zeng, Yufei Huang, Chaojun Xiao, Chi Han, Yi~Ren Fung, Yusheng Su, Huadong Wang, Cheng Qian, Runchu Tian, Kunlun Zhu, Shihao Liang, Xingyu Shen, Bokai Xu, Zhen Zhang, Yining Ye, Bowen Li, Ziwei Tang, Jing Yi, Yuzhang Zhu, Zhenning Dai, Lan Yan, Xin Cong, Yaxi Lu, Weilin Zhao, Yuxiang Huang, Junxi Yan, Xu~Han, Xian Sun, Dahai Li, Jason Phang, Cheng Yang, Tongshuang Wu, Heng Ji, Zhiyuan Liu, and Maosong Sun. 2023.
\newblock \href {https://doi.org/10.48550/arXiv.2304.08354} {Tool learning with foundation models}.
\newblock \emph{CoRR}, abs/2304.08354.

\bibitem[{Qiu et~al.(2022)Qiu, Liu, Zhou, and Huang}]{qiu2022adversarial}
Shilin Qiu, Qihe Liu, Shijie Zhou, and Wen Huang. 2022.
\newblock Adversarial attack and defense technologies in natural language processing: A survey.
\newblock \emph{Neurocomputing}, 492:278--307.

\bibitem[{Radford et~al.(2018)Radford, Narasimhan, Salimans, and Sutskever}]{radford2018improving}
Alec Radford, Karthik Narasimhan, Tim Salimans, and Ilya Sutskever. 2018.
\newblock Improving language understanding by generative pre-training.

\bibitem[{Radford et~al.(2019)Radford, Wu, Child, Luan, Amodei, Sutskever et~al.}]{radford2019language}
Alec Radford, Jeffrey Wu, Rewon Child, David Luan, Dario Amodei, Ilya Sutskever, et~al. 2019.
\newblock Language models are unsupervised multitask learners.
\newblock \emph{OpenAI blog}, 1(8):9.

\bibitem[{Radhakrishnan(2022)}]{lesswrongRLHFLessWrong}
Ansh Radhakrishnan. 2022.
\newblock {R}{L}{H}{F}.
\newblock \url{https://www.lesswrong.com/posts/rQH4gRmPMJyjtMpTn/rlhf}.

\bibitem[{Rae et~al.(2021)Rae, Borgeaud, Cai, Millican, Hoffmann, Song, Aslanides, Henderson, Ring, Young et~al.}]{rae2021scaling}
Jack~W Rae, Sebastian Borgeaud, Trevor Cai, Katie Millican, Jordan Hoffmann, Francis Song, John Aslanides, Sarah Henderson, Roman Ring, Susannah Young, et~al. 2021.
\newblock Scaling language models: Methods, analysis \& insights from training gopher.
\newblock \emph{arXiv preprint arXiv:2112.11446}.

\bibitem[{Rafailov et~al.(2023)Rafailov, Sharma, Mitchell, Ermon, Manning, and Finn}]{rafailov2023direct}
Rafael Rafailov, Archit Sharma, Eric Mitchell, Stefano Ermon, Christopher~D Manning, and Chelsea Finn. 2023.
\newblock Direct preference optimization: Your language model is secretly a reward model.
\newblock \emph{arXiv preprint arXiv:2305.18290}.

\bibitem[{Raffel et~al.(2020)Raffel, Shazeer, Roberts, Lee, Narang, Matena, Zhou, Li, and Liu}]{DBLP:journals/jmlr/RaffelSRLNMZLL20}
Colin Raffel, Noam Shazeer, Adam Roberts, Katherine Lee, Sharan Narang, Michael Matena, Yanqi Zhou, Wei Li, and Peter~J. Liu. 2020.
\newblock \href {http://jmlr.org/papers/v21/20-074.html} {Exploring the limits of transfer learning with a unified text-to-text transformer}.
\newblock \emph{J. Mach. Learn. Res.}, 21:140:1--140:67.

\bibitem[{Razavi et~al.(2010)Razavi, Inkpen, Uritsky, and Matwin}]{razavi2010offensive}
Amir~H Razavi, Diana Inkpen, Sasha Uritsky, and Stan Matwin. 2010.
\newblock Offensive language detection using multi-level classification.
\newblock In \emph{Advances in Artificial Intelligence: 23rd Canadian Conference on Artificial Intelligence, Canadian AI 2010, Ottawa, Canada, May 31--June 2, 2010. Proceedings 23}, pages 16--27. Springer.

\bibitem[{Renduchintala and Williams(2021)}]{DBLP:journals/corr/abs-2104-07838}
Adithya Renduchintala and Adina Williams. 2021.
\newblock \href {http://arxiv.org/abs/2104.07838} {Investigating failures of automatic translation in the case of unambiguous gender}.
\newblock \emph{CoRR}, abs/2104.07838.

\bibitem[{Rigaki and Garcia(2020)}]{rigaki2020survey}
Maria Rigaki and Sebastian Garcia. 2020.
\newblock A survey of privacy attacks in machine learning.
\newblock \emph{arXiv preprint arXiv:2007.07646}.

\bibitem[{Ross et~al.(2017)Ross, Rist, Carbonell, Cabrera, Kurowsky, and Wojatzki}]{ross2017measuring}
Bj{\"o}rn Ross, Michael Rist, Guillermo Carbonell, Benjamin Cabrera, Nils Kurowsky, and Michael Wojatzki. 2017.
\newblock Measuring the reliability of hate speech annotations: The case of the european refugee crisis.
\newblock \emph{arXiv preprint arXiv:1701.08118}.

\bibitem[{Rudinger et~al.(2018)Rudinger, Naradowsky, Leonard, and Van~Durme}]{rudinger2018gender}
Rachel Rudinger, Jason Naradowsky, Brian Leonard, and Benjamin Van~Durme. 2018.
\newblock Gender bias in coreference resolution.
\newblock \emph{arXiv preprint arXiv:1804.09301}.

\bibitem[{Russell(1998)}]{russell1998learning}
Stuart~J Russell. 1998.
\newblock Learning agents for uncertain environments.
\newblock In \emph{Proceedings of the eleventh annual conference on Computational learning theory}, pages 101--103.

\bibitem[{Russell and Norvig(2010)}]{russell2010artificial}
Stuart~J Russell and Peter Norvig. 2010.
\newblock \emph{Artificial intelligence a modern approach}.
\newblock Pearson Education, Inc.

\bibitem[{Saleem and Anderson()}]{saleemarabs}
Muniba Saleem and Craig~A Anderson.
\newblock Arabs as terrorists: Effects of stereotypes within violent contexts on attitudes, perceptions, and affect.

\bibitem[{Sandbrink(2023)}]{sandbrink2023artificial}
Jonas~B Sandbrink. 2023.
\newblock Artificial intelligence and biological misuse: Differentiating risks of language models and biological design tools.
\newblock \emph{arXiv preprint arXiv:2306.13952}.

\bibitem[{Sap et~al.(2019)Sap, Gabriel, Qin, Jurafsky, Smith, and Choi}]{sap2019social}
Maarten Sap, Saadia Gabriel, Lianhui Qin, Dan Jurafsky, Noah~A Smith, and Yejin Choi. 2019.
\newblock Social bias frames: Reasoning about social and power implications of language.
\newblock \emph{arXiv preprint arXiv:1911.03891}.

\bibitem[{Scao et~al.(2022)Scao, Fan, Akiki, Pavlick, Ilic, Hesslow, Castagn{\'{e}}, Luccioni, Yvon, Gall{\'{e}}, Tow, Rush, Biderman, Webson, Ammanamanchi, Wang, Sagot, Muennighoff, del Moral, Ruwase, Bawden, Bekman, McMillan{-}Major, Beltagy, Nguyen, Saulnier, Tan, Suarez, Sanh, Lauren{\c{c}}on, Jernite, Launay, Mitchell, Raffel, Gokaslan, Simhi, Soroa, Aji, Alfassy, Rogers, Nitzav, Xu, Mou, Emezue, Klamm, Leong, van Strien, Adelani, and et~al.}]{DBLP:journals/corr/abs-2211-05100}
Teven~Le Scao, Angela Fan, Christopher Akiki, Ellie Pavlick, Suzana Ilic, Daniel Hesslow, Roman Castagn{\'{e}}, Alexandra~Sasha Luccioni, Fran{\c{c}}ois Yvon, Matthias Gall{\'{e}}, Jonathan Tow, Alexander~M. Rush, Stella Biderman, Albert Webson, Pawan~Sasanka Ammanamanchi, Thomas Wang, Beno{\^{\i}}t Sagot, Niklas Muennighoff, Albert~Villanova del Moral, Olatunji Ruwase, Rachel Bawden, Stas Bekman, Angelina McMillan{-}Major, Iz~Beltagy, Huu Nguyen, Lucile Saulnier, Samson Tan, Pedro~Ortiz Suarez, Victor Sanh, Hugo Lauren{\c{c}}on, Yacine Jernite, Julien Launay, Margaret Mitchell, Colin Raffel, Aaron Gokaslan, Adi Simhi, Aitor Soroa, Alham~Fikri Aji, Amit Alfassy, Anna Rogers, Ariel~Kreisberg Nitzav, Canwen Xu, Chenghao Mou, Chris Emezue, Christopher Klamm, Colin Leong, Daniel van Strien, David~Ifeoluwa Adelani, and et~al. 2022.
\newblock \href {https://doi.org/10.48550/arXiv.2211.05100} {{BLOOM:} {A} 176b-parameter open-access multilingual language model}.
\newblock \emph{CoRR}, abs/2211.05100.

\bibitem[{Scheurer et~al.(2023)Scheurer, Campos, Korbak, Chan, Chen, Cho, and Perez}]{scheurer2023training}
J{\'e}r{\'e}my Scheurer, Jon~Ander Campos, Tomasz Korbak, Jun~Shern Chan, Angelica Chen, Kyunghyun Cho, and Ethan Perez. 2023.
\newblock Training language models with language feedback at scale.
\newblock \emph{arXiv preprint arXiv:2303.16755}.

\bibitem[{Schick et~al.(2021)Schick, Udupa, and Sch{\"u}tze}]{schick2021self}
Timo Schick, Sahana Udupa, and Hinrich Sch{\"u}tze. 2021.
\newblock Self-diagnosis and self-debiasing: A proposal for reducing corpus-based bias in nlp.
\newblock \emph{Transactions of the Association for Computational Linguistics}, 9:1408--1424.

\bibitem[{Schmidt and Wiegand(2017)}]{DBLP:conf/acl-socialnlp/SchmidtW17}
Anna Schmidt and Michael Wiegand. 2017.
\newblock \href {https://doi.org/10.18653/v1/w17-1101} {A survey on hate speech detection using natural language processing}.
\newblock In \emph{Proceedings of the Fifth International Workshop on Natural Language Processing for Social Media, SocialNLP@EACL 2017, Valencia, Spain, April 3, 2017}, pages 1--10. Association for Computational Linguistics.

\bibitem[{Schulman et~al.(2017)Schulman, Wolski, Dhariwal, Radford, and Klimov}]{schulman2017proximal}
John Schulman, Filip Wolski, Prafulla Dhariwal, Alec Radford, and Oleg Klimov. 2017.
\newblock Proximal policy optimization algorithms.
\newblock \emph{arXiv preprint arXiv:1707.06347}.

\bibitem[{Schwartz et~al.(2012)Schwartz, Cieciuch, Vecchione, Davidov, Fischer, Beierlein, Ramos, Verkasalo, L{\"o}nnqvist, Demirutku et~al.}]{schwartz2012refining}
Shalom~H Schwartz, Jan Cieciuch, Michele Vecchione, Eldad Davidov, Ronald Fischer, Constanze Beierlein, Alice Ramos, Markku Verkasalo, Jan-Erik L{\"o}nnqvist, Kursad Demirutku, et~al. 2012.
\newblock Refining the theory of basic individual values.
\newblock \emph{Journal of personality and social psychology}, 103(4):663.

\bibitem[{Segerie(2023)}]{lesswrongTaskDecomposition}
Charbel-Raphaël Segerie. 2023.
\newblock {T}ask decomposition for scalable oversight ({A}{G}{I}{S}{F} {D}istillation).
\newblock \url{https://www.lesswrong.com/posts/FFz6H35Gy6BArHxkc/task-decomposition-for-scalable-oversight-agisf-distillation}.

\bibitem[{Shah et~al.(2019)Shah, Schwartz, and Hovy}]{shah2019predictive}
Deven Shah, H~Andrew Schwartz, and Dirk Hovy. 2019.
\newblock Predictive biases in natural language processing models: A conceptual framework and overview.
\newblock \emph{arXiv preprint arXiv:1912.11078}.

\bibitem[{Shah(2023)}]{categorizing_alignment}
Rohin Shah. 2023.
\newblock Categorizing failures as “outer” or “inner” misalignment is often confused.
\newblock \url{https://www.lesswrong.com/posts/JKwrDwsaRiSxTv9ur/categorizing-failures-as-outer-or-inner-misalignment-is}.

\bibitem[{Shah et~al.(2022)Shah, Varma, Kumar, Phuong, Krakovna, Uesato, and Kenton}]{shah2022goal}
Rohin Shah, Vikrant Varma, Ramana Kumar, Mary Phuong, Victoria Krakovna, Jonathan Uesato, and Zac Kenton. 2022.
\newblock Goal misgeneralization: Why correct specifications aren't enough for correct goals.
\newblock \emph{arXiv preprint arXiv:2210.01790}.

\bibitem[{Sheng et~al.(2022)Sheng, Han, Li, and Chang}]{sheng2022survey}
Xuan Sheng, Zhaoyang Han, Piji Li, and Xiangmao Chang. 2022.
\newblock A survey on backdoor attack and defense in natural language processing.
\newblock In \emph{2022 IEEE 22nd International Conference on Software Quality, Reliability and Security (QRS)}, pages 809--820. IEEE.

\bibitem[{Shevlane et~al.(2023)Shevlane, Farquhar, Garfinkel, Phuong, Whittlestone, Leung, Kokotajlo, Marchal, Anderljung, Kolt et~al.}]{shevlane2023model}
Toby Shevlane, Sebastian Farquhar, Ben Garfinkel, Mary Phuong, Jess Whittlestone, Jade Leung, Daniel Kokotajlo, Nahema Marchal, Markus Anderljung, Noam Kolt, et~al. 2023.
\newblock Model evaluation for extreme risks.
\newblock \emph{arXiv preprint arXiv:2305.15324}.

\bibitem[{Shi et~al.(2023)Shi, Liu, Zhou, and Sun}]{shi2023badgpt}
Jiawen Shi, Yixin Liu, Pan Zhou, and Lichao Sun. 2023.
\newblock Bad{G}{P}{T}: Exploring security vulnerabilities of {C}hat{G}{P}{T} via backdoor attacks to {I}nstruct{G}{P}{T}.
\newblock \emph{arXiv preprint arXiv:2304.12298}.

\bibitem[{Smith et~al.(2022)Smith, Hall, Kambadur, Presani, and Williams}]{smith2022m}
Eric~Michael Smith, Melissa Hall, Melanie Kambadur, Eleonora Presani, and Adina Williams. 2022.
\newblock “i’m sorry to hear that”: Finding new biases in language models with a holistic descriptor dataset.
\newblock In \emph{Proceedings of the 2022 Conference on Empirical Methods in Natural Language Processing}, pages 9180--9211.

\bibitem[{Soares(2015{\natexlab{a}})}]{soares2015aligning}
Nate Soares. 2015{\natexlab{a}}.
\newblock Aligning superintelligence with human interests: An annotated bibliography.
\newblock \emph{Intelligence}, 17(4):391--444.

\bibitem[{Soares(2015{\natexlab{b}})}]{intelligenceResearchGuide}
Nate Soares. 2015{\natexlab{b}}.
\newblock {R}esearch {G}uide - {M}achine {I}ntelligence {R}esearch {I}nstitute.
\newblock \url{https://intelligence.org/research-guide}.

\bibitem[{Song et~al.(2022)Song, Wu, Washington, Sadler, Chao, and Su}]{DBLP:journals/corr/abs-2212-04088}
Chan~Hee Song, Jiaman Wu, Clayton Washington, Brian~M. Sadler, Wei{-}Lun Chao, and Yu~Su. 2022.
\newblock \href {https://doi.org/10.48550/arXiv.2212.04088} {Llm-planner: Few-shot grounded planning for embodied agents with large language models}.
\newblock \emph{CoRR}, abs/2212.04088.

\bibitem[{Song and Raghunathan(2020)}]{song2020information}
Congzheng Song and Ananth Raghunathan. 2020.
\newblock Information leakage in embedding models.
\newblock In \emph{Proceedings of the 2020 ACM SIGSAC conference on computer and communications security}, pages 377--390.

\bibitem[{Song and Shmatikov(2019)}]{song2019auditing}
Congzheng Song and Vitaly Shmatikov. 2019.
\newblock Auditing data provenance in text-generation models.
\newblock In \emph{Proceedings of the 25th ACM SIGKDD International Conference on Knowledge Discovery \& Data Mining}, pages 196--206.

\bibitem[{Song et~al.(2023)Song, Yu, Li, Yu, Huang, Li, and Wang}]{song2023preference}
Feifan Song, Bowen Yu, Minghao Li, Haiyang Yu, Fei Huang, Yongbin Li, and Houfeng Wang. 2023.
\newblock Preference ranking optimization for human alignment.
\newblock \emph{arXiv preprint arXiv:2306.17492}.

\bibitem[{Soral et~al.(2018)Soral, Bilewicz, and Winiewski}]{soral2018exposure}
Wiktor Soral, Micha{\l} Bilewicz, and Miko{\l}aj Winiewski. 2018.
\newblock Exposure to hate speech increases prejudice through desensitization.
\newblock \emph{Aggressive behavior}, 44(2):136--146.

\bibitem[{Sousa and Kern(2023)}]{sousa2023keep}
Samuel Sousa and Roman Kern. 2023.
\newblock How to keep text private? a systematic review of deep learning methods for privacy-preserving natural language processing.
\newblock \emph{Artificial Intelligence Review}, 56(2):1427--1492.

\bibitem[{Srivastava et~al.(2022)Srivastava, Rastogi, Rao, Shoeb, Abid, Fisch, Brown, Santoro, Gupta, Garriga-Alonso et~al.}]{srivastava2022beyond}
Aarohi Srivastava, Abhinav Rastogi, Abhishek Rao, Abu Awal~Md Shoeb, Abubakar Abid, Adam Fisch, Adam~R Brown, Adam Santoro, Aditya Gupta, Adri{\`a} Garriga-Alonso, et~al. 2022.
\newblock Beyond the imitation game: Quantifying and extrapolating the capabilities of language models.
\newblock \emph{arXiv preprint arXiv:2206.04615}.

\bibitem[{Stanovsky et~al.(2019)Stanovsky, Smith, and Zettlemoyer}]{DBLP:conf/acl/StanovskySZ19}
Gabriel Stanovsky, Noah~A. Smith, and Luke Zettlemoyer. 2019.
\newblock \href {https://doi.org/10.18653/v1/p19-1164} {Evaluating gender bias in machine translation}.
\newblock In \emph{Proceedings of the 57th Conference of the Association for Computational Linguistics, {ACL} 2019, Florence, Italy, July 28- August 2, 2019, Volume 1: Long Papers}, pages 1679--1684. Association for Computational Linguistics.

\bibitem[{Stiennon et~al.(2020)Stiennon, Ouyang, Wu, Ziegler, Lowe, Voss, Radford, Amodei, and Christiano}]{stiennon2020learning}
Nisan Stiennon, Long Ouyang, Jeffrey Wu, Daniel Ziegler, Ryan Lowe, Chelsea Voss, Alec Radford, Dario Amodei, and Paul~F Christiano. 2020.
\newblock Learning to summarize with human feedback.
\newblock \emph{Advances in Neural Information Processing Systems}, 33:3008--3021.

\bibitem[{Stray(2020)}]{stray2020aligning}
Jonathan Stray. 2020.
\newblock Aligning ai optimization to community well-being.
\newblock \emph{International Journal of Community Well-Being}, 3(4):443--463.

\bibitem[{Stray et~al.(2021)Stray, Vendrov, Nixon, Adler, and Hadfield-Menell}]{stray2021you}
Jonathan Stray, Ivan Vendrov, Jeremy Nixon, Steven Adler, and Dylan Hadfield-Menell. 2021.
\newblock What are you optimizing for? aligning recommender systems with human values.
\newblock \emph{arXiv preprint arXiv:2107.10939}.

\bibitem[{Sue et~al.(2007)Sue, Capodilupo, Torino, Bucceri, Holder, Nadal, and Esquilin}]{sue2007racial}
Derald~Wing Sue, Christina~M Capodilupo, Gina~C Torino, Jennifer~M Bucceri, Aisha Holder, Kevin~L Nadal, and Marta Esquilin. 2007.
\newblock Racial microaggressions in everyday life: Implications for clinical practice.
\newblock \emph{American Psychologist}, 62(4):271--286.

\bibitem[{Sun et~al.(2023{\natexlab{a}})Sun, Zhang, Deng, Cheng, and Huang}]{sun2023safety}
Hao Sun, Zhexin Zhang, Jiawen Deng, Jiale Cheng, and Minlie Huang. 2023{\natexlab{a}}.
\newblock Safety assessment of chinese large language models.
\newblock \emph{arXiv preprint arXiv:2304.10436}.

\bibitem[{Sun et~al.(2023{\natexlab{b}})Sun, Shen, Zhou, Zhang, Chen, Cox, Yang, and Gan}]{sun2023principle}
Zhiqing Sun, Yikang Shen, Qinhong Zhou, Hongxin Zhang, Zhenfang Chen, David Cox, Yiming Yang, and Chuang Gan. 2023{\natexlab{b}}.
\newblock Principle-driven self-alignment of language models from scratch with minimal human supervision.
\newblock \emph{arXiv preprint arXiv:2305.03047}.

\bibitem[{Tamkin et~al.(2021)Tamkin, Brundage, Clark, and Ganguli}]{tamkin2021understanding}
Alex Tamkin, Miles Brundage, Jack Clark, and Deep Ganguli. 2021.
\newblock \href {http://arxiv.org/abs/2102.02503} {Understanding the capabilities, limitations, and societal impact of large language models}.

\bibitem[{Tay et~al.(2020)Tay, Ong, Fu, Chan, Chen, Luu, and Pal}]{tay2020would}
Yi~Tay, Donovan Ong, Jie Fu, Alvin Chan, Nancy Chen, Anh~Tuan Luu, and Christopher Pal. 2020.
\newblock Would you rather? a new benchmark for learning machine alignment with cultural values and social preferences.
\newblock In \emph{Proceedings of the 58th Annual Meeting of the Association for Computational Linguistics}, pages 5369--5373.

\bibitem[{Touvron et~al.(2023{\natexlab{a}})Touvron, Lavril, Izacard, Martinet, Lachaux, Lacroix, Rozi{\`{e}}re, Goyal, Hambro, Azhar, Rodriguez, Joulin, Grave, and Lample}]{DBLP:journals/corr/abs-2302-13971}
Hugo Touvron, Thibaut Lavril, Gautier Izacard, Xavier Martinet, Marie{-}Anne Lachaux, Timoth{\'{e}}e Lacroix, Baptiste Rozi{\`{e}}re, Naman Goyal, Eric Hambro, Faisal Azhar, Aur{\'{e}}lien Rodriguez, Armand Joulin, Edouard Grave, and Guillaume Lample. 2023{\natexlab{a}}.
\newblock \href {https://doi.org/10.48550/arXiv.2302.13971} {Llama: Open and efficient foundation language models}.
\newblock \emph{CoRR}, abs/2302.13971.

\bibitem[{Touvron et~al.(2023{\natexlab{b}})Touvron, Martin, Stone, Albert, Almahairi, Babaei, Bashlykov, Batra, Bhargava, Bhosale, Bikel, Blecher, Canton{-}Ferrer, Chen, Cucurull, Esiobu, Fernandes, Fu, Fu, Fuller, Gao, Goswami, Goyal, Hartshorn, Hosseini, Hou, Inan, Kardas, Kerkez, Khabsa, Kloumann, Korenev, Koura, Lachaux, Lavril, Lee, Liskovich, Lu, Mao, Martinet, Mihaylov, Mishra, Molybog, Nie, Poulton, Reizenstein, Rungta, Saladi, Schelten, Silva, Smith, Subramanian, Tan, Tang, Taylor, Williams, Kuan, Xu, Yan, Zarov, Zhang, Fan, Kambadur, Narang, Rodriguez, Stojnic, Edunov, and Scialom}]{DBLP:journals/corr/abs-2307-09288}
Hugo Touvron, Louis Martin, Kevin Stone, Peter Albert, Amjad Almahairi, Yasmine Babaei, Nikolay Bashlykov, Soumya Batra, Prajjwal Bhargava, Shruti Bhosale, Dan Bikel, Lukas Blecher, Cristian Canton{-}Ferrer, Moya Chen, Guillem Cucurull, David Esiobu, Jude Fernandes, Jeremy Fu, Wenyin Fu, Brian Fuller, Cynthia Gao, Vedanuj Goswami, Naman Goyal, Anthony Hartshorn, Saghar Hosseini, Rui Hou, Hakan Inan, Marcin Kardas, Viktor Kerkez, Madian Khabsa, Isabel Kloumann, Artem Korenev, Punit~Singh Koura, Marie{-}Anne Lachaux, Thibaut Lavril, Jenya Lee, Diana Liskovich, Yinghai Lu, Yuning Mao, Xavier Martinet, Todor Mihaylov, Pushkar Mishra, Igor Molybog, Yixin Nie, Andrew Poulton, Jeremy Reizenstein, Rashi Rungta, Kalyan Saladi, Alan Schelten, Ruan Silva, Eric~Michael Smith, Ranjan Subramanian, Xiaoqing~Ellen Tan, Binh Tang, Ross Taylor, Adina Williams, Jian~Xiang Kuan, Puxin Xu, Zheng Yan, Iliyan Zarov, Yuchen Zhang, Angela Fan, Melanie Kambadur, Sharan Narang, Aur{\'{e}}lien Rodriguez, Robert Stojnic, Sergey Edunov,
  and Thomas Scialom. 2023{\natexlab{b}}.
\newblock \href {https://doi.org/10.48550/arXiv.2307.09288} {Llama 2: Open foundation and fine-tuned chat models}.
\newblock \emph{CoRR}, abs/2307.09288.

\bibitem[{Turner et~al.(2021)Turner, Smith, Shah, Critch, and Tadepalli}]{NEURIPS2021_c26820b8}
Alex Turner, Logan Smith, Rohin Shah, Andrew Critch, and Prasad Tadepalli. 2021.
\newblock \href {https://proceedings.neurips.cc/paper_files/paper/2021/file/c26820b8a4c1b3c2aa868d6d57e14a79-Paper.pdf} {Optimal policies tend to seek power}.
\newblock In \emph{Advances in Neural Information Processing Systems}, volume~34, pages 23063--23074. Curran Associates, Inc.

\bibitem[{Turner and Tadepalli(2022)}]{NEURIPS2022_cb3658b9}
Alex Turner and Prasad Tadepalli. 2022.
\newblock \href {https://proceedings.neurips.cc/paper_files/paper/2022/file/cb3658b9983f677670a246c46ece553d-Paper-Conference.pdf} {Parametrically retargetable decision-makers tend to seek power}.
\newblock In \emph{Advances in Neural Information Processing Systems}, volume~35, pages 31391--31401. Curran Associates, Inc.

\bibitem[{Van~Wynsberghe(2021)}]{van2021sustainable}
Aimee Van~Wynsberghe. 2021.
\newblock Sustainable ai: Ai for sustainability and the sustainability of ai.
\newblock \emph{AI and Ethics}, 1(3):213--218.

\bibitem[{Vaswani et~al.(2017)Vaswani, Shazeer, Parmar, Uszkoreit, Jones, Gomez, Kaiser, and Polosukhin}]{vaswani2017attention}
Ashish Vaswani, Noam Shazeer, Niki Parmar, Jakob Uszkoreit, Llion Jones, Aidan~N Gomez, {\L}ukasz Kaiser, and Illia Polosukhin. 2017.
\newblock Attention is all you need.
\newblock In \emph{Proceedings of the 31st International Conference on Neural Information Processing Systems}, pages 6000--6010.

\bibitem[{Vidgen et~al.(2021)Vidgen, Thrush, Waseem, and Kiela}]{DBLP:conf/acl/VidgenTWK20}
Bertie Vidgen, Tristan Thrush, Zeerak Waseem, and Douwe Kiela. 2021.
\newblock \href {https://doi.org/10.18653/v1/2021.acl-long.132} {Learning from the worst: Dynamically generated datasets to improve online hate detection}.
\newblock In \emph{Proceedings of the 59th Annual Meeting of the Association for Computational Linguistics and the 11th International Joint Conference on Natural Language Processing, {ACL/IJCNLP} 2021, (Volume 1: Long Papers), Virtual Event, August 1-6, 2021}, pages 1667--1682. Association for Computational Linguistics.

\bibitem[{Vilone and Longo(2020)}]{vilone2020explainable}
Giulia Vilone and Luca Longo. 2020.
\newblock \href {http://arxiv.org/abs/2006.00093} {Explainable artificial intelligence: a systematic review}.

\bibitem[{Wallace et~al.(2019)Wallace, Feng, Kandpal, Gardner, and Singh}]{wallace2019universal}
Eric Wallace, Shi Feng, Nikhil Kandpal, Matt Gardner, and Sameer Singh. 2019.
\newblock Universal adversarial triggers for attacking and analyzing nlp.
\newblock \emph{arXiv preprint arXiv:1908.07125}.

\bibitem[{Wallace et~al.(2020)Wallace, Zhao, Feng, and Singh}]{wallace2020concealed}
Eric Wallace, Tony~Z Zhao, Shi Feng, and Sameer Singh. 2020.
\newblock Concealed data poisoning attacks on nlp models.
\newblock \emph{arXiv preprint arXiv:2010.12563}.

\bibitem[{Wang et~al.(2022{\natexlab{a}})Wang, Variengien, Conmy, Shlegeris, and Steinhardt}]{wang2022interpretability}
Kevin~Ro Wang, Alexandre Variengien, Arthur Conmy, Buck Shlegeris, and Jacob Steinhardt. 2022{\natexlab{a}}.
\newblock Interpretability in the wild: a circuit for indirect object identification in {G}{P}{T}-2 small.
\newblock In \emph{The Eleventh International Conference on Learning Representations}.

\bibitem[{Wang et~al.(2023{\natexlab{a}})Wang, Li, Chen, Zhu, Lin, Cao, Liu, Liu, and Sui}]{wang2023large}
Peiyi Wang, Lei Li, Liang Chen, Dawei Zhu, Binghuai Lin, Yunbo Cao, Qi~Liu, Tianyu Liu, and Zhifang Sui. 2023{\natexlab{a}}.
\newblock Large language models are not fair evaluators.
\newblock \emph{arXiv preprint arXiv:2305.17926}.

\bibitem[{Wang et~al.(2023{\natexlab{b}})Wang, Yu, Zeng, Yang, Wang, Chen, Jiang, Xie, Wang, Xie et~al.}]{wang2023pandalm}
Yidong Wang, Zhuohao Yu, Zhengran Zeng, Linyi Yang, Cunxiang Wang, Hao Chen, Chaoya Jiang, Rui Xie, Jindong Wang, Xing Xie, et~al. 2023{\natexlab{b}}.
\newblock Pandalm: An automatic evaluation benchmark for llm instruction tuning optimization.
\newblock \emph{arXiv preprint arXiv:2306.05087}.

\bibitem[{Wang et~al.(2023{\natexlab{c}})Wang, Ivison, Dasigi, Hessel, Khot, Chandu, Wadden, MacMillan, Smith, Beltagy et~al.}]{wang2023far}
Yizhong Wang, Hamish Ivison, Pradeep Dasigi, Jack Hessel, Tushar Khot, Khyathi~Raghavi Chandu, David Wadden, Kelsey MacMillan, Noah~A Smith, Iz~Beltagy, et~al. 2023{\natexlab{c}}.
\newblock How far can camels go? exploring the state of instruction tuning on open resources.
\newblock \emph{arXiv preprint arXiv:2306.04751}.

\bibitem[{Wang et~al.(2022{\natexlab{b}})Wang, Kordi, Mishra, Liu, Smith, Khashabi, and Hajishirzi}]{wang2022self}
Yizhong Wang, Yeganeh Kordi, Swaroop Mishra, Alisa Liu, Noah~A Smith, Daniel Khashabi, and Hannaneh Hajishirzi. 2022{\natexlab{b}}.
\newblock Self-instruct: Aligning language model with self generated instructions.
\newblock \emph{arXiv preprint arXiv:2212.10560}.

\bibitem[{Wang et~al.(2023{\natexlab{d}})Wang, Zhong, Li, Mi, Zeng, Huang, Shang, Jiang, and Liu}]{wang2023aligning}
Yufei Wang, Wanjun Zhong, Liangyou Li, Fei Mi, Xingshan Zeng, Wenyong Huang, Lifeng Shang, Xin Jiang, and Qun Liu. 2023{\natexlab{d}}.
\newblock Aligning large language models with human: A survey.
\newblock \emph{arXiv preprint arXiv:2307.12966}.

\bibitem[{Warner and Hirschberg(2012)}]{warner2012detecting}
William Warner and Julia Hirschberg. 2012.
\newblock Detecting hate speech on the world wide web.
\newblock \emph{NAACL-HLT 2012}, page~19.

\bibitem[{Waseem(2016)}]{DBLP:conf/acl-nlpcss/Waseem16}
Zeerak Waseem. 2016.
\newblock \href {https://doi.org/10.18653/v1/W16-5618} {Are you a racist or am {I} seeing things? annotator influence on hate speech detection on twitter}.
\newblock In \emph{Proceedings of the First Workshop on {NLP} and Computational Social Science, NLP+CSS@EMNLP 2016, Austin, TX, USA, November 5, 2016}, pages 138--142. Association for Computational Linguistics.

\bibitem[{Waseem and Hovy(2016)}]{waseem2016hateful}
Zeerak Waseem and Dirk Hovy. 2016.
\newblock Hateful symbols or hateful people? predictive features for hate speech detection on twitter.
\newblock In \emph{Proceedings of the NAACL student research workshop}, pages 88--93.

\bibitem[{Wei et~al.(2022)Wei, Wang, Schuurmans, Bosma, Xia, Chi, Le, Zhou et~al.}]{wei2022chain}
Jason Wei, Xuezhi Wang, Dale Schuurmans, Maarten Bosma, Fei Xia, Ed~Chi, Quoc~V Le, Denny Zhou, et~al. 2022.
\newblock Chain-of-thought prompting elicits reasoning in large language models.
\newblock \emph{Advances in Neural Information Processing Systems}, 35:24824--24837.

\bibitem[{Weidinger et~al.(2021)Weidinger, Mellor, Rauh, Griffin, Uesato, Huang, Cheng, Glaese, Balle, Kasirzadeh et~al.}]{weidinger2021ethical}
Laura Weidinger, John Mellor, Maribeth Rauh, Conor Griffin, Jonathan Uesato, Po-Sen Huang, Myra Cheng, Mia Glaese, Borja Balle, Atoosa Kasirzadeh, et~al. 2021.
\newblock Ethical and social risks of harm from language models.
\newblock \emph{arXiv preprint arXiv:2112.04359}.

\bibitem[{Wentworth(2020)}]{inner_alignemt_and_outer_alignment}
John Wentworth. 2020.
\newblock “{I}nner alignment failures” which are actually outer alignment failures.
\newblock \url{https://www.lesswrong.com/posts/HYERofGZE6j9Tuigi/inner-alignment-failures-which-are-actually-outer-alignment}.

\bibitem[{Wiener(1960)}]{wiener1960some}
Norbert Wiener. 1960.
\newblock Some moral and technical consequences of automation: As machines learn they may develop unforeseen strategies at rates that baffle their programmers.
\newblock \emph{Science}, 131(3410):1355--1358.

\bibitem[{Wulczyn et~al.(2017)Wulczyn, Thain, and Dixon}]{wulczyn2017ex}
Ellery Wulczyn, Nithum Thain, and Lucas Dixon. 2017.
\newblock Ex machina: Personal attacks seen at scale.
\newblock In \emph{Proceedings of the 26th international conference on world wide web}, pages 1391--1399.

\bibitem[{Xu et~al.(2023{\natexlab{a}})Xu, Sun, Zheng, Geng, Zhao, Feng, Tao, and Jiang}]{xu2023wizardlm}
Can Xu, Qingfeng Sun, Kai Zheng, Xiubo Geng, Pu~Zhao, Jiazhan Feng, Chongyang Tao, and Daxin Jiang. 2023{\natexlab{a}}.
\newblock Wizardlm: Empowering large language models to follow complex instructions.
\newblock \emph{arXiv preprint arXiv:2304.12244}.

\bibitem[{Xu et~al.(2022)Xu, He, He, and McAuley}]{xu2022leashing}
Canwen Xu, Zexue He, Zhankui He, and Julian McAuley. 2022.
\newblock Leashing the inner demons: Self-detoxification for language models.
\newblock In \emph{Proceedings of the AAAI Conference on Artificial Intelligence}, volume~36, pages 11530--11537.

\bibitem[{Xu et~al.(2023{\natexlab{b}})Xu, Song, Iyyer, and Choi}]{xu2023critical}
Fangyuan Xu, Yixiao Song, Mohit Iyyer, and Eunsol Choi. 2023{\natexlab{b}}.
\newblock A critical evaluation of evaluations for long-form question answering.
\newblock \emph{arXiv preprint arXiv:2305.18201}.

\bibitem[{Xu et~al.(2020)Xu, Ju, Li, Boureau, Weston, and Dinan}]{xu2020recipes}
Jing Xu, Da~Ju, Margaret Li, Y-Lan Boureau, Jason Weston, and Emily Dinan. 2020.
\newblock Recipes for safety in open-domain chatbots.
\newblock \emph{arXiv preprint arXiv:2010.07079}.

\bibitem[{Yang et~al.(2021)Yang, Li, Zhang, Ren, Sun, and He}]{yang2021careful}
Wenkai Yang, Lei Li, Zhiyuan Zhang, Xuancheng Ren, Xu~Sun, and Bin He. 2021.
\newblock Be careful about poisoned word embeddings: Exploring the vulnerability of the embedding layers in nlp models.
\newblock \emph{arXiv preprint arXiv:2103.15543}.

\bibitem[{Yang et~al.(2019)Yang, Dai, Yang, Carbonell, Salakhutdinov, and Le}]{DBLP:conf/nips/YangDYCSL19}
Zhilin Yang, Zihang Dai, Yiming Yang, Jaime~G. Carbonell, Ruslan Salakhutdinov, and Quoc~V. Le. 2019.
\newblock \href {https://proceedings.neurips.cc/paper/2019/hash/dc6a7e655d7e5840e66733e9ee67cc69-Abstract.html} {Xlnet: Generalized autoregressive pretraining for language understanding}.
\newblock In \emph{Advances in Neural Information Processing Systems 32: Annual Conference on Neural Information Processing Systems 2019, NeurIPS 2019, December 8-14, 2019, Vancouver, BC, Canada}, pages 5754--5764.

\bibitem[{Ye et~al.(2023)Ye, Kim, Kim, Hwang, Kim, Jo, Thorne, Kim, and Seo}]{ye2023flask}
Seonghyeon Ye, Doyoung Kim, Sungdong Kim, Hyeonbin Hwang, Seungone Kim, Yongrae Jo, James Thorne, Juho Kim, and Minjoon Seo. 2023.
\newblock Flask: Fine-grained language model evaluation based on alignment skill sets.
\newblock \emph{arXiv preprint arXiv:2307.10928}.

\bibitem[{Yuan et~al.(2023)Yuan, Yuan, Tan, Wang, Huang, and Huang}]{yuan2023rrhf}
Zheng Yuan, Hongyi Yuan, Chuanqi Tan, Wei Wang, Songfang Huang, and Fei Huang. 2023.
\newblock Rrhf: Rank responses to align language models with human feedback without tears.
\newblock \emph{arXiv preprint arXiv:2304.05302}.

\bibitem[{Yudkowsky(2004)}]{yudkowsky2004coherent}
Eliezer Yudkowsky. 2004.
\newblock Coherent extrapolated volition.
\newblock \emph{Singularity Institute for Artificial Intelligence}.

\bibitem[{Zampieri et~al.(2019)Zampieri, Malmasi, Nakov, Rosenthal, Farra, and Kumar}]{zampieri2019predicting}
Marcos Zampieri, Shervin Malmasi, Preslav Nakov, Sara Rosenthal, Noura Farra, and Ritesh Kumar. 2019.
\newblock Predicting the type and target of offensive posts in social media.
\newblock \emph{arXiv preprint arXiv:1902.09666}.

\bibitem[{Zeng et~al.(2023)Zeng, Liu, Du, Wang, Lai, Ding, Yang, Xu, Zheng, Xia, Tam, Ma, Xue, Zhai, Chen, Liu, Zhang, Dong, and Tang}]{DBLP:conf/iclr/ZengLDWL0YXZXTM23}
Aohan Zeng, Xiao Liu, Zhengxiao Du, Zihan Wang, Hanyu Lai, Ming Ding, Zhuoyi Yang, Yifan Xu, Wendi Zheng, Xiao Xia, Weng~Lam Tam, Zixuan Ma, Yufei Xue, Jidong Zhai, Wenguang Chen, Zhiyuan Liu, Peng Zhang, Yuxiao Dong, and Jie Tang. 2023.
\newblock \href {https://openreview.net/pdf?id=-Aw0rrrPUF} {{GLM-130B:} an open bilingual pre-trained model}.
\newblock In \emph{The Eleventh International Conference on Learning Representations, {ICLR} 2023, Kigali, Rwanda, May 1-5, 2023}. OpenReview.net.

\bibitem[{Zeng et~al.(2021)Zeng, Ren, Su, Wang, Liao, Wang, Jiang, Yang, Wang, Zhang, Li, Gong, Yao, Huang, Wang, Yu, Guo, Yu, Zhang, Wang, Tao, Yan, Yi, Peng, Jiang, Zhang, Deng, Zhang, Lin, Zhang, Zhang, Guo, Gu, Fan, Wang, Jin, Liu, and Tian}]{DBLP:journals/corr/abs-2104-12369}
Wei Zeng, Xiaozhe Ren, Teng Su, Hui Wang, Yi~Liao, Zhiwei Wang, Xin Jiang, ZhenZhang Yang, Kaisheng Wang, Xiaoda Zhang, Chen Li, Ziyan Gong, Yifan Yao, Xinjing Huang, Jun Wang, Jianfeng Yu, Qi~Guo, Yue Yu, Yan Zhang, Jin Wang, Hengtao Tao, Dasen Yan, Zexuan Yi, Fang Peng, Fangqing Jiang, Han Zhang, Lingfeng Deng, Yehong Zhang, Zhe Lin, Chao Zhang, Shaojie Zhang, Mingyue Guo, Shanzhi Gu, Gaojun Fan, Yaowei Wang, Xuefeng Jin, Qun Liu, and Yonghong Tian. 2021.
\newblock \href {http://arxiv.org/abs/2104.12369} {Pangu-{\(\alpha\)}: Large-scale autoregressive pretrained chinese language models with auto-parallel computation}.
\newblock \emph{CoRR}, abs/2104.12369.

\bibitem[{Zha et~al.(2023)Zha, Yang, Li, and Hu}]{zha2023alignscore}
Yuheng Zha, Yichi Yang, Ruichen Li, and Zhiting Hu. 2023.
\newblock Alignscore: Evaluating factual consistency with a unified alignment function.
\newblock \emph{arXiv preprint arXiv:2305.16739}.

\bibitem[{Zhang et~al.(2021)Zhang, Benz, Lin, Karjauv, Wu, and Kweon}]{zhang2021survey}
Chaoning Zhang, Philipp Benz, Chenguo Lin, Adil Karjauv, Jing Wu, and In~So Kweon. 2021.
\newblock A survey on universal adversarial attack.
\newblock \emph{arXiv preprint arXiv:2103.01498}.

\bibitem[{Zhang et~al.(2023)Zhang, Li, Wu, Zhang, Lin, Geng, Wang, and Fu}]{zhang2023corgi}
Ge~Zhang, Yizhi Li, Yaoyao Wu, Linyuan Zhang, Chenghua Lin, Jiayi Geng, Shi Wang, and Jie Fu. 2023.
\newblock Corgi-pm: A chinese corpus for gender bias probing and mitigation.
\newblock \emph{arXiv preprint arXiv:2301.00395}.

\bibitem[{Zhang et~al.(2022)Zhang, Roller, Goyal, Artetxe, Chen, Chen, Dewan, Diab, Li, Lin, Mihaylov, Ott, Shleifer, Shuster, Simig, Koura, Sridhar, Wang, and Zettlemoyer}]{DBLP:journals/corr/abs-2205-01068}
Susan Zhang, Stephen Roller, Naman Goyal, Mikel Artetxe, Moya Chen, Shuohui Chen, Christopher Dewan, Mona~T. Diab, Xian Li, Xi~Victoria Lin, Todor Mihaylov, Myle Ott, Sam Shleifer, Kurt Shuster, Daniel Simig, Punit~Singh Koura, Anjali Sridhar, Tianlu Wang, and Luke Zettlemoyer. 2022.
\newblock \href {https://doi.org/10.48550/arXiv.2205.01068} {{OPT:} open pre-trained transformer language models}.
\newblock \emph{CoRR}, abs/2205.01068.

\bibitem[{Zhang et~al.(2020)Zhang, Sheng, Alhazmi, and Li}]{zhang2020adversarial}
Wei~Emma Zhang, Quan~Z Sheng, Ahoud Alhazmi, and Chenliang Li. 2020.
\newblock Adversarial attacks on deep-learning models in natural language processing: A survey.
\newblock \emph{ACM Transactions on Intelligent Systems and Technology (TIST)}, 11(3):1--41.

\bibitem[{Zhao et~al.(2018)Zhao, Wang, Yatskar, Ordonez, and Chang}]{zhao2018gender}
Jieyu Zhao, Tianlu Wang, Mark Yatskar, Vicente Ordonez, and Kai-Wei Chang. 2018.
\newblock Gender bias in coreference resolution: Evaluation and debiasing methods.
\newblock In \emph{Proceedings of the 2018 Conference of the North American Chapter of the Association for Computational Linguistics: Human Language Technologies, Volume 2 (Short Papers)}, pages 15--20.

\bibitem[{Zhao et~al.(2023{\natexlab{a}})Zhao, Wen, Tuan, Zhao, and Fu}]{zhao2023prompt}
Shuai Zhao, Jinming Wen, Luu~Anh Tuan, Junbo Zhao, and Jie Fu. 2023{\natexlab{a}}.
\newblock Prompt as triggers for backdoor attack: Examining the vulnerability in language models.
\newblock \emph{arXiv preprint arXiv:2305.01219}.

\bibitem[{Zhao et~al.(2023{\natexlab{b}})Zhao, Zhou, Li, Tang, Wang, Hou, Min, Zhang, Zhang, Dong et~al.}]{zhao2023survey}
Wayne~Xin Zhao, Kun Zhou, Junyi Li, Tianyi Tang, Xiaolei Wang, Yupeng Hou, Yingqian Min, Beichen Zhang, Junjie Zhang, Zican Dong, et~al. 2023{\natexlab{b}}.
\newblock A survey of large language models.
\newblock \emph{arXiv preprint arXiv:2303.18223}.

\bibitem[{Zhao et~al.(2023{\natexlab{c}})Zhao, Joshi, Liu, Khalman, Saleh, and Liu}]{zhao2023slic}
Yao Zhao, Rishabh Joshi, Tianqi Liu, Misha Khalman, Mohammad Saleh, and Peter~J Liu. 2023{\natexlab{c}}.
\newblock Slic-hf: Sequence likelihood calibration with human feedback.
\newblock \emph{arXiv preprint arXiv:2305.10425}.

\bibitem[{Zhao et~al.(2022)Zhao, Khalman, Joshi, Narayan, Saleh, and Liu}]{zhao2022calibrating}
Yao Zhao, Misha Khalman, Rishabh Joshi, Shashi Narayan, Mohammad Saleh, and Peter~J Liu. 2022.
\newblock Calibrating sequence likelihood improves conditional language generation.
\newblock \emph{arXiv preprint arXiv:2210.00045}.

\bibitem[{Zheng et~al.(2023{\natexlab{a}})Zheng, Chiang, Sheng, Zhuang, Wu, Zhuang, Lin, Li, Li, Xing et~al.}]{zheng2023judging}
Lianmin Zheng, Wei-Lin Chiang, Ying Sheng, Siyuan Zhuang, Zhanghao Wu, Yonghao Zhuang, Zi~Lin, Zhuohan Li, Dacheng Li, Eric Xing, et~al. 2023{\natexlab{a}}.
\newblock Judging llm-as-a-judge with mt-bench and chatbot arena.
\newblock \emph{arXiv preprint arXiv:2306.05685}.

\bibitem[{Zheng et~al.(2023{\natexlab{b}})Zheng, Dou, Gao, Shen, Wang, Liu, Jin, Liu, Xiong, Chen et~al.}]{zheng2023secrets}
Rui Zheng, Shihan Dou, Songyang Gao, Wei Shen, Binghai Wang, Yan Liu, Senjie Jin, Qin Liu, Limao Xiong, Lu~Chen, et~al. 2023{\natexlab{b}}.
\newblock Secrets of rlhf in large language models part i: Ppo.
\newblock \emph{arXiv preprint arXiv:2307.04964}.

\bibitem[{Zhou et~al.(2023{\natexlab{a}})Zhou, Liu, Xu, Iyer, Sun, Mao, Ma, Efrat, Yu, Yu et~al.}]{zhou2023lima}
Chunting Zhou, Pengfei Liu, Puxin Xu, Srini Iyer, Jiao Sun, Yuning Mao, Xuezhe Ma, Avia Efrat, Ping Yu, Lili Yu, et~al. 2023{\natexlab{a}}.
\newblock Lima: Less is more for alignment.
\newblock \emph{arXiv preprint arXiv:2305.11206}.

\bibitem[{Zhou et~al.(2022)Zhou, Deng, Mi, Li, Wang, Huang, Jiang, Liu, and Meng}]{zhou2022towards}
Jingyan Zhou, Jiawen Deng, Fei Mi, Yitong Li, Yasheng Wang, Minlie Huang, Xin Jiang, Qun Liu, and Helen Meng. 2022.
\newblock Towards identifying social bias in dialog systems: Frame, datasets, and benchmarks.
\newblock \emph{arXiv preprint arXiv:2202.08011}.

\bibitem[{Zhou et~al.(2023{\natexlab{b}})Zhou, Jurafsky, and Hashimoto}]{zhou2023navigating}
Kaitlyn Zhou, Dan Jurafsky, and Tatsunori Hashimoto. 2023{\natexlab{b}}.
\newblock Navigating the grey area: Expressions of overconfidence and uncertainty in language models.
\newblock \emph{arXiv preprint arXiv:2302.13439}.

\bibitem[{Zhu et~al.(2023)Zhu, Jiao, and Jordan}]{zhu2023principled}
Banghua Zhu, Jiantao Jiao, and Michael~I Jordan. 2023.
\newblock Principled reinforcement learning with human feedback from pairwise or $ k $-wise comparisons.
\newblock \emph{arXiv preprint arXiv:2301.11270}.

\bibitem[{Ziebart et~al.(2008)Ziebart, Maas, Bagnell, Dey et~al.}]{ziebart2008maximum}
Brian~D Ziebart, Andrew~L Maas, J~Andrew Bagnell, Anind~K Dey, et~al. 2008.
\newblock Maximum entropy inverse reinforcement learning.
\newblock In \emph{Aaai}, volume~8, pages 1433--1438. Chicago, IL, USA.

\bibitem[{Ziegler et~al.(2019)Ziegler, Stiennon, Wu, Brown, Radford, Amodei, Christiano, and Irving}]{ziegler2019fine}
Daniel~M Ziegler, Nisan Stiennon, Jeffrey Wu, Tom~B Brown, Alec Radford, Dario Amodei, Paul Christiano, and Geoffrey Irving. 2019.
\newblock Fine-tuning language models from human preferences.
\newblock \emph{arXiv preprint arXiv:1909.08593}.

\bibitem[{Zou et~al.(2023)Zou, Wang, Kolter, and Fredrikson}]{zou2023universal}
Andy Zou, Zifan Wang, J~Zico Kolter, and Matt Fredrikson. 2023.
\newblock Universal and transferable adversarial attacks on aligned language models.
\newblock \emph{arXiv preprint arXiv:2307.15043}.

\end{thebibliography}
\bibliographystyle{acl_natbib}

\end{document}